\documentclass[twoside,11pt]{article}

\usepackage{jmlr2e}

\usepackage{amsmath,amssymb,amsfonts,graphicx,nicefrac,mathtools,bm}
\usepackage{hyperref}
\usepackage[english]{babel}
\usepackage{times}
\usepackage[T1]{fontenc}
\usepackage{bbm}
\usepackage{color}
\usepackage{xspace}
\usepackage{rotating,multirow}
\usepackage{graphicx,subfigure}
\usepackage{xfrac}
\usepackage{booktabs}
\usepackage{arydshln}
\usepackage[]{algorithm2e}
\SetKwInOut{Parameter}{Parameters}
\usepackage{adjustbox}

\newtheorem{assumption}{Assumption}

\DeclareMathOperator{\diag}{diag}

\DeclareMathOperator{\supp}{support}

\DeclareMathOperator*{\argmin}{arg\,min}

\newcommand{\norm}[1]{\lVert{#1}\rVert}

\newcommand{\Abs}[1]{\left\lvert{#1}\right\rvert}
\newcommand{\abs}[1]{\lvert{#1}\rvert}
\newcommand{\PP}[1]{\mathbb{P}\left\{{#1}\right\}} 
\newcommand{\EE}[1]{\mathbb{E}\left[{#1}\right]} 

\newcommand{\PPst}[2]{\mathbb{P}\left\{{#1}\ \middle| \ {#2}\right\}} 

\renewcommand{\O}[1]{\mathcal{O}\left({#1}\right)}
\def\R{\mathbb{R}}

\newcommand{\ident}{I}
\newcommand{\ones}{\mathbf{1}}

\newcommand{\iidsim}{\stackrel{\mathrm{iid}}{\sim}}

\newcommand{\ignore}[1]{}

\newcommand{\xh}{\widehat{x}}

\newcommand{\xhr}{\widehat{x}^{\textnormal{R}}}
\newcommand{\xcr}{\check{x}^{\textnormal{R}}}

\newcommand{\ph}{\widehat{x}}

\newcommand{\Acal}{\mathcal{A}}
\newcommand{\xs}{x^{\star}}

\newcommand{\seed}{\mathbf{s}}

\newcommand{\E}[1]{\mathbb{E}[{#1}]} 
\newcommand{\vol}{\textnormal{Vol}}

\newcommand{\lap}{L}

\newcommand{\fp}{f}

\ShortHeadings{Statistical guarantees for local graph clustering}{W.~Ha and K.~Fountoulakis and M.~Mahoney}
\firstpageno{1}

\begin{document}

\title{Statistical guarantees for local graph clustering}

\author{\name Wooseok~Ha\thanks{Equal contribution}\textsuperscript{ \hspace{0.1cm}} \email haywse@berkeley.edu \\ 
        \addr Department of Statistics, \\ University of California at Berkeley, \\ Berkeley, CA, USA.
        \AND
        \name Kimon~Fountoulakis\footnotemark[1]\textsuperscript{ \hspace{0.1cm}} \email kfountou@uwaterloo.ca \\ 
        \addr School of Computer Science, \\ University of Waterloo, \\ Waterloo, ON, Canada.
        \AND
        \name Michael~W.~Mahoney \email mmahoney@stat.berkeley.edu \\ 
        \addr ICSI and Department of Statistics, \\ University of California at Berkeley, \\ Berkeley, CA, USA.
}

\editor{}

\maketitle

\begin{abstract}
Local graph clustering methods aim to find small clusters in very large graphs.\ 
These methods take as input a graph and a seed node, and they return as output a good cluster in a running time that depends on the size of the output cluster but that is independent of the size of the input graph.\
In this paper, we adopt a statistical perspective on local graph clustering, and we analyze the performance of the $\ell_1$-regularized PageRank method~\citep{fountoulakis2019variational} for the recovery of a single target cluster, given a seed node inside the cluster.\
Assuming the target cluster has been generated by a random model, we present two results.\
In the first, we show that the optimal support of $\ell_1$-regularized PageRank recovers the full target cluster, with bounded false positives.\
In the second, we show that if the seed node is connected solely to the target cluster then the optimal support of $\ell_1$-regularized PageRank recovers exactly the target cluster.\ 
We also show empirically that $\ell_1$-regularized PageRank has a state-of-the-art performance on many real graphs, demonstrating the superiority of the method.
From a computational perspective, we show that the solution path of $\ell_1$-regularized PageRank is monotonic.\ This allows for the application of the
forward stagewise algorithm, which approximates the solution path in running time that does not depend on the size of the whole graph.\
Finally, we show that $\ell_1$-regularized PageRank and approximate personalized PageRank (APPR)~\citep{andersen2006local}, another very popular method for local graph clustering, are equivalent in the sense that we can lower and upper bound the output of one with the output of the other.
Based on this relation, we establish for APPR similar results to those we establish for $\ell_1$-regularized PageRank. 
\end{abstract}

\begin{keywords}
  clustering, local graph clustering, Page-Rank, seed expansion, $\ell_1$-regularization.
\end{keywords}

\section{Introduction}
In many data applications, one is interested in finding small-scale structure in a very large data set.\
As an example, consider the following version of the so-called \emph{local graph clustering problem}: given a large graph and a seed node in that graph, quickly find a good small cluster that includes that seed node.\ 
From an algorithmic perspective, one typically considers worst-case input graphs, and one is interested in running time guarantees, e.g., to find a good cluster in a time that depends linearly or sub-linearly on the size of the entire graph.\
From a statistical perspective, such a local graph clustering problem can be understood as a recovery problem.\ 
One assumes that there exists a target cluster in a given large graph, where the graph is assumed to have been generated by a random model, and the objective is to recover the target cluster from one node inside the cluster.\ 

In this paper, we consider the so-called \emph{$\ell_1$-regularized PageRank algorithm}~\citep{fountoulakis2019variational}, a popular algorithm for the local graph clustering problem, and we establish statistical recoverability guarantees for it.\ 
Previous theoretical analysis on local graph clustering, e.g.,~\citet{andersen2006local,zhu2013local}, is based on the notion of conductance (a cluster quality metric that considers the internal versus external connectivity of a cluster) and considers running time performance for worst-case input graphs.\ 
In contrast, our goal will be to study the average-case performance of the $\ell_1$-regularized PageRank algorithm, under a certain type of a local random graph model.\ 
The model we consider is very general; it concerns the target cluster and its adjacent nodes; and it encompasses the stochastic block model~\citep{holland1983stochastic,abbe2017community} and the planted clustering model~\citep{alon1998finding,arias2014community} as special~cases.\ 

Within this random graph model, we provide theoretical guarantees for the unique optimal solution of the $\ell_1$-regularized PageRank optimization problem.\
In particular, the cluster is recovered through the support set of the $\ell_1$-regularized PageRank vector and we give rigorous bounds on the false positives and false negatives of the recovered cluster.\
Furthermore, observe that our statistical perspective is more aligned with statistical guarantees for the sparse regression problem (and the lasso problem~\citep{tibshirani1996regression}), where the objective is to recover the true parameter and/or support from noisy data.\ 
Given this connection, we also establish a result for the exact support recovery of $\ell_1$-regularized PageRank. Empirically we demonstrate the ability of the method to recover the target cluster in a range of real-world data graphs.\
Finally, we establish an equivalence between $\ell_1$-regularized PageRank and the very popular local graph clustering algorithm from~\citet{andersen2006local,zhu2013local}.\ 
This allows us to prove similar average case guarantees for the algorithm of~\citet{andersen2006local,zhu2013local} as well.

\subsection{Literature review}


The origins of local graph clustering are with the work of \citet{spielman2013local}.\ 
Subsequent to their original results, there has been a great deal of follow-up work on local graph clustering procedures, including with random walks \citep{andersen2006local}, local Lanczos spectral approximations \citep{shi2017local}, evolving sets \citep{andersen2016almost}, seed expansion methods \citep{kloumann2014community}, optimization-based approaches \citep{fountoulakis2019variational,fountoulakis2017optimization}, and local flow methods \citep{wang2017capacity}.\ 
There also exist  local higher-order clustering \citep{yin2017local},  linear algebra approaches \citep{shi2017local}, spectral methods based on Heat Kernel PageRank \citep{kloster2014heat}, newer seed set selection techniques for local flow methods \citep{veldt2019flow}, and parallel local spectral approaches \citep{shun2016parallel}.\ 
In all of these cases, given a seed node, or a seed set of nodes, the goal of existing local graph clustering approaches is to compute a cluster ``nearby'' the seed that is related to the ``best'' cluster nearby the seed.\ 
Here, ``best'' and ``nearby'' are intentionally left under-specified, as they can be formalized in one of a few different but related ways.\ 
For example, ``best'' is usually related to a clustering score such as conductance.\
In fact, many existing methods for local graph clustering with theoretical guarantees are motivated through the problem of finding a cluster that is near the seed node and that also has small conductance value~\citep{spielman2013local,andersen2006local,fountoulakis2019variational,veldt2019flow,wang2017capacity}.\ 
Recently, \citet{green2019local} studied local graph clustering from a traditional statistical learning setup to identify density clusters.

There are also numerous papers in statistics on partitioning random graphs.\ 
Arguably, the stochastic block model (SBM) is the most commonly employed random model for graph partitioning, and it has been extensively studied~\citep{abbe2015community,abbe2015exact,zhang2016minimax,massoulie2014community,mossel2018proof,newman2002random,mossel2015reconstruction,rohe2011spectral,amini2018semidefinite,abbe2017community}.\ 
Recent work has also generalized the SBM to a degree-corrected block model, to capture degree heterogeneity of the network~\citep{chen2018convexified,gulikers2017spectral,zhao2012consistency,gao2018community}.\ 
The literature in this area is too extensive to cover in this paper, but we refer the readers to excellent survey papers on the graph partitioning problem~\citep{abbe2017community}.\ 

We should emphasize that the traditional graph partitioning problem is quite different than the local graph clustering problem that we consider in this paper.\
Among other things, the former partitions all the vertices of a graph into different clusters, while for the latter problem our objective is to find a single cluster given a seed node in the cluster; the former takes as input a graph, while the latter takes as input a graph and a seed set of nodes; and the former runs in time depending on the size of the graph, while the latter runs in time depending on the size of the output, but is otherwise independent of the size of the graph.\

\subsection{Notation}

We write $[n]=\{1,\dots,n\}$ for any $n\geq 1$.\ 
Throughout the paper we assume  we have a connected, undirected graph $G=(V,E)$, where $V$ denotes the set of nodes, with $\Abs{V}=n$, and  $E\subset (V\times V)$ denotes the set of edges.\ We denote by $A$ the adjacency matrix of $G$, i.e., $A_{ij}=w_{ij}$ if $(i,j)\in E$, and $0$ otherwise. For an unweighted graph, $w_{ij}$ is set to $1$ for all $(i,j)\in E$.\ We denote by $D$  the diagonal degree matrix of $G$, i.e., $D_{ii}\coloneqq d_i =\sum_{j:(i,j)\in E}w_{ij}$, where $d_i$ is the weighted degree of node $i$.\ In this case, $d=(d_i)\in\R^n$ denotes the degree vector, and the volume of a subset of nodes is defined as $\vol(B) = \sum_{i\in B}d_i$ for $B\subseteq V$. We denote by $\lap = D-A$ the graph Laplacian; and $Q\coloneqq \alpha D + \frac{1-\alpha}{2}\lap$.  

For given sets of indexes $B_1,B_2 \subseteq [n]$, we write $M_{B_1,B_2}$ to denote the submatrix of $M$ indexed by $B_1$ and $B_2$. If $B_1=\{i\}$ is a singleton, we use $M_{i,B_2}$ to indicate the $i$-th row of $M$ whose columns are indexed by $B_2$.\ Analogously, we use $M_{B_1,j}$ to indicate the $j$-th column of $M$ whose rows are indexed by $B_1$.\ We denote by $B_1\backslash B_2$ a set difference between $B_1$ and $B_2$, and denote by $B_1^c=[n]\backslash B_1$ the complement of $B_1$.

\section{Local graph clustering from a variational point of view}
In this section, we briefly review and motivate $\ell_1$-regularized PageRank~\citep{fountoulakis2019variational}, an optimization formulation of local graph clustering to find a local cluster around a given seed node. 
We study some properties of the output vector of  $\ell_1$-regularized PageRank and illuminate a novel connection to approximate personalized PageRank (APPR)~\citep{andersen2006local}.

\subsection{Background on $\ell_1$-regularized PageRank}\label{sec:background}

PageRank~\citep{page1999pagerank,brin1998anatomy} is a popular approach for ranking the importance of the nodes given a graph.\ 
It is defined as the stationary distribution of a Markov chain, which is encoded by a convex combination of the input distribution $\seed \in\R^n$ and the (lazy) random walk on the graph, i.e.,\
\begin{equation}\label{eqn:pagerank}
p^{\text{PR}} = \alpha \seed + (1-\alpha) W p^{\text{PR}}, 
\end{equation}
where $W=(\ident + AD^{-1})/2$ is the lazy random walk operator and where $\alpha\in (0,1)$ is the teleportation parameter.\ 
To measure the ranking or importance of the nodes of the ``whole'' graph, PageRank  is often computed by setting the input vector $\seed$ to be a uniform distribution over $\{1,2,\ldots,n\}$.\

For local graph clustering, where the aim is to identify a target cluster, given a seed node in the cluster, the input distribution $\seed$ is set to be equal to one for the seed node and zero everywhere else.\ 
For example, when the node $i$ is given as the seed node, we consider the input distribution $\seed$ to be the discrete Dirac measure such that $\seed_i = 1$ and zero elsewhere.\
This ``personalized'' PageRank \citep{haveliwala2002topic} measures the closeness or similarity of the nodes to the given seed node, and it outputs  a ranking of the nodes that is ``personalized'' with respect to the seed node (as opposed to the original PageRank, which considers the entire graph).\ 
From an operational point of view, the underlying diffusion process in~\eqref{eqn:pagerank} defining personalized PageRank performs a lazy random walk with probability $1-\alpha$ and ``teleports'' a random walker back to the original seed node with probability $\alpha$.\

From the definition itself, the personalized PageRank vector can be obtained by solving the linear system~\eqref{eqn:pagerank}. Unfortunately, this step can be prohibitively expensive, especially when there is a single seed node or a small seed set of seed nodes, and when one is interested in very small clusters in a very large graph.\ 
In the seminal work of~\citet{andersen2006local}, the authors propose an iterative algorithm, called \emph{approximate personalized PageRank (APPR)}, to solve this running time problem.
They do so by approximating the personalized PageRank vector, while running in time \emph{independent} of the size of the entire graph.\
APPR was developed from an algorithmic (or ``theoretical computer science'') perspective, but it is equivalent to applying a coordinate descent type algorithm to the linear system~\eqref{eqn:pagerank} with a particular scheme of early stopping (see Section~\ref{sec:APPR} for more details on the APPR algorithm).\ 
Motivated by this, \citet{fountoulakis2019variational} proposed the \emph{$\ell_1$-regularized PageRank optimization problem}.\ 
Unlike APPR, the solution method for the $\ell_1$-regularized PageRank optimization problem is purely optimization-based.\
It uses an $\ell_1$ norm regularization to set automatically to be zero all nodes that are dissimilar to the seed node, thereby resulting in a highly sparse output.\ 
In this manner, $\ell_1$-regularized PageRank can estimate the personalized ranking, while maintaining the most relevant nodes at the same time.\ 
Prior work \citep{fountoulakis2019variational} also showed that proximal gradient descent (ISTA) can solve the $\ell_1$-regularized PageRank minimization problem, with access to only a small portion of the entire graph, i.e., without even touching the entire graph, therefore allowing the method to easily scale to very large-scale graphs~\citep{shun2016parallel}.\

In this paper, we investigate the statistical performance of $\ell_1$-regularized PageRank by reformulating the local graph clustering into the problem of sparse support recovery.\ 
Here we give a more precise definition of the $\ell_1$-regularized PageRank optimization problem from~\citet{fountoulakis2019variational} that we consider.\


\begin{definition}[\textbf{$\ell_1$-regularized PageRank}]
\label{def:pagerank2}
Given a graph $G=(V,E)$, with $|V|=n$, and a seed vector $\seed \in\mathbb{R}^{n}$, the $\ell_1$-regularized PageRank  vector on the graph is defined as
\begin{equation}
\label{eqn:pagerank2}
\ph = \argmin_{x\in\R^n} \left\{ \underbrace{\frac{1}{2}x^\top Q x - \alpha x^\top \seed }_{\coloneqq \fp(x)}+ \rho\alpha \norm{D x}_1\right\} ,
\end{equation}
where recall $Q=\alpha D + \frac{1-\alpha}{2}\lap$, and where $\rho>0$ is a user-specified parameter that controls the amount of the regularization. 
\end{definition}
Note that our definition of $\ell_1$-regularized PageRank is consistent with the  original formulation~\citep[Equation (8)]{fountoulakis2019variational} by change of variables $D^{1/2}x=q$.
To better understand the objective function of~\eqref{eqn:pagerank2}, when the regularization parameter $\rho$ is set to $0$,  one can easily check that the re-scaled version of the output solution  $D\xh$ recovers the original PageRank solution, that is, $D\xh = p^{\text{PR}}$ satisfies the stationary equation~\eqref{eqn:pagerank}. 
For the stationary personalized PageRank vector $p^{\text{PR}}$, mass is concentrated around the seed node, meaning that after ordering, it has long tail for nodes far away from the seed node.
Importantly, we can then efficiently cut this tale  using $\ell_1$ norm regularization, without even having to compute the long tail. 


\subsection{Properties}

Here, we state some properties of the $\ell_1$-regularized PageRank vector that will be useful for our analysis, as well as for gaining further insight into the method.
The proof of these lemmas can be found in Appendix~\ref{sec:additional_proof}.

First, the following lemma guarantees that the $\ell_1$-regularized PageRank vector is non-negative.\
This result should be natural, because $\ell_1$-regularized PageRank computes the importance scores of the nodes relative to the seed node, which cannot be negative.

\begin{lemma}[\textbf{Non-negativity of $\ell_1$-reg. PageRank.}]
\label{lem:nonneg}
Let $\xh$ be the optimal output solution given in Definition~\ref{def:pagerank2}. Then $\xh$ is  non-negative, i.e., $\xh_j\geq 0$ for all $j\in V$.\
\end{lemma}

The next lemma guarantees that the gradient of $\fp$ at the optimal solution $\xh$ must be non-positive. Since the $\ell_1$-regularized PageRank problem is strongly convex (the minimum eigenvalue of $Q$ is $>0$), by KKT condition, this characterization is both necessary and sufficient  for $\xh$ to be the unique solution. We frequently  use this lemma  in the proof of our subsequent results.\

\begin{lemma}[\textbf{Optimality condition for $\ell_1$-reg. PR minimization.}]
\label{lem:nonneg_grad}
Let $\supp(\ph):=\{i \in V \ | \ \ph_i\neq 0\}$ be the support set of the optimal solution.\ Then
\begin{align*}\label{eqn:opt_cond}
&\nabla_i\fp(\xh)=(Q \xh)_i -\alpha \seed{}_i =\nonumber
&\begin{cases} -\rho\alpha d_i, & \mbox{if } \ \xh_i >0 , \\
		       \in [ -\rho\alpha d_i, 0], & \mbox{if } \ \xh_i = 0 \text{ and $i$ is a neighbor of nonzero node}, \\
		       0, & \text{otherwise.}
\end{cases}
\end{align*}
\end{lemma}

Finally, the next result shows that the solution path of the $\ell_1$-regularized PageRank problem is monotone, meaning that when the output of $\ell_1$-regularized PageRank  is parametrized as a function of $\rho>0$, the trajectory $\xh(\rho)$ changes in a monotonic manner as  $\rho$ varies.
\begin{lemma}[\textbf{Monotonicity of $\ell_1$-reg. PageRank.}]\label{lem:monotone}
Let $\xh(\rho)$ denote the solution for~\eqref{eqn:pagerank2} indexed by $\rho>0$. Then, $\ph(\rho)$ is monotone as a function of $\rho$, i.e., $\ph(\rho_0) \leq \ph(\rho_1)$ whenever $\rho_0 > \rho_1$, where $\leq $ is applied component-wise. 
\end{lemma}
Lemma~\ref{lem:monotone} shows that once a node enter the model at some number $\rho>0$,  then it will never leave the model thereafter. 
This result is  intuitive since the importance score of the nodes will increase as the amount of regularization or shrinkage decreases.
In Section~\ref{sec:stagewise}, we will use this monotonic property in a crucial way to develop an iterative algorithm that approximates the entire solution path.

\subsection{Connection to approximate personalized PageRank (APPR)}\label{sec:APPR}

 
Approximate personalized PageRank (APPR), first studied in~\citet{andersen2006local} and followed up by a number of subsequent works, is at the heart of local graph clustering whose central idea is to approximate the original PageRank vector without ever touching the entire graph. 
 \citet{fountoulakis2019variational} observed that the APPR algorithm is equivalent to the iterative coordinate descent solver applied to the linear system~\eqref{eqn:pagerank} with a specifically designed termination criterion. 
 In our notation, we can write the algorithm in a simple and compact way:
 \begin{itemize}
 \item Given a parameter $\rho>0$, initialize $x^{(0)}=0$;
 \item For each $k\geq0$, while $\norm{D^{-1}\nabla f(x^{(k)})}_\infty \geq \rho\alpha$, iterate the steps\footnote{It can be shown that if APPR is initialized at $x^{(0)}=0$ then $\nabla f(x^{(k)})\leq 0$ for all $k\geq 0$. In this case, the termination criterion of APPR reads as $\nabla_i f(x^{(k)})> -\rho\alpha d_i$ for all $i\in [n]$.}
 \begin{equation}\label{eqn:appr}
\begin{cases}
\text{Choose an $i\in [n]$ such that $\nabla_i f(x^{(k)})\leq-\rho\alpha d_i$};\\
\text{Update $x^{(k+1)}_i=x^{(k)}_i - d_i^{-1}\nabla_i f(x^{(k)})$. }
\end{cases}
\end{equation}
\item Output $x^{(k^\star)}=:x^{\text{APPR}}$.
 \end{itemize}
 The output of the APPR algorithm is the personalized score vector $x^{\text{APPR}}$ that efficiently approximates the original PageRank. 
Note that the updating formula~\eqref{eqn:appr} requires one evaluation of the gradient which can be done by accessing only neighboring nodes of the selected entry $i$.
As a result the APPR algorithm can solve the PageRank equation~\eqref{eqn:pagerank} extremely efficient; and in particular  the running time of the algorithm depends on the size of the output rather than the size of the whole~graph.

APPR was  developed purely from an algorithmic perspective and the output of the algorithm depends on which coordinate $i$ is chosen at every iteration (see the updating step~\eqref{eqn:appr}). This property, while allowing the algorithm to enjoy the {\em locality} property and resulting in the output  vector that is highly sparse, can lead to different results of local clustering in principle (albeit the results may look more or less similar). On the other hand, $\ell_1$-regularized PageRank eliminates this issue by decoupling the locality/sparsity of the local clustering output vector from the algorithmic issue of running in time independent of the size of the graph. In particular, any convex optimization algorithm that exploits the locality of the problem, such as proximal gradient descent (ISTA), can be employed to minimize the $\ell_1$-regularized objective function~\eqref{eqn:pagerank2} while touching only the neighbors of the current non-zero nodes. Therefore the formulation via $\ell_1$-regularized PageRank  achieves two objectives of the graph processing that are desirable for local graph clustering, i.e., both the locality/sparsity of the solution and the ``strongly local'' running time of the algorithm. 

It is shown in~\citet{fountoulakis2019variational} that $\ell_1$-regularized PageRank can be viewed as a variational formulation of APPR, in the sense that the optimality condition for the $\ell_1$-regularized objective function  (Lemma~\ref{lem:nonneg_grad}) implies the termination criterion of APPR, i.e., $\norm{D^{-1}\nabla f(x^{(k)})}_\infty <\rho\alpha$. 
In this work, we strengthen the connection between the output of APPR and the output of $\ell_1$-regularized PageRank. 
In particular, we show that with appropriate choices of the regularization parameters, both approaches become equivalent in terms of cluster recovery.
\begin{theorem}[\textbf{Equivalence between $\ell_1$-reg. PR and APPR.}]\label{thm:appr_vs_l1}
Let $\ph(\rho)$ be the $\ell_1$-regularized PageRank vector~\eqref{eqn:pagerank2}, let $x^{\text{APPR}}(\rho)$ be the output of APPR~\eqref{eqn:appr} at the regularization parameter $\rho$.
Then for any $\rho_0>0$, 
\[\supp(\ph(\rho_0)) \subseteq \supp(x^{\text{APPR}}(\rho_0)) \subseteq \supp(\ph((1-\alpha)\cdot \rho_0/2)) .\]
That is, for any parameter $\rho=\rho_0>0$, the support set of the output of APPR is, respectively, a superset and a subset of that of $\ell_1$-regularized PageRank at $\rho=\rho_0$ and $\rho=(1-\alpha)\cdot \rho_0/2$.
\end{theorem}
Theorem \ref{thm:appr_vs_l1} establishes rather a stronger equivalence between the $\ell_1$-regularized PageRank and the APPR approaches than was established in the prior work of~\citet{fountoulakis2019variational}. 
Importantly, using this result, various statistical guarantees for the output cluster of APPR directly follow from the results we establish for the output of $\ell_1$-regularized PageRank. 
While analyzing   APPR's output in the random graph setting involves technical challenges due to its algorithmic nature, the fact that the output cluster of $\ell_1$-regularized PageRank is given by the optimization solution allows us to analyze statistical properties more easily.
In the next section, we will study the recovery guarantees for $\ell_1$-regularized PageRank under a random graph model, and we will show how this guarantee can be transferred to the output cluster of APPR using Theorem \ref{thm:appr_vs_l1}.

\section{Statistical guarantees under random model}
\label{sec:statistical_guarantee}


 We begin by introducing a random model that we consider for generating a target cluster. In particular we will assume the graph is generated according to the following model.
\begin{definition}[\textbf{Local random model.}]
\label{def:model}
Given a graph $G=(V,E)$ that has $n$ vertices, let $K\subseteq V$ be a target cluster inside the graph, and let $K^c$ denote the complement of $K$.\
If two vertices $i$ and $j$ belong to $K$, then we draw an edge between $i$ and $j$ with probability $p$, independently of all other edges; if $i\in K$ and $j\in K^c$, then we draw an edge with probability $q$, independently of all other edges; and otherwise, we allow any (deterministic or random) model to generate edges among vertices in $K^c$.\ 
\end{definition}

\noindent
According to Definition \ref{def:model}, the adjacency matrix $A\in\R^{n\times n}$ is symmetric, and for any $i,j\in V$, we have that $A_{ij}$ is  an independent draw from a Bernoulli distribution with probability $p$ if $i,j\in K$, and from a Bernoulli distribution with probability $q$ if $i\in K$ and $j\in K^c$.
For the rest of the graph, i.e., when both $i$ and $j$ belong to $K^c$, $A_{ij}$ can be generated from an arbitrary fixed model.\
Under this definition, we can also naturally define the population version of the graph, which is the graph induced by the expected adjacency matrix $\EE{A}$, where the expectation is taken with respect to the distribution defined by our  random model.
That is, the population graph is an undirected graph $\overline{G}=(V,E)$ whose adjacency matrix is $\EE{A}$,~where
\begin{equation}\label{eqn:adj_avg}\EE{A_{ij}} = \begin{cases} p & \text{if $i\in K$ and $j\in K$}, \\
q &  \text{if $i\in K$ and $j\in K^c$},\\
\text{Any value} &  \text{if $i\in K^c$ and $j\in K^c$}.
 \end{cases} \end{equation}
The expected degree matrix is similarly denoted by $\EE{D}$ and the expected graph Laplacian is defined as $\EE{\lap} = \EE{D} - \EE{A}$. 
The model in Definition \ref{def:model} allows us to formulate the problem of local graph clustering as the recovery of a target cluster.\
Since we are interested in recovering a single target cluster, it is natural to make assumptions only for nodes in the target cluster and nodes adjacent to the target cluster, and to leave the interactions between other nodes unspecified.\

This random model is fairly general, and it covers several popular random graph models appearing in the literature, including the stochastic block model (SBM)~\citep{holland1983stochastic,abbe2017community} and the planted clustering model~\citep{alon1998finding,arias2014community,chen2016statistical}.\
For instance, if the subgraph with the vertices within $K^c$ is generated from the SBM, then the entire graph $G=(V,E)$ follows the SBM.
On the other hand, if the subgraph of $K^c$ is generated from the classical Erd\H{o}s-R\'enyi model with probability $q$, the entire graph $G=(V,E)$ follows the Planted Densest Subgraph (in this case nodes in $K^c$ do not belong to any clusters).\ 
Hence, the results we obtain here for our model holds more broadly across these different random graph models.\

Before we move on to our results, we need additional piece of notation.\ 
We write $S\subseteq K$ to denote a singleton of the given seed node.\ Let $k=\Abs{K}$ denote the cardinality of the target cluster.\ According to our local model, any node in the target cluster has the same expected degree, $\EE{d_i}=p(k-1)+q(n-k)$ for all $i\in K$, which we denote by $\bar{d}$.\ For the nodes $\ell$ outside $K$, we write $\EE{d_\ell}$ to denote its expected degree, where the expectation is taken with respect to a distribution that generates the graph in $K^c$.\ 
Conductance measures the weight of the edges that are being removed over the volume of the cluster---formally it is defined as the ratio $\mbox{Cut}(S,S^c)/\min\left(\vol(S), \vol(S^c)\right)$, where $\mbox{Cut}(S,S^c):=\sum_{i\in S,j\in S^c} A_{ij}$.\
From Definition~\ref{def:model}, the conductance of the target cluster of the population graph $\overline{G}$ is given by
\begin{equation}
\label{eqn:gamma} 
\overline{\text{Cond}} = 1-\gamma, \;\text{ where } \gamma \coloneqq \frac{p\cdot (k-1)}{\bar{d}} \in (0,1).
\end{equation}
Here $\gamma$ can be viewed as the ratio of the random walker staying inside $K$ under the population graph.\
(Note $\bar{d}$ is the expected degree of the target cluster and $p\cdot (k-1)$ is the expected degree of the target cluster when restricted to the subgraph within $K$.)\ 

As in the worst-case analysis of local graph clustering \citep{andersen2006local,zhu2013local}, conductance $\overline{\text{Cond}}$, or equivalently the number $\gamma$, will play a crucial role in determining the behavior of $\ell_1$-regularized PageRank under the local graph model.\
In particular, we see that in the extreme scenario where $\gamma=1$, we have $q=0$ indicating perfect separability of the target cluster from the rest, while for $\gamma=0$, we have $p=0$ meaning there is no signal to ever recover.\ 
With this definition, we can also write $p (k-1)=\gamma\bar{d}$ and $q(n-k) = (1-\gamma)\bar{d}$.\


\subsection{Recovery of target cluster with bounded false positives}
\label{sec:full_recovery}

Here, we investigate the performance of $\ell_1$-regularized PageRank on the graph generated by the local random model as in Definition \ref{def:model}, and we state two of our main theorems.\ 


Our first main result guarantees  full recovery of the target cluster for an appropriate choice of the regularization parameter.\
Specifically, for $\delta>0$, define 
\begin{equation}
\label{eqn:rho}
\rho(\delta) \coloneqq \left(\frac{1-\alpha}{1+\alpha}\right)^2  \left(\frac{1-\delta}{1+\delta} \right)^2 \frac{\gamma p}{(1+\delta) \bar{d}^2} = \O{\frac{\gamma p}{\bar{d}^2}},
\end{equation}
where $\alpha$ is the teleportation constant and where $p,\gamma, \bar{d}$ are the parameters of the random model defined in~\eqref{eqn:gamma}.
Then, if we solve the convex problem~\eqref{eqn:pagerank2} with $\rho\leq \rho(\delta)$, the optimal  solution fully recovers the target cluster $K$, as long as the seed node is initialized inside $K$.\

\begin{theorem}[\textbf{Full recovery.}]
\label{thm:full_recovery}
Suppose that $p^2 k \geq \O{\delta^{-2} \log k}$.\ 
If the regularization parameter satisfies $\rho\leq \rho(\delta)$ where $\rho(\delta)$ is defined in~\eqref{eqn:rho}, then the solution to Problem~\eqref{eqn:pagerank2} fully recovers the cluster $K$, i.e.,
\[K\subseteq \supp(\xh),\]
with probability at least $1-6\exp(-\O{\delta^2 p^2 k})$.\footnote{More precisely, we assume $(1-\delta)p^2k\geq c_0^{-1}\delta^{-2}\log k$ for a fixed constant $c_0>0$. Then with probability at least $1-6e^{-c_0 \delta^2 (1-\delta)p^2 k}$, the statement in the theorem holds.} 
\end{theorem}

Our next main result provides an upper bound on the false positives present in the support set of the $\ell_1$-regularized PageRank vector.\
By ``false positives'', we mean the nonzero nodes that belong to $K^c$.\ We measure the size of false positives using a notion of volume, where we recall the volume of a subset of vertices $B\subseteq V$ is given by $\vol(B) = \sum_{i\in B}d_i$.\

\begin{theorem}[\textbf{Bounds on false positives.}]
\label{thm:bound_FP}
Suppose the same conditions as in Theorem \ref{thm:full_recovery}.\
If the regularization parameter satisfies $\rho\geq\rho(\delta)$, then the solution to Problem~\eqref{eqn:pagerank2} satisfies the bound
\begin{equation}
\label{eqn:bound_FP}
\vol(\text{FP}) \leq \vol(K) \big[ \underbrace{\left(\frac{1+\alpha}{1-\alpha}\right)^2  \left(\frac{1+\delta}{1-\delta} \right)^3 \frac{1}{\gamma^2 }  -1}_{=\O{\frac{1}{\gamma^2}} - 1} \big],
\end{equation}
with probability at least $1-6\exp(-\O{\delta^2 p^2 k})$,\footnote{The same probability bound holds as in Theorem \ref{thm:full_recovery}.} where $\text{FP}=\{i\in \supp(\xh): i\in K^c \}$ is the collection of false positive nodes.\
\end{theorem}
The above results, Theorem \ref{thm:full_recovery} and Theorem \ref{thm:bound_FP}, show several regimes where $\ell_1$-regularized PageRank can fully recover the target cluster with nonvanishing probability.\ 
In particular, when $p=\O{1}$, the size of the target cluster, $k$, is required to be larger than $\O{\log k}$, which includes the constant size $k=\O{1}$.\ 
This is often the regime of interest for local graph clustering, where the goal is to find small- and meso-scale clusters in massive graphs \citep{leskovec2009community,leskovec2010empirical}.\ 
In addition, Theorem \ref{thm:full_recovery} indicates that if $\gamma$ is small,  we need to set $\rho$ to be small to recover the entire cluster.\ 
Intuitively, more mass will leak out to $K^c$ for small $\gamma$, so we need to run more steps of random walk (equivalently a smaller $\rho$  in our optimization framework) to find the right cluster.\ 
However, this means that the $\ell_1$-regularized PageRank vector will also pick up many nonzero nodes in $K^c$, resulting in many false positives in the support set.\ 
Indeed, Theorem \ref{thm:bound_FP} shows that the volume of false positives grows quadratically as $1/\gamma$, so we need $\gamma$ to be well-bounded to get a meaningful recovery from local clustering.\ 
In the case of $p=\O{1}, k=\O{1}$, this amounts to requiring that $q=\O{\frac{1}{n}}$ in order for the recovered cluster to keep high mass  inside $K$.\

It is also worth making several other comments regarding the results.
First, we suspect the current bound we obtain in~\eqref{eqn:bound_FP} may not be tight with respect to $\alpha$ and other constants, and especially the factor $(\frac{1+\alpha}{1-\alpha})^2$ may be an artifact of our proof.\
Studying the lower bound on the performance of the method, as well as obtaining an improved bound on false positives, is therefore an interesting future direction to pursue.\ 
Furthermore,  on the basis of our empirical results, $\ell_1$-regularized PageRank performs well across a broad range of $\alpha$ values, and we have not seen much difference in terms of performance among different $\alpha$'s.\ 
The role of $\alpha$ in $\ell_1$-regularized PageRank is closely tied to the regularization parameter $\rho$, and we leave the question of selecting optimal $\alpha$ for future work.\



\subsection{Exact recovery of target cluster with no false positives}
\label{sec:no_false_pos}

Next, we study the scenarios under which $\ell_1$-regularized PageRank can exhibit a stronger recovery guarantee. Specifically, under some additional conditions, we show that the support set of the optimal solution~\eqref{eqn:pagerank2} identifies the target cluster exactly, without making any false positives.\ 
For this stronger exact recovery result, we require the following assumption about the parameters of the~model.\

\begin{assumption}\label{assump:rate}
We assume $p=\O{1}, k=\O{1}$, i.e., the within-cluster connectivity and the size of the target cluster do not scale with the size of the graph $n$.\ Also, we assume $q=\frac{c}{n}$ for a fixed numerical constant $c>0$.\
\end{assumption}

\noindent
As we noted previously, the setting $k=\O{1}$ is often the case of interest for local graph clustering, where we would like to identify small- and medium-scale structure in large graphs \citep{leskovec2009community,leskovec2010empirical}.\ 
In this case, Assumption \ref{assump:rate} requires $p=\O{1}$, so that the  underlying ``signal'' of the problem does not vanish as the size of the graph grows, $n\to \infty$.\ 
As discussed earlier, this means $q$ must also scale as $\O{n^{-1}}$ for the local clustering algorithm to find the target without making many false positives.



Now we turn to the statement of exact recovery guarantees for $\ell_1$-regularized PageRank when applied to the noisy graph generated from Definition \ref{def:model}.\ 
In particular, the fact that $q=\O{n^{-1}}$ from Assumption \ref{assump:rate} allows that with nonvanishing probability there is a  node in the target cluster that is solely connected to $K$.\ 
This node will serve as a ``good'' seed node input in the $\ell_1$-regularized PageRank.\ 
With this choice of seed node, we now give  conditions under which the optimal solution $\xh$ has no false positives with nonvanishing probability.\ 

\begin{theorem}[\textbf{No false positives.}]
\label{thm:no_false_pos}
Suppose the same conditions as in Theorem \ref{thm:full_recovery}. Assume also that Assumption \ref{assump:rate} holds and that the size of the target cluster $k$ is $\geq 2(c+3)$.
Fix $\delta\leq 0.1$. 
If the regularization parameter satisfies $\rho\geq \rho(\delta)$, then  there is a good starting node in $K$ such that the solution to Problem~\eqref{eqn:pagerank2} with that node as a seed node and with teleportation parameter $\alpha \in [0.1,0.9]$ satisfies
\[ \supp(\xh) \subseteq K,\]
with probability at least $1 - 6\exp(-\O{\delta^2 p^2 k})- (1-\exp(-1.5c))^k -\O{n^{-1}}$,\footnote{More precisely, we assume $(1-\delta)p^2 k\geq c_0^{-1}\delta^{-2}\log k$ for a fixed constant $c_0>0$. Then with probability at least $1-6e^{-c_0 \delta^2(1-\delta) p k} -(1-\exp(-1.5c))^k -\O{n^{-1}}$, the statement in the theorem holds.} as long as
\begin{equation}
\label{eqn:degree_cond}
\frac{C(0.5c+1)}{\gamma p} = \O{\frac{1}{\gamma p}}< d_j,
\end{equation}
for all node $j\in K^c$ adjacent to $K$, where $C>0$ is a universal constant.\
\end{theorem}

\noindent
We require both  $\alpha\in [0.1,0.9]$ and $\delta\leq 0.1$ in the theorem to avoid overly complicated constants; while this simplifies the presentation of the theorem, it is not difficult to show that a similar result holds more generally.\
Importantly,  when combined with Theorem \ref{thm:full_recovery} (full recovery of the cluster), our result Theorem \ref{thm:no_false_pos} immediately establishes that $\ell_1$-regularized PageRank recovers the target cluster exactly, even when the target cluster is constant-sized.\  
We state this result informally in the following, which requires no proof.
\begin{corollary}[\textbf{Exact recovery; informal statement.}]
\label{cor:exact_recovery}
Under the same assumptions as in Theorem \ref{thm:full_recovery} and Theorem \ref{thm:no_false_pos}, there is a good starting node in $K$ such that $\ell_1$-regularized PageRank parameterized with that node as a seed node satisfies 
\[ \supp(\xh) = K,\]
with nonvanishing probability.
\end{corollary}

\noindent 
It should be noted that a sort of condition like~\eqref{eqn:degree_cond} about the realized degree seems necessary in order that the $\ell_1$-regularized PageRank has no false positives.\ 
The optimization program~\eqref{eqn:pagerank2} assigns less weights to low degree nodes in the $\ell_1$ penalty, so any nodes adjacent to $K$ will become active unless the $\ell_1$-regularized PageRank penalizes them with nontrivial weights.\ Unlike Theorem \ref{thm:full_recovery} and Theorem \ref{thm:bound_FP},  condition~\eqref{eqn:degree_cond} rules out some specific models to which Theorem \ref{thm:no_false_pos} can be applied.\ 
For example, planted clustering model with $p=\O{1}$ and $q=\O{1/n}$ does not satisfy this condition because the degrees in $K^c$ do not concentrate.\ For the stochastic block model, this condition is still satisfied if nodes adjacent to the target cluster belong to the clusters with degree larger than $\O{1/\gamma p}=\O{1}$.\ 
In practice,  condition~\eqref{eqn:degree_cond} may not be always applicable for every node adjacent to $K$, in which case the nodes that violate this condition may enter the model as false positives.\ We require the condition here though, since our model is essentially local and we do not have control outside $K$ beyond its~neighbors.\

\subsection{Comparison with existing  results}\label{sec:comparison}
The local graph clustering problem has been relatively well-studied in the area of theoretical computer science, and the existing works largely focus on the worst-case guarantees. 
We now compare our results through the random graph model with the  current known state-of-the-art worst-case results, given by~\citet{zhu2013local}. 

First, the main result Theorem 1 of \citet{zhu2013local}, when applied to our  population graph $\overline{G}$, implies that $\vol(\text{FP}) ,\vol(\text{FN})\leq \vol(K)\cdot \O{(1-\gamma)\log k}$, as long as $\text{Gap}=\O{1/((1-\gamma)\log k) } \geq \O{1}$. When $\gamma=\O{1}\in (0,1)$, our Theorem \ref{thm:full_recovery} states that if $pk^2 \geq \O{\log k}$, the output of the $\ell_1$-regularized PageRank model does not contain any false negative.  This cannot be deduced from \citet{zhu2013local}.\
In addition, our general bound on false positive, i.e., $\vol(\text{FP})\leq \vol(K)\cdot( \O{1/\gamma^2} - 1)$ in Theorem \ref{thm:bound_FP}, is better than the worst-case bound of \citet{zhu2013local} in the regime of large $\gamma$, which is typically the case for  many interesting  scenarios.  For instance, when the expected target conductance $\overline{\text{Cond}}=1-\gamma$ is small and fixed, the bound of the worst-case result degrades as the size of the target cluster $k$ increases, whereas our result is improved by increasing the probability bound. In the regime of $p=\O{1}, q=\O{1/n}$, and $k=\O{1}$ (hence $\gamma=\O{1}$), our Theorem \ref{thm:no_false_pos} shows that the output even contains no false positive. In this particular case, the strong separability ($p=\O{1}, q=\O{1/n}$) corresponds to a constant signal-to-noise ratio, since even for $q=\O{1/n}$ there are still a constant amount of edges outgoing from the target cluster, while the internal edges inside the target cluster is also constant.\ 
Although in practice the exact recovery of the target cluster may be a strong requirement, nevertheless, for real world clusters with high signal-to-noise ratio, $\ell_1$-regularized PageRank can still reconstruct the ground truth clusters more or less exactly (see, for instance, Section~\ref{sec:real_data}). 

Finally we briefly give a comparison of our theoretical results with the information theoretic results of~\citet{chen2016statistical} in the special case of planted clustering model. 
To ease and simplify the comparison, we only consider the case where $p=O(1)$ and $q=O(\log n/n)$. In this case, \citet[Theorem 2.5]{chen2016statistical}   implies that  a solution to a SDP achieves exact recovery  as long as $k\geq O(\log n)$, whereas our Theorem \ref{thm:full_recovery} and \ref{thm:bound_FP} suggest that in the same regime $\ell_1$-regularized PageRank fully recovers the target cluster while picking up a constant proportion of false positives. 
Importantly, $\ell_1$-regularized PageRank is essentially a local method, while \citet{chen2016statistical}'s SDP is a global method that explores the entire graph.

\subsection{Approximate personalized PageRank on random graph}\label{sec:appr_random}

One  advantage of variational formulation of local graph clustering, via $\ell_1$-regularized convex program~\eqref{eqn:pagerank2}, is that it allows tractable analysis of the method in random graphs. 
Given the connection between $\ell_1$-regularized PageRank and APPR established in Theorem~\ref{thm:appr_vs_l1}, 
this further allows us to obtain the statistical guarantees of the APPR algorithm under the random graph model. 
We formally state this result in the following theorem, which is a simple consequence of Theorem~\ref{thm:appr_vs_l1}, in combination with those  obtained in Section~\ref{sec:full_recovery} and~\ref{sec:no_false_pos}.
 (We write $\text{FP}(x^{\text{APPR}})=\{i\in \supp(x^{\text{APPR}}): i\in K^c \}$ to denote the collection of false positive nodes in the APPR's output.)
\begin{theorem}[\textbf{Recovery guarantees for APPR.}]\label{thm:appr_thm}
Consider the APPR algorithm given in~\eqref{eqn:appr} with parameter $\rho=\rho(\delta)$.
Under the same conditions as Theorem \ref{thm:full_recovery}, the output vector $x^{\text{APPR}}(\rho(\delta))$ satisfies 
\[K\subseteq \supp(x^{\text{APPR}}) \;\;\text{ and }\;\; \vol(\text{FP}(x^{\text{APPR}})) \leq \vol(K)\left[\O{\frac{1}{\gamma^2}} - 1 \right], \]
with nonvanishing probability. Furthermore, under the same conditions as Theorem \ref{thm:full_recovery}, there is a good starting node in $K$ such that the APPR's output vector parameterized with that node as a seed node satisfies 
\[ \supp(x^{\text{APPR}}) = K,\]
with nonvanishing probability, as long as
\begin{equation*}
 \O{\frac{1}{\gamma p}}< d_j,
\end{equation*}
for all node $j\in K^c$ adjacent to $K$.
\end{theorem}

\section{Stagewise PageRank and solution paths}\label{sec:stagewise}

The $\ell_1$-regularized PageRank problem~\eqref{eqn:pagerank2} is convex and there are numerous ways to solve it using convex optimization techniques.\
In~\citet{fountoulakis2019variational}, the authors apply proximal gradient descent (ISTA) 
and show that the algorithm  enjoys  the {\em locality} property, meaning that the algorithm touches at most the optimal support set and its neighbors, while inheriting the fast convergence property of the proximal gradient descent.\
Optimization algorithms typically solve the problem at a single value of $\rho$, or a sequence of multiple values. On the other hand, ``path algorithms'' are designed to solve the problem for all values of a regularization parameter $\rho\in (0,\infty]$, or a subset of it when terminated early.  This is more desirable when we need  solutions over a list of parameter~values.\

The idea of designing path algorithms has already gained much attention in the sparse regression literature, first pioneered by~\citet{efron2004least} and further developed by~\citet{zou2007degrees,hastie2004entire,arnold2016efficient}.  This has rendered the exploration of full regression coefficient paths relatively inexpensive.\
Unlike regression setting, however, this type of path algorithm has been less studied in local graph clustering~\citep{gleich2016seeded}.\
Motivated by the path algorithms developed in the sparse regression setting, we consider the following coordinate-wise algorithm for local graph clustering: 
 \begin{itemize}
 \item Initialize $x^{(0)}=0$;
 \item For each $k\geq0$, iterate the steps
 \begin{equation}\label{eqn:stagewise}
\begin{cases}
\text{Choose an $i\in [n]$ such that $d_i^{-1}\nabla_i \fp (x^{(t)}) \leq 0$ is the smallest among $[n]$};\\
\text{Update $x^{(k+1)}_i=x^{(k)}_i + d_i^{-1}\eta$. }
\end{cases}
\end{equation}
\item Output a sequence $\{x^{(0)},x^{(1)},\ldots\}$.
 \end{itemize}
The two main features of this algorithm are: 1) we greedily select the coordinate $i$ at each iteration that maximizes the magnitude of gradient, and 2) we update the current iterate by adding a small step size $\eta$ to the $i$th coordinate.\ This conservative update of the variable counterbalances the greedy selection step, thus making the algorithm more stable.\


The above algorithm is called the ``stagewise'' algorithm, and in fact it has been widely studied by many authors~\citep{efron2004least,hastie2007forward,rosset2004boosting,rosset2007piecewise,zhao2007stagewise,tibshirani2015general}.\
The stagewise algorithm is known to have implicit regularization effect closely related to $\ell_1$ norm regularization~\citep{tibshirani2015general}, and in particular, if each component of the $\ell_1$-regularized solution has a monotone path, then the stagewise path exactly coincides with that of $\ell_1$ regularization, for $\eta\to 0$~\citep{efron2004least,rosset2004boosting}.\ 
Recalling our earlier result on the monotone path property of $\ell_1$-regularized PageRank (Lemma~\ref{lem:monotone}), the following result is then immediate.
\begin{corollary}\label{cor:stagewise_equivalence}
The output sequence of stagewise algorithm described in~\eqref{eqn:stagewise} converges to the $\ell_1$-regularized PageRank solution path as the step size goes to $0$, i.e., $\eta\to 0$.
\end{corollary}
Therefore, the stagewise algorithm allows us to {\em provably} explore the entire $\ell_1$-regularization path via a single run of  simple iterative steps.\
Another advantage of the stagewise algorithm is that it enjoys the locality property, in that the algorithm only touches the chosen nodes and its neighbors as it progresses.\
This is obvious from the update step of~\eqref{eqn:stagewise} and the expression of the gradient $\nabla_i \fp(x)$.\
Thus, when terminated early, the algorithm produces an approximate and partial solution path without making access to the entire graph.\

For the $\ell_1$-regularized PageRank method, the  parameter $\rho$  controls the extent to which the random walk has moved farther from the seed, so different values of $\rho$  reveal various scales of local clustering structure around the seed node.\ Therefore, in the setting of local graph clustering, the stagewise algorithm allows us to provably and efficiently  track the evolution of a $\ell_1$-regularized PageRank diffusion and better understand the local cluster properties of the graph.\ 
This is well-suited for the purpose of exploratory graph analysis, and the idea of using path algorithms for exploring the graph has been also studied in~\citet{gleich2016seeded}.\ 
In addition to the exploratory analysis, the stagewise algorithm can be a competitive algorithm to find the target cluster if  the size of the target cluster is small and/or medium and one needs a fine scale resolution of the solution path.\ 
However, when the  size of the target cluster is quite large, using optimization algorithms with a coarse grid of regularization parameter may lead to better computational savings without exploring the entire solution path  from scratch.\ 
Overall, the stagewise algorithm must be used in a complementary way to the optimization algorithms that directly solve~\eqref{eqn:pagerank2}. We also refer the readers to~\citet{tibshirani2015general} for comprehensive study of  the stagewise algorithm for general sparse modeling~problem.

\begin{figure}
\centering     
\subfigure[$\ell_1$-reg. path]{\label{fig:solution_path_a}\includegraphics[scale=0.35]{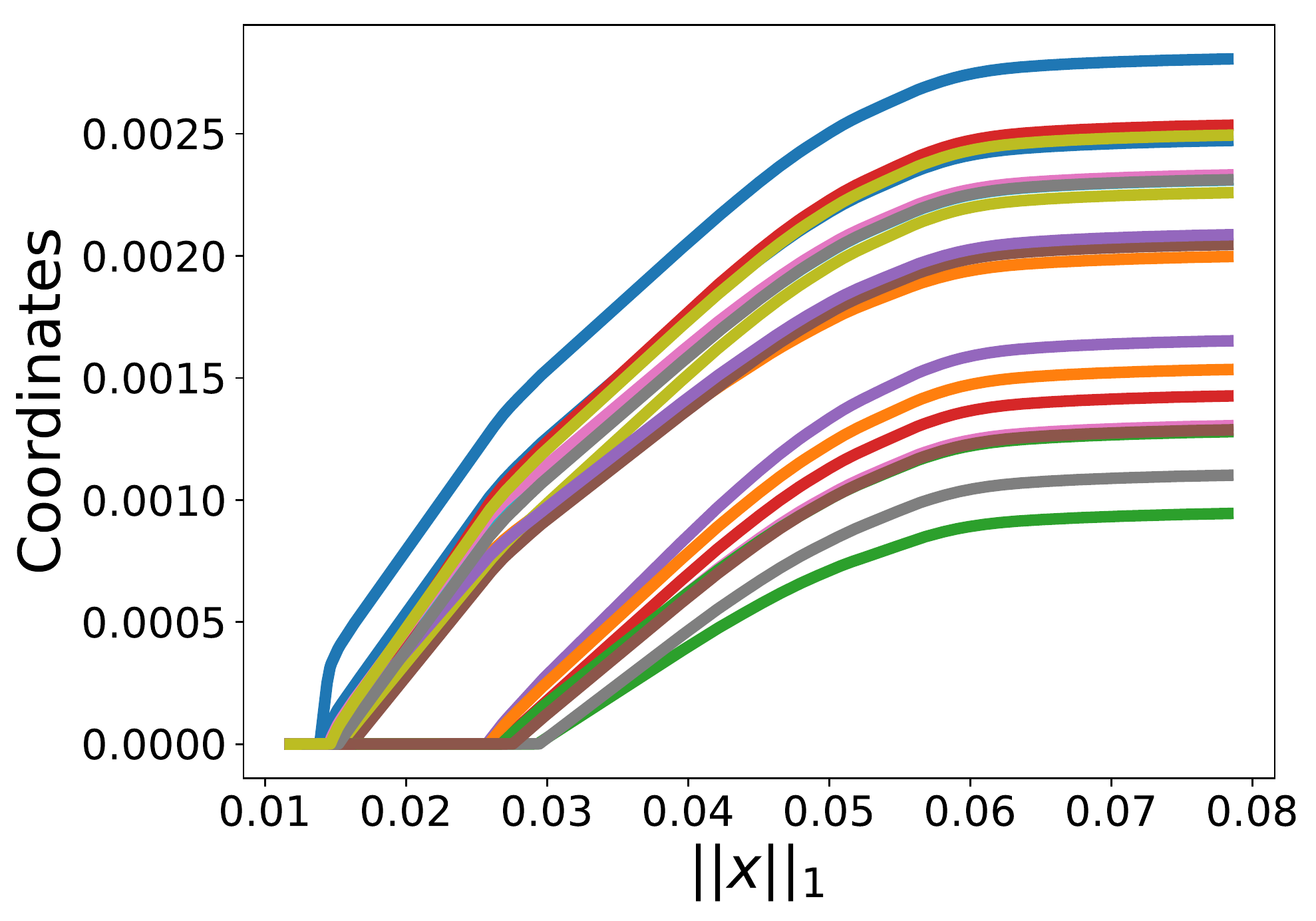}}
\subfigure[Stage. path $\eta=10^{-4}$]{\label{fig:solution_path_b}\includegraphics[scale=0.35]{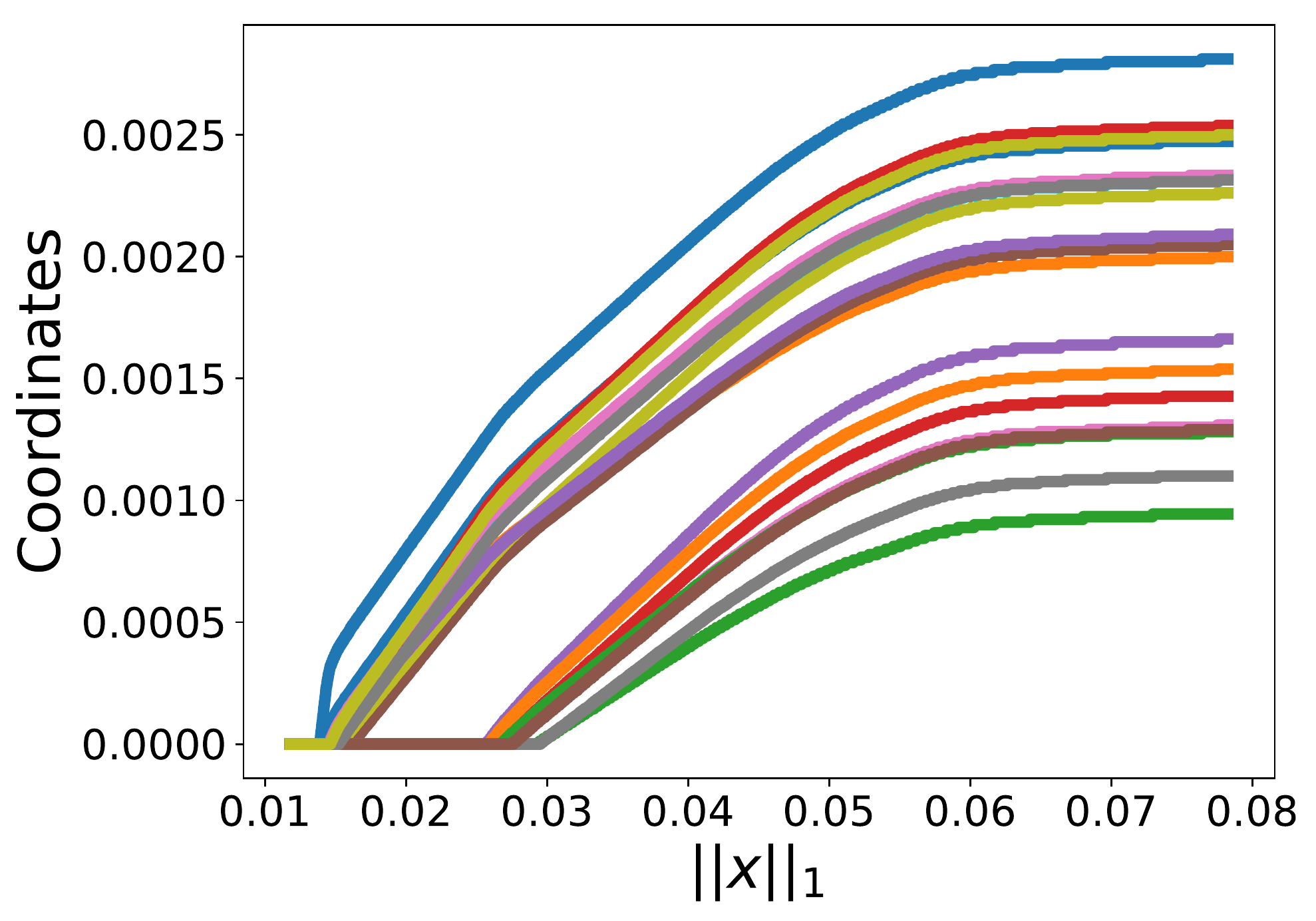}}
\subfigure[Stage. path  $\eta=5\cdot  10^{-3}$]{\label{fig:solution_path_c}\includegraphics[scale=0.35]{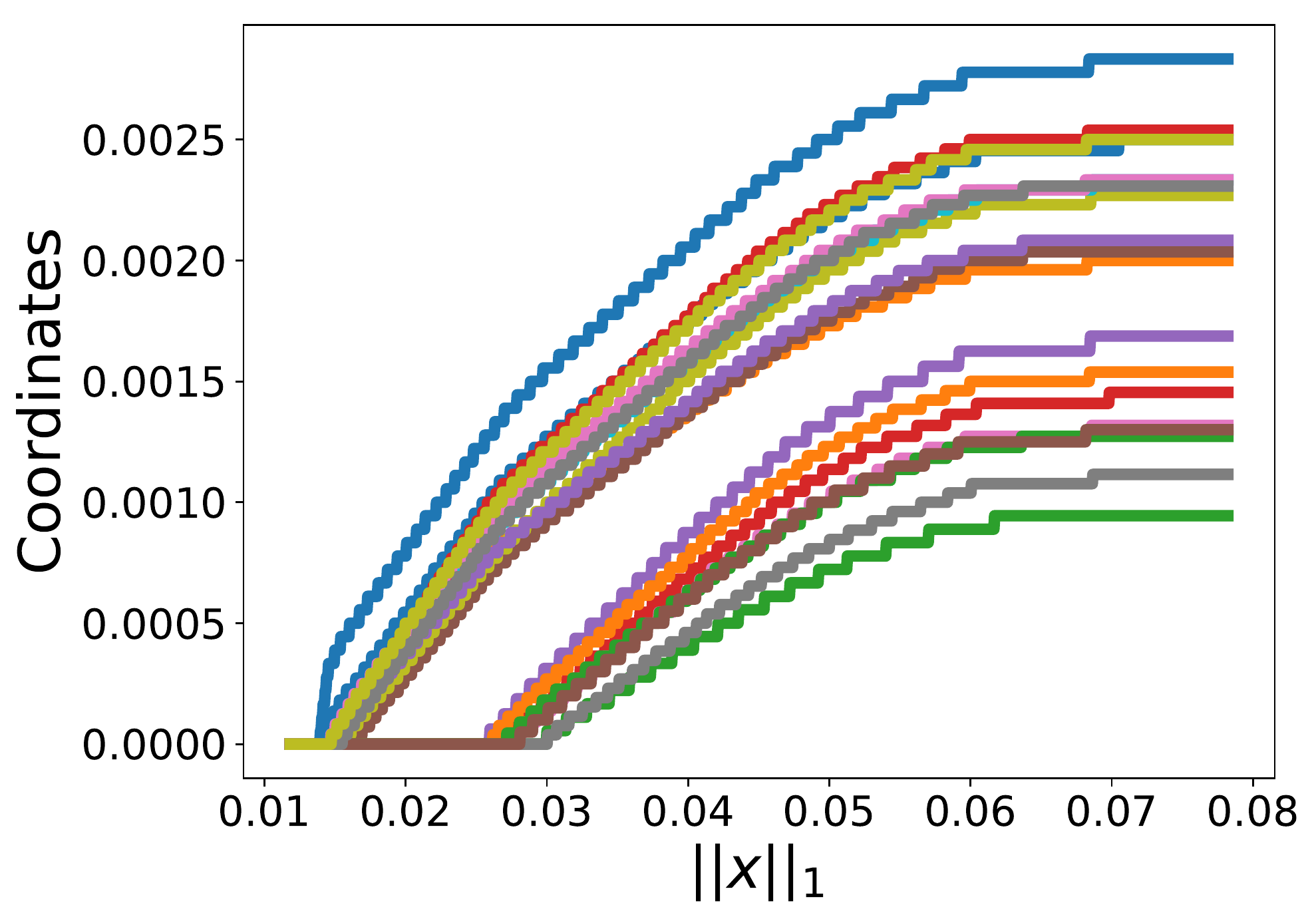}}
\subfigure[Stage. path  $\eta=10^{-3}$]{\label{fig:solution_path_d}\includegraphics[scale=0.35]{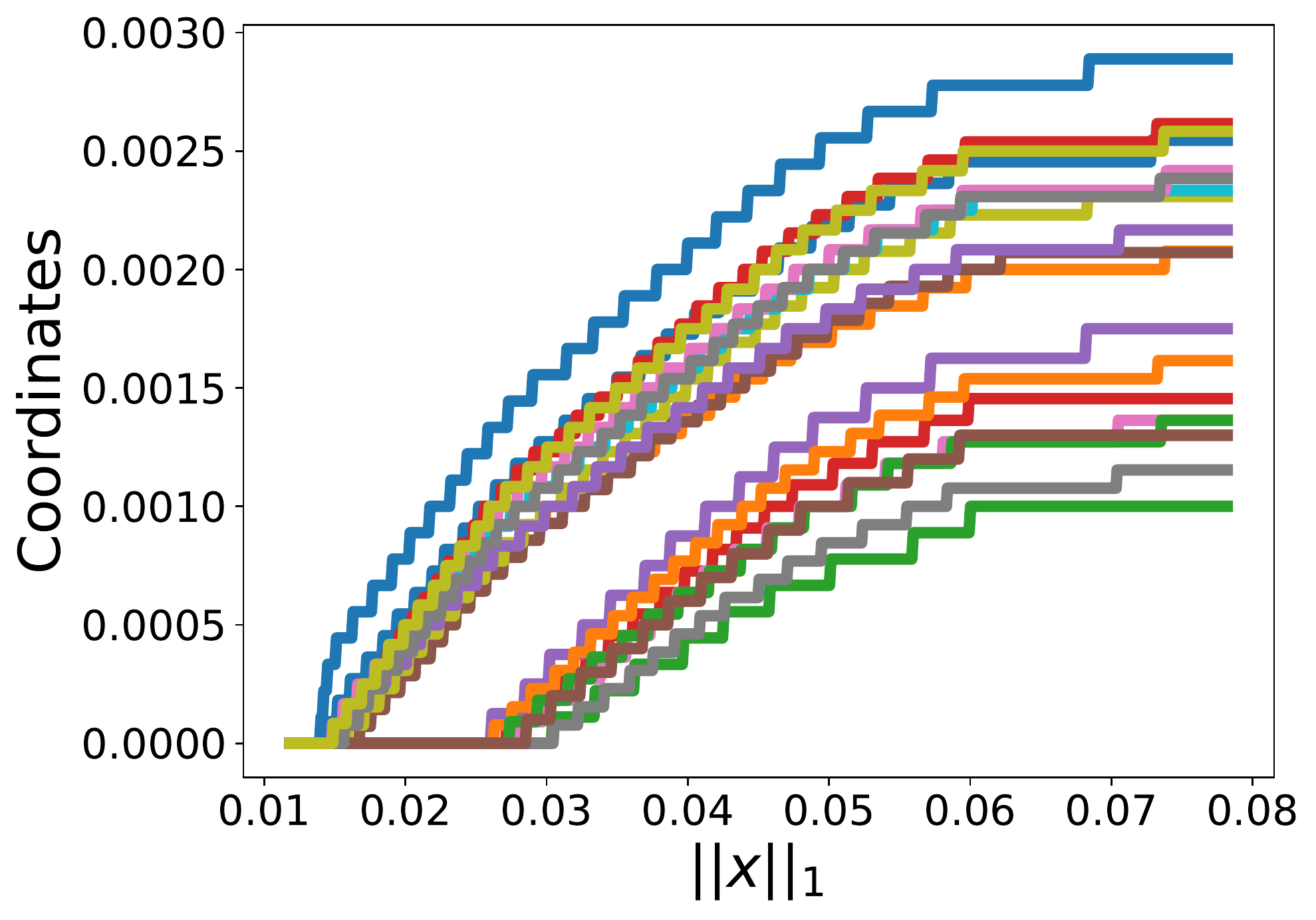}}
\caption{Comparison of $\ell_1$ regularization path and stagewise paths for different step sizes $\eta$.\ The profiles are  shown only for  nodes in $K\backslash S$ among $n=1000$ nodes. Each color in the plot corresponds to different  nodes. The $x$-axis is the $\ell_1$ norm of the current estimates.\ For stagewise, the results are obtained with $7706, 1527$, and $764$ iterations respectively.}
\label{fig:solution_path}
\end{figure}

In Figure \ref{fig:solution_path}, we compare the actual solution path to the stagewise algorithm paths for different step sizes.\ We generate  data from the stochastic block model with $p=0.5, q=0.002$ and $r=50$. Each cluster has $20$ nodes. Figure~\ref{fig:solution_path} shows the $\ell_1$-regularization path and stagewise component paths for one particular draw from the stochastic block model.\ Here we  only show the solution paths for nodes in the target cluster without seed node, among $n=1000$ nodes (each color line corresponds to different nodes in the target cluster).\ Note that when the step size is small, $\eta=0.0001$, the stagewise path appears to closely match to the $\ell_1$-regularization path; for moderate step size, $\eta=0.0005$, the stagewise path exhibits some jagged pattern but nevertheless accurately approximates the optimal path; and for relatively large step size, $\eta=0.001$, while the jagged pattern becomes more evident visually, it is clear that the overall trend still coincides well with that of the $\ell_1$ solution path.\ 


\section{Numerical evaluation}
\label{sec:experiments}

In this section, we provide a detailed numerical evaluation, to illustrate the performance of $\ell_1$-regularized PageRank on synthetic and real data.\ 
We have conducted a comprehensive experiments to demonstrate the state-of-the-art performance of the method across a wide range of real graphs, which has not been investigated at this level of extent in the prior works.
To measure the quality of the recovered cluster, we define Precision and Recall as $\vol(\mbox{TP})/ \vol(\supp(\xh))$ and $\vol(\mbox{TP})/ \vol(K)$, respectively.
The F1score is the harmonic mean of precision and recall, $\left(2/\left(\text{Precision}^{-1}+\text{Recall}^{-1}\right)\right)$.
We will also make use of conductance, where recall $\text{Cond}=\mbox{Cut}(S,S^c)/\min\left(\vol(S), \vol(S^c)\right)$.\
The lower the conductance value is, the better  the quality of the cluster is.\
For our experiments, we solve problem~\eqref{eqn:pagerank2} using a proximal coordinate descent algorithm, which enjoys both the ``strongly local'' running time property (running time depends on the size of cluster rather than the entire graph) as well as linear convergence~\citep{fountoulakis2019variational}.\

\subsection{Simulated data}\label{sec:simulated_data}

\begin{figure}
\centering     
\subfigure[F1score against $\gamma$]{\label{fig:f1vsgamma_a}\includegraphics[scale=0.7]{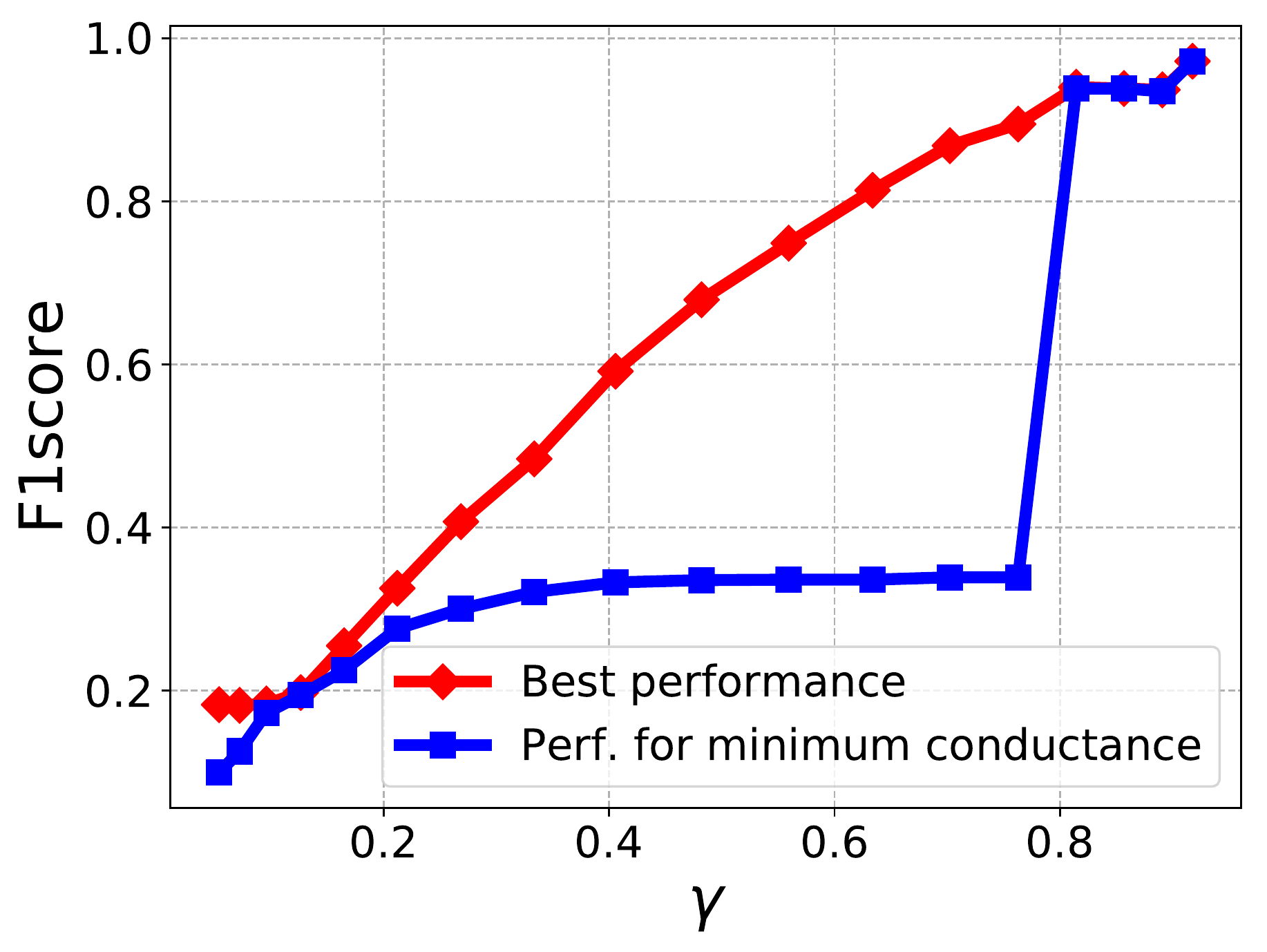}}
\caption{Performance (F1score) of $\ell_1$-regularized PageRank as $\gamma$ increases. We demonstrate two scenarios. The first scenario (red line with diamonds, ``Best performance'') shows performance in the case that we could select the regularization $\rho$ that corresponds to the largest F1score. We obtain this result by using the stage-wise algorithm that generates the whole solution path for any $\rho$, and we select the solution with the largest F1score. The second scenario (blue line with squares, ``Perf. for minimum conductance'') shows performance in case we select the solution with minimum conductance out of all solutions produced by the stage-wise algorithm. For larger values of $\gamma$, the two scenarios follow each other quite closely, but for smaller values there can be a large difference.}
\label{fig:f1vsgamma}
\end{figure}

In Figure \ref{fig:f1vsgamma}, we demonstrate the performance of $\ell_1$-regularized PageRank as $\gamma$ increases (where, recall, $\gamma$ is defined in Eqn.~(\ref{eqn:gamma})). 
We illustrate the performance of $\ell_1$-regularized PageRank in two cases. 
The first (red line with diamonds, ``Best performance'') is a best-case scenario where we select the solution with the best F1score out of all the solutions that are produced by the stage-wise algorithm \eqref{eqn:stagewise}.
The second scenario (blue line with squares, ``Perf. for minimum conductance'') shows a more realistic case where we select the solution with minimum conductance out of all the solutions that are produced by the stage-wise algorithm. 
We note that F1score for the best-case scenario scales linearly as a function of $\gamma$. On the other hand, the scenario where we select the solution with minimum conductance scales sub-linearly as a function of $\gamma$ until a phase-transition around $\gamma\approx0.75$, after which the scenario of minimum conductance matches the best-case scenario. The gap between the best-case scenario and the minimum conductance scenario indicates the need for new post-processing methods beyond minimum conductance, which we leave for future work.
For the experiments in Figure \ref{fig:f1vsgamma}, we fix the teleportation parameter $\alpha=0.1$. We generate graphs from the stochastic block model which consists of $10$ clusters, each of which has $20$ nodes, and only one of which is the target cluster $K$.\ 
We use the same parameters $p$ and $q$ across different clusters to generate edges within and between clusters.\ 
Here we set $p=0.5$ and $q$ is varying in order to generate various $\gamma$ as is shown in Figure~\ref{fig:f1vsgamma}.\ 
The results are averaged over $30$ trials.\

\begin{figure}
\centering     
\subfigure[$\gamma = 0.91$]{\label{fig:FR_TP_a}\includegraphics[scale=0.40]{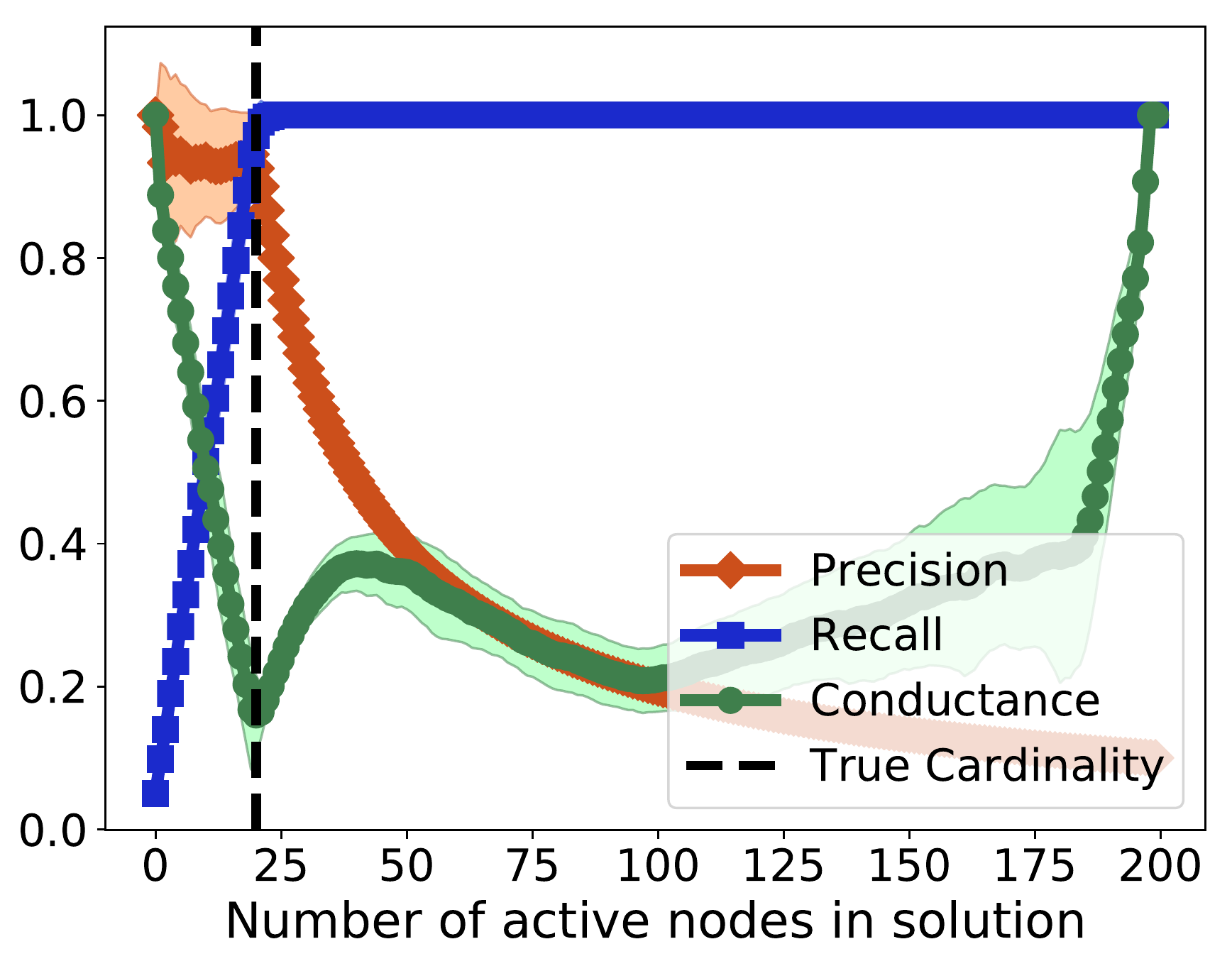}}
\subfigure[$\gamma = 0.86$]{\label{fig:FR_TP_b}\includegraphics[scale=0.40]{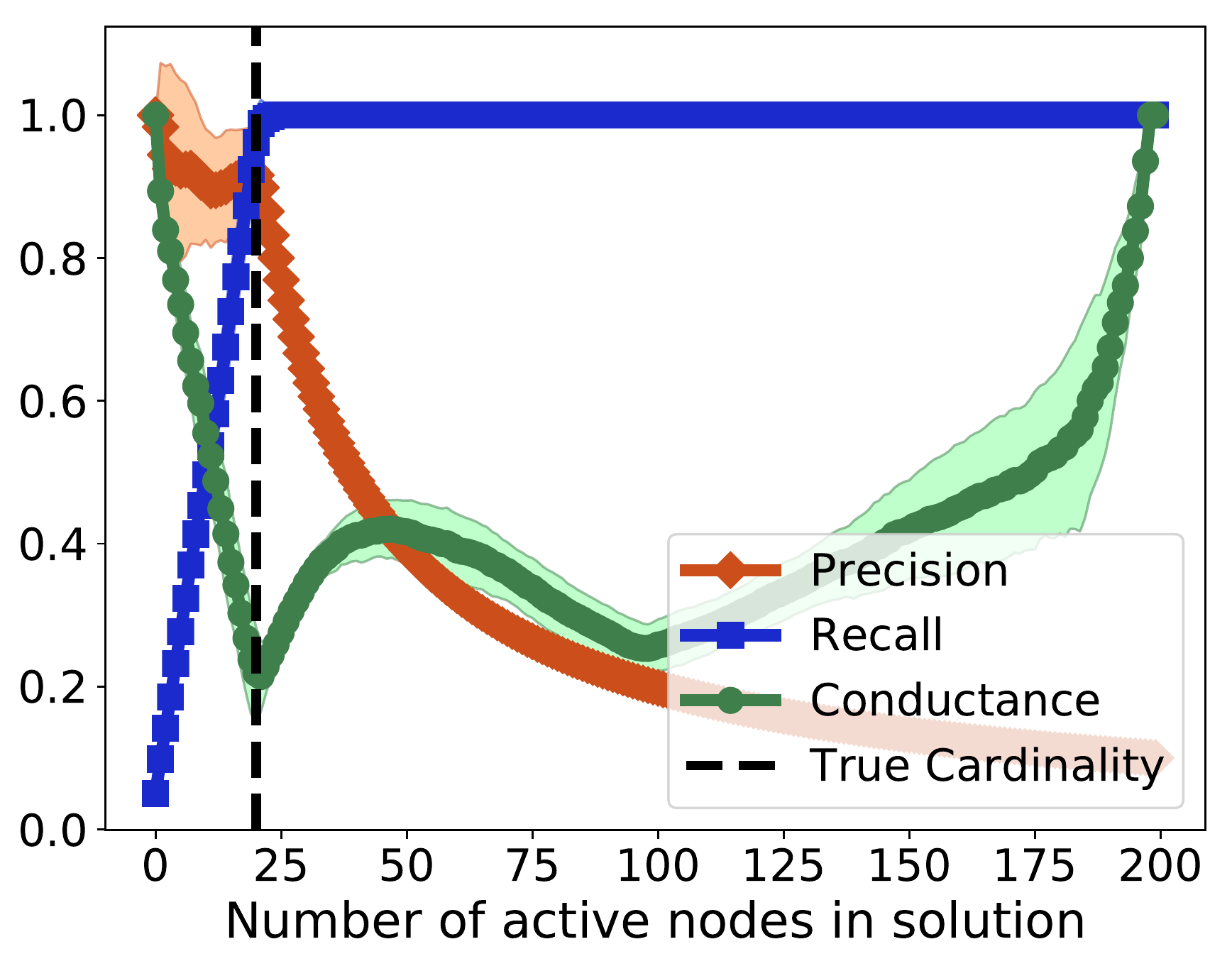}}
\subfigure[$\gamma = 0.65$]{\label{fig:FR_TP_c}\includegraphics[scale=0.40]{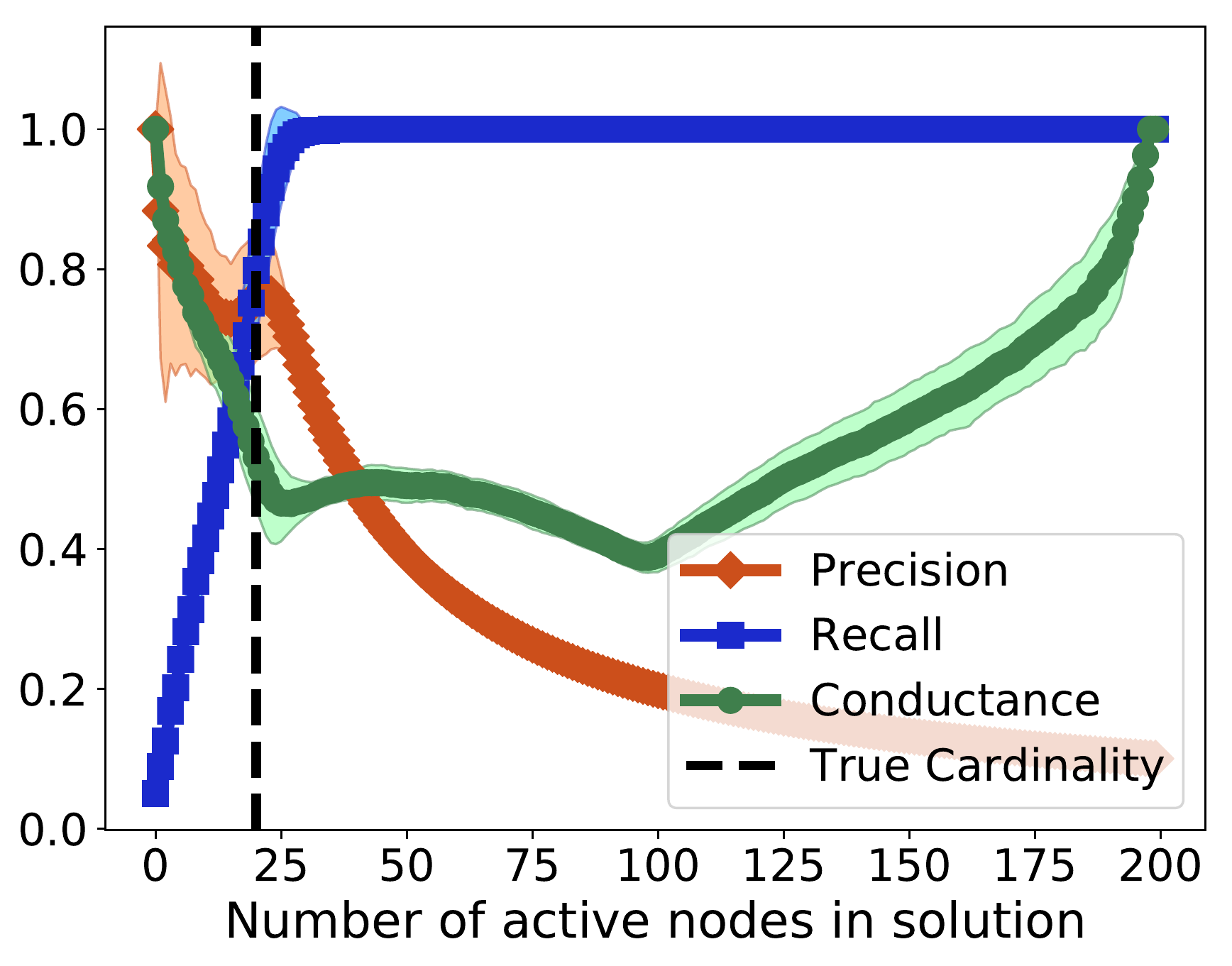}}
\subfigure[$\gamma = 0.44$]{\label{fig:FR_TP_d}\includegraphics[scale=0.40]{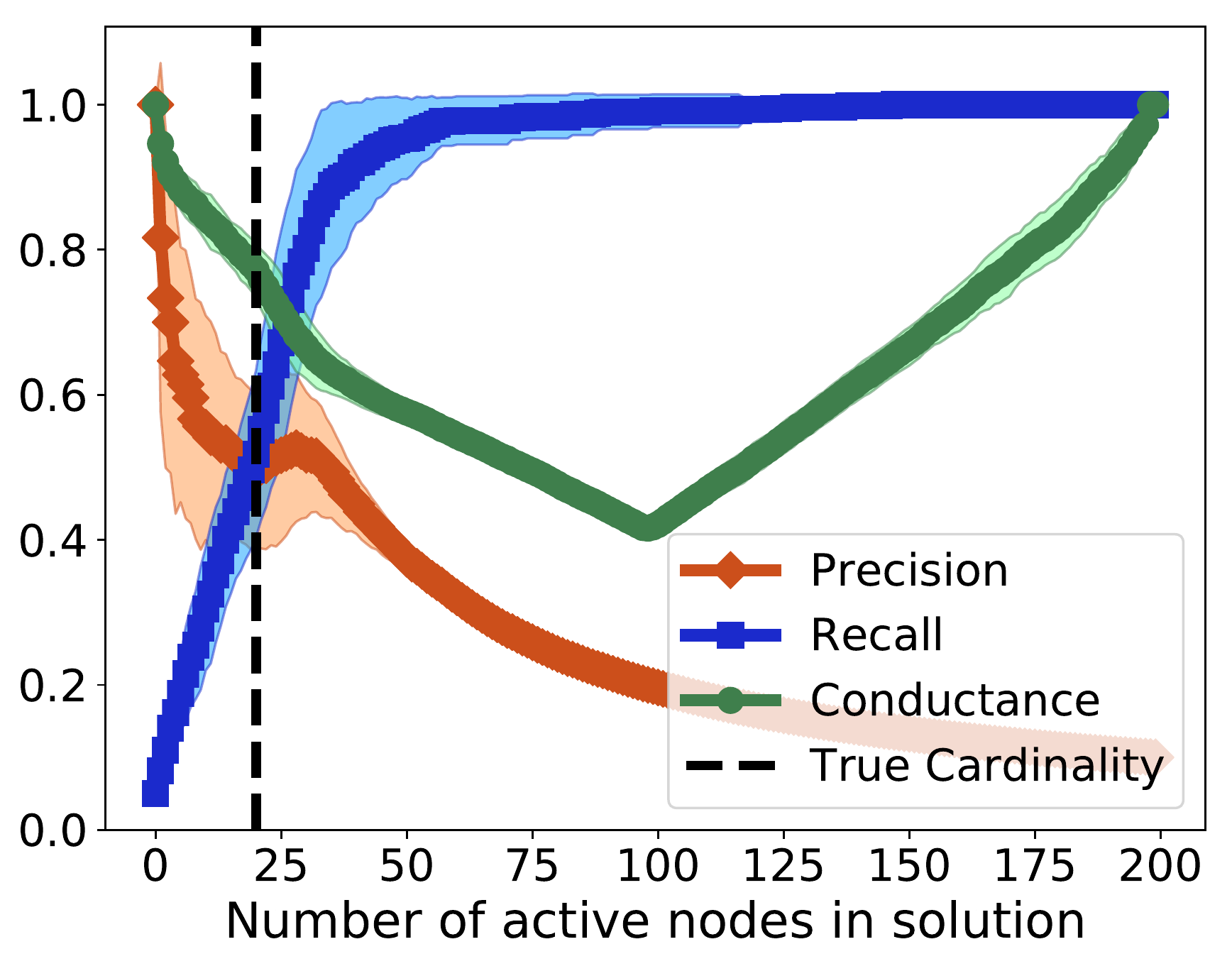}}
\caption{In Figures \ref{fig:FR_TP_a}, \ref{fig:FR_TP_b} and \ref{fig:FR_TP_c}, we illustrate that for large and medium values of $\gamma$, $\ell_1$-regularized PageRank recovers the target cluster. 
On the other hand, for smaller $\gamma$, in Figure \ref{fig:FR_TP_d}, $\ell_1$-regularized PageRank does not recover the target cluster with high accuracy. The $x$-axis gives the number of nonzero nodes in the solution of $\ell_1$-regularized PageRank as the parameter $\rho$ decreases. The vertical line indicates the  cardinality of the target cluster ($k=20$).}
\label{fig:FR_TP}
\end{figure}

In Figure \ref{fig:FR_TP}, we demonstrate a much more detailed view of the performance of $\ell_1$-regularized PageRank for four representative values of $\gamma$.
More precisely, we show that for large and medium values of $\gamma$, there exists a parameter $\rho$ such that the support of the $\ell_1$-regularized PageRank solution recovers the target cluster; 
and, for smaller $\gamma < 0.5$, $\ell_1$-regularized PageRank does not recover the target cluster with high accuracy. For Figure \ref{fig:FR_TP} we use the same experiment setting as in Figure \ref{fig:f1vsgamma}.
For large $\gamma$, Figure \ref{fig:FR_TP_a}, we observe that when $\ell_1$-regularized PageRank recovers about $20$ nodes, then these nodes correspond to very high precision and recall, and as the number of nodes in the solution increases then precision decreases. We also observe that conductance of the recovered cluster is a good metric for finding the target cluster.\ By this, we mean that we will find the target cluster with high precision and recall if out of all solutions on the path we choose the one with minimum conductance.\
As $\gamma$ gets smaller, e.g., in Figure~\ref{fig:FR_TP_c}, the minimum conductance does not relate to the target cluster.\ However, it is clear from Figures \ref{fig:FR_TP_b} and \ref{fig:FR_TP_c} that $\ell_1$-regularized PageRank with minimum conductance still finds the target cluster $K$ with good accuracy if the algorithm is terminated early.\
Finally, in Figure \ref{fig:FR_TP_d}, we demonstrate a case where $\gamma$ is small and conductance of solution fails to relate to the target cluster and  there is no output of $\ell_1$-regularized PageRank that recovers the target cluster accurately.

\subsection{Real data experiments}\label{sec:real_data}

Next, we test the performance of local graph clustering methods using biology networks and social networks. All the graphs that we consider are unweighted and undirected.
For a summary of the datasets see Table \ref{tab:sumdata}. All datasets that are used come with \textit{suggested} ground-truth clusters for social networks and with a gold standard for biology networks.
However, we filter the given ground-truth clusters by measuring their conductance values. In particular, we keep only the ground-truth clusters that have conductance value less than or equal to $0.6$.
This means we keep the clusters that the number of edges that cross the cluster is $60\%$ of the volume of the cluster.
This way we obtain a wide range of clusters from ``good'' (small conductance) to noisy (large conductance, i.e., close to~$0.6$).

\begin{table}[h] 
    \caption{Summary of datasets}
\vspace{2mm}
  \centering
  \begin{adjustbox}{max width=\textwidth}
    \begin{tabular}{cccl}
    dataset & number of nodes & number of edges &  \multicolumn{1}{c}{\rotatebox[origin=c]{0}{description}}\\
    \midrule
    Sfld & $232$ & $15570$ & pairwise similarities of blasted sequences of proteins\\    
     PPI-mips & $1096$ & $13221$ & protein-protein interaction network\\  
     FB-Johns55 & $5157$ & $186572$ & Facebook social network for Johns Hopkins University\\   
     Colgate88 & $3482$ & $155043$ & Facebook social network for Colgate University\\   
     Orkut & $3072441$ & $117185083$ & Large-scale on-line social network\\   
    \midrule
    \end{tabular}%
    \end{adjustbox}
  \label{tab:sumdata}%
\end{table}%

%

\subsubsection{Datasets}

\textbf{Sfld}.\ This dataset contains pairwise similarities of blasted sequences of $232$ proteins belonging to the amidohydrolase superfamily~\citep{brown2006gold}. There are $232$ nodes and $31140$ edges in this graph. 
A gold standard is provided describing families within the given superfamily. According to the gold standard the amidrohydrolase superfamily contains $29$ families/clusters. However, after filtering the families we find that only two have conductance value less than $0.6$.
(This should not be surprising, given that the graph is so dense.)
In Table \ref{tab:clusters_biology} we present properties of the two clusters that correspond to urease.0 and AMP amidrohydrolase superfamilies.

\noindent \textbf{PPI-mips}.\ This dataset is a protein-protein interaction graph of mammalian species~\citep{pagel2004mips}. There are $1096$ nodes and $26442$ edges in this graph. In Table \ref{tab:clusters_biology} we provide all \textit{suggested} ground-truth clusters that have conductance less than $0.6$.

\begin{table}[h] 
    \caption{``Ground truth'' clusters for the Sfld and PPI-mips datasets. Full details about the datasets can be found in the original paper~\citep{pagel2004mips}. In this table we report the volume, the number of nodes and the conductance of each ground truth cluster.}
\vspace{2mm}
  \centering
    \begin{adjustbox}{max width=\textwidth}
    \begin{tabular}{cllccc}
    dataset &  \multicolumn{1}{c}{feature} & \multicolumn{1}{c}{feature (short name)} &volume & nodes & conductance\\
    \midrule
     \multirow{2}{*}{\rotatebox[origin=c]{0}{Sfld}}           & urease.0 & urease & $16209$  &$100$ & $0.42$\\
      								                  & AMP & AMP & $1721$  &$28$ & $0.56$\\\hdashline
     \multirow{19}{*}{\rotatebox[origin=c]{0}{PPI-mips}} & Actin-associated-proteins & Actin & 870 & 24 & 0.36 \\
                 								 & Anaphase-promoting-complex & Anaphase & 165 & 11  & 0.33 \\
                 								 & Cdc28p-complexes &Cdc28p & 135 & 10 & 0.33 \\
                 								 & Coat-complexes &Coat & 676 & 19 &0.49 \\
                 								 & cytoplasmic-ribosomal-large-subunit & ct-large & 9720 & 81 & 0.33\\
                 								 & cytoplasmic-ribosomal-small-subunit & ct-small & 4788 & 57 & 0.33\\
                 								 & F0-F1-ATP-synthase & F0-F1-ATP & 315 & 15 & 0.33\\
                 								 & H+-transporting-ATPase-vacuolar & H+& 315 & 15 & 0.33 \\
                 								 & mitochondrial-ribosomal-large-subunit & mc-large & 1488 & 32 & 0.33 \\
                 								 & mitochondrial-ribosomal-small-subunit & mc-small & 273 & 14 &0.33 \\
                 								 & Mitochondrial-translation-complexes &mc-complex & 281 & 12 &0.53 \\
                 								 & mRNA-splicing &mRNA & 1861 & 33 & 0.43\\
                 								 & Nuclear-pore-complex & Nuclear& 614 & 18 &0.50 \\
                 								 & RNA-polymerase-II-holoenzyme &RNA & 1487 & 29 &0.45 \\
                 								 & Spindle-pole-body &Spindle & 981 & 22  &0.52 \\
                 								 & TRAPP-complex & TRAPP& 135 & 10 & 0.33\\
                 								 & tRNA-splicing &tRNA & 165 & 11 &0.33 \\
                 								 & 19-22S-regulator &19-22S & 459 & 18 &0.33 \\
                 								 & 20S-proteasome & 20S & 315 & 15 &0.33 \\
    \midrule
    \end{tabular}%
    \end{adjustbox}
  \label{tab:clusters_biology}%
\end{table}%


\noindent \textbf{FB-Johns55}.\
This graph is a Facebook anonymized dataset on a particular day in September $2005$ for a student social network at John Hopkins university. The graph is unweighted and it represents ``friendship'' ties.
The data form a subset of the Facebook100 data-set from~\citet{traud2011comparing,traud2012social}.
This graph has $5157$ nodes and $186572$ edges. 
This dataset comes along with $6$ features, i.e., second major, high school, gender, dorm, major index and year. 
We construct ``ground truth'' clusters by using the features for each node.
In particular, we consider nodes with the same value of a feature to be a cluster, e.g., students of year $2009$. 
As analyzed in the original paper that introduced these datasets~\citep{traud2012social}, for FB-Johns55 the year and major index features give non-trivial ``assortativity coefficients'', which is a ``local measure of homophily''. This agrees with the ground truth clusters we find after applying our filtering technique. The clusters per graph are shown in Table \ref{tab:clusters}. 

\noindent \textbf{Colgate88}.\ This graph is constructed similarly to the FB-Johns55 graph but for Colgate University. We apply the same filtering techniques as for FB-Johns55 and we present the filtered clusters in Table \ref{tab:clusters}. There are $3482$ nodes and $155043$ edges in this graph. 

\begin{table}[h] 
    \caption{``Ground truth'' clusters for the FB-Johns55 and Colgate88 datasets.}
\vspace{2mm}
  \centering
    \begin{adjustbox}{max width=\textwidth}
    \begin{tabular}{ccccc}
    dataset & feature & volume & nodes  & conductance\\
    \midrule
     \multirow{9}{*}{\rotatebox[origin=c]{0}{FB-Johns55}} & year $2006$& $81893$ & $845$& $0.54$\\
										    & year $2007$& $89021$  & $842$ & $0.49$\\     										    
										    & year $2008$& $82934$  & $926$& $0.39$\\
     										    & year $2009$&  $33059$ & $910$& $0.21$\\
										    & major index $217$ & $10697$ & $201$ & $0.26$ \\
										    & second major $0$ & $178034$ & $2844$ & $0.51$\\
										    & dorm $0$ &$137166$ & $2121$ & $0.52$\\
							                             & gender $1$ & $181656$  &$2144$ & $0.46$\\
							                             & gender $2$ & $173524$  & $2598$ & $0.48$\\\hdashline
     \multirow{10}{*}{\rotatebox[origin=c]{0}{Colgate88}}    & year $2004$ & $14888$  &$230$ & 0.54\\
     										    & year $2005$ & $50643$  & $501$ & $0.50$\\
     										    & year $2006$ & $62065$  & $557$ & $0.48$\\
     										    & year $2007$ & $68382$  & $589$ & $0.41$\\
							                             & year $2008$ & $62430$  & $641$ & $0.29$\\
							                             & year $2009$ & $35379$  & $641$ & $0.11$\\
							                             & secondMajor $0$ & $175239$  &$2107$ & $0.54$\\
							                             & dorm $0$ & $100414$  & $1157$& $0.53$\\
							                             & gender $1$ & $162759$  & $1695$ & $0.48$\\
							                             & gender $2$ & $123724$  & $1485$& $0.55$\\
							                             
    \midrule
    \end{tabular}%
    \end{adjustbox}
  \label{tab:clusters}%
\end{table}%

\noindent \textbf{Orkut}.\ Orkut is a free on-line social network where users form friendship each other. Orkut also allows users form a group which other members can then join. This dataset has $3072441$ nodes and $117185083$ edges. It can be downloaded from~\citet{snapnets}.
This dataset comes with $5000$ ground truth communities, which we filter by maintaining the clusters with conductance less than or equal to $0.6$. Out of the $5000$ communities only $282$ pass our filtering test. 

The above real graphs may not be generated from the random model that we consider in Section~\ref{sec:statistical_guarantee}, nonetheless, we have evaluated the method to illustrate the scenarios that are not just idealized. 
Moreover, the Facebook social networks and the biology networks are expected to exhibit  homophily structure for some ground truth clusters that are highly inter-connected relative  to the rest of the graph. 
These block-structured networks belong to a special class of networks and we believe that random graph models, such as stochastic block model, can closely approximate it. 

\subsubsection{Methods}

We compare the performance of $\ell_1$-regularized PageRank with state-of-the-art local graph clustering algorithms. There are two categories of local graph clustering methods: local spectral \citep{andersen2006local} and it's follow-up work \citep{zhu2013local}; and local flow-based methods \citep{lang2004flow,andersen2008algorithm,orecchia2014flow,veldt2016simple}.

\textit{Approximate personalized PageRank (APPR):}\ In \citet{andersen2006local} and \citet{zhu2013local} an approximate personalized PageRank (APPR) algorithm is proposed, where the personalized PageRank linear system is solved approximately using a local diffusion process. Both papers study nearly identical algorithms, with the latter paper giving better theoretical guarantees based on a stronger assumption. In particular, the authors in \citet{zhu2013local} assume that the internal connectivity of the target cluster (minimum conductance of the induced subgraph for the target cluster) is quadratically stronger than the conductance of the target cluster. 


\textit{SimpleLocal (SL):}\ Out of the four flow-based methods \citet{lang2004flow,andersen2008algorithm,orecchia2014flow,veldt2016simple}, the one proposed in \citet{veldt2016simple} (SimpleLocal) simplifies and generalizes the methods proposed in \citet{lang2004flow,andersen2008algorithm,orecchia2014flow}, while having similar theoretical and practical guarantees in terms of quality of the output. Depending on the parameter tuning of SimpleLocal, one can obtain any of the flow-based methods proposed in \citet{lang2004flow,andersen2008algorithm,orecchia2014flow}.
Therefore, we will only use SimpleLocal in our experiments, and we will use different parameter tuning such that we obtain performance for a wide-range of methods. Moreover, SimpleLocal requires initialization with a set of seed nodes that ideally has some overlap with the target cluster. Therefore, in this paper, we will use SimpleLocal with three different initialization techniques. Below we comment on why we choose the following inputs to SimpleLocal.

\begin{enumerate}
\item{{$\ell_1$-reg. PR-SL}:} SimpleLocal using the output of $\ell_1$-reg. PR as input.
\item{{BFS-SL}:} We will initialize SimpleLocal using the output of a breadth-first-search-type (BFS) algorithm starting from a given seed node. The algorithm that is used for initialization of SimpleLocal is shown in Algorithm \ref{algo:bfs}. Details about how we use this algorithm are given in the next subsection. Briefly, Algorithm \ref{algo:bfs} is used as a seed node expansion technique, which has also been used in \citet{veldt2016simple}.
\end{enumerate}

Note that all these flow-based methods are ``improve'' methods, i.e., they are not stand-alone, and they require initial input from some other stand-alone method, such as APPR or $\ell_1$-reg. PR. This is both a theoretical and a practical argument. It is a theoretical argument because in the theoretical analysis of SimpleLocal (see Theorem 3 in \citet{orecchia2014flow}), the input to SimpleLocal is required to have sufficient overlap with the target cluster. Otherwise, currently there is no guarantee that flow-improve methods output a reasonable approximation to the target cluster in terms of its conductance value (Theorem 1 in \citet{orecchia2014flow}) or in terms of false and true positives (Theorem 3 in \citet{orecchia2014flow}). Theorem 3 in \citet{orecchia2014flow} suggests that sufficient overlap can be achieved either by using a local spectral method such as APPR or $\ell_1$-reg. PR. In this paper, we use only the latter. This is because, as shown in \citet{fountoulakis2019variational}, theoretically and empirically APPR is very similar to $\ell_1$-reg. PR. We also show extensive experiments in this section about the similarity of APPR and $\ell_1$-reg. PR.
In practice one could use heuristics such as a BFS-type algorithm to slightly expand a seed node within a target cluster and then provide the output of  the BFS-type as input to SimpleLocal, i.e., see also \citet{veldt2016simple}. In Algorithm \ref{algo:bfs} we provide a pseudo-code for the BFS-type algorithm that we use in this paper. Basically, the only difference with a standard BFS algorithm is that we explore neighbors in batches, which allows us to terminate the BFS algorithm after a given number of steps.

\subsubsection{Parameter tuning}



APPR and $\ell_1$-reg. PR have two parameters, the teleportation parameter $\alpha$ and a tolerance/$\ell_1$-regularization parameter $\rho$. The teleportation parameter should be set according to the reciprocal of the mixing time 
of a random walk within the target cluster, which is equal to the smallest nonzero eigenvalue of the normalized Laplacian for the subgraph that corresponds to the target cluster. See \citet{zhu2013local} for details.
Let us denote the eigenvalue with $\lambda$. 
In our case the target cluster is a given ground truth cluster. We use this information to set parameter $\alpha$. In particular, for each node in the target cluster we run APPR and $\ell_1$-reg. PR $4$ times where $\alpha$ is set based on a range of values in $[\lambda/2,2\lambda]$ with a step of $(2\lambda - \lambda/2)/4$. The regularization parameter of $\ell_1$-reg. PR has nearly identical purpose as the tolerance parameter of APPR. Both parameters are set to be proportional to inverse of the volume of the target cluster, as suggested by theoretical arguments in this paper as well as in \citet{andersen2006local,zhu2013local,fountoulakis2019variational}. For each parameter setting we use sweep cut to round the output of APPR and $\ell_1$-reg. PR to find a cluster of low conductance. The sweep cut rounding procedure is a common technique to post-process the output of local graph clustering methods---for details, we refer the reader to \citet{andersen2006local,zhu2013local,fountoulakis2019variational}. We use a proximal coordinate descent algorithm \citep{fountoulakis2019variational} to solve the $\ell_1$-reg. PR problem. 
Over all parameter 
settings, we return the cluster with the lowest conductance value as an output of APPR and $\ell_1$-reg. PR.


SimpleLocal has only one parameter denoted by $\delta$. The $\delta$ parameter controls how localized the output is going to be around the input seed nodes \citep{veldt2016simple}, it also controls the quality of the output in terms of its conductance value (Theorem 1 in \citet{orecchia2014flow}) or in terms of false and true positives (Theorem 3 in \citet{orecchia2014flow}). The parameter has to satisfy $\delta \ge 0$, and according to \citet{orecchia2014flow} it should be set such that $\vol(R)/\vol(V-R) + \delta = 1/(3/\sigma + 3)$, where $R$ is the set of seed nodes that is given as input to SimpleLocal, and $\sigma$ is a parameter that satisfies $\sigma\in[\vol(R)/\vol(V-R),1]$. It is suggested in \citet{orecchia2014flow} to set $\sigma = \mathcal{O}(\vol(R\cap K) / \vol(K))$, where $K$ denotes the target cluster. Therefore, in our experiments we set 
$$\sigma = \min\left[\max\left(\frac{\vol(R)}{\vol(V-R)},\frac{\vol(R\cap K)}{\vol(K)}\right),1\right].$$ 
When we use $\ell_1$-reg. PR as input to SimpleLocal, this parameter setting also provides the theoretical guarantees which are claimed in \citet{orecchia2014flow} (assuming that all assumptions about the target cluster are satisfied).

We set parameter \textit{steps} $=n$ in Algorithm \ref{algo:bfs} and we terminate Algorithm \ref{algo:bfs} when the current set of seed nodes, denoted by \textit{seeds}, satisfies $\vol(\mbox{\textit{seeds}} \cap K) / \vol(K) \ge 0.75$, i.e., the output of Algorithm \ref{algo:bfs} has to have at least $75\%$ overlap with the target cluster, or when $\vol(\mbox{\textit{seeds}})/\vol(G) \ge 0.25$, i.e., the volume of the output is larger than $25\%$ the volume of the graph. This setting is also motivated by Theorem 3 in \citet{orecchia2014flow}, which mentions that the input to SimpleLocal has $75\%$ overlap with the target cluster. However, since this relies on the assumption that the input set of nodes is the output of APPR or $\ell_1$-reg. PR, which are themselves clustering algorithms, while Algorithm \ref{algo:bfs} is not, then we also add the second termination criterion that the output cannot have volume more than $25\%$ the volume of the graph.

\begin{algorithm}[H]
\SetKwInOut{Input}{Input}
\SetKwInOut{Output}{Output}
\Input{\textit{seeds} - set of seed nodes that we want to expand}
\Output{\textit{seeds} - updated/expanded set of seeds}
\Parameter{\textit{steps} - number of steps of the algorithm}
 Create queue Q\;
 Set all nodes in \textit{seeds} as visited\;
 Set \textit{step} $=1$\;
 \While{step $\le$ steps}{
  \textit{k} $=1$ and \textit{l} $=$ size of Q\;
  \While{k $\le$ l}{
    remove node \textit{u} from head of Q\;
    mark and enqueue all (unvisited) neighbours of \textit{u}\;
    Add newly visited nodes in set \textit{seeds}\;
    increase \textit{k}\;
  }
  increase \textit{step}\;
 }
 \caption{A modified BFS algorithm for seed set expansion}\label{algo:bfs}
\end{algorithm}


\subsubsection{Results}

\begin{figure}
\centering     
\subfigure[F1score vs conductance]{\label{fig:f1_orkut_appr_vs_l1}\includegraphics[scale=0.35]{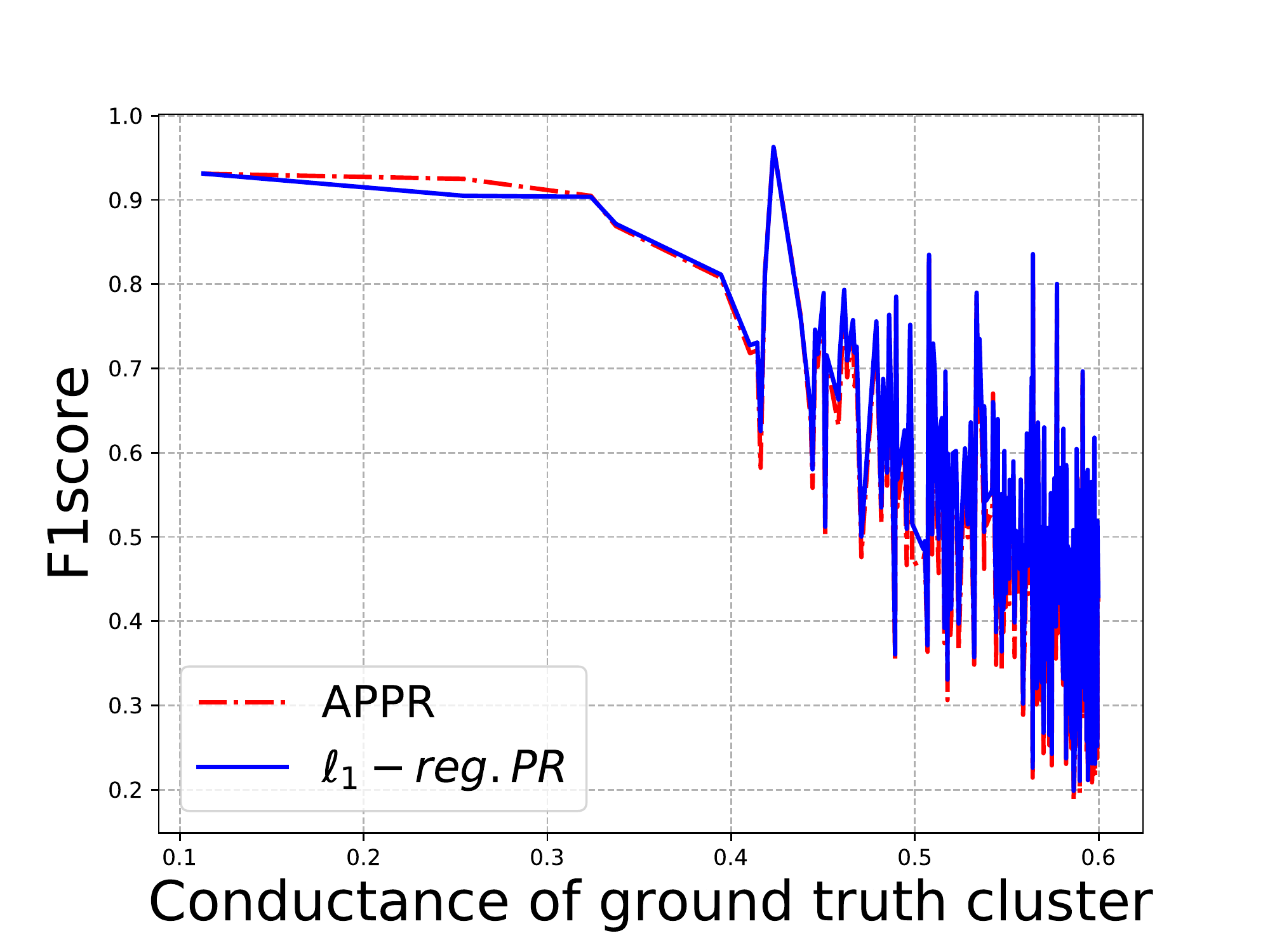}}
\subfigure[Recall vs conductance]{\label{fig:re_orkut_appr_vs_l1}\includegraphics[scale=0.35]{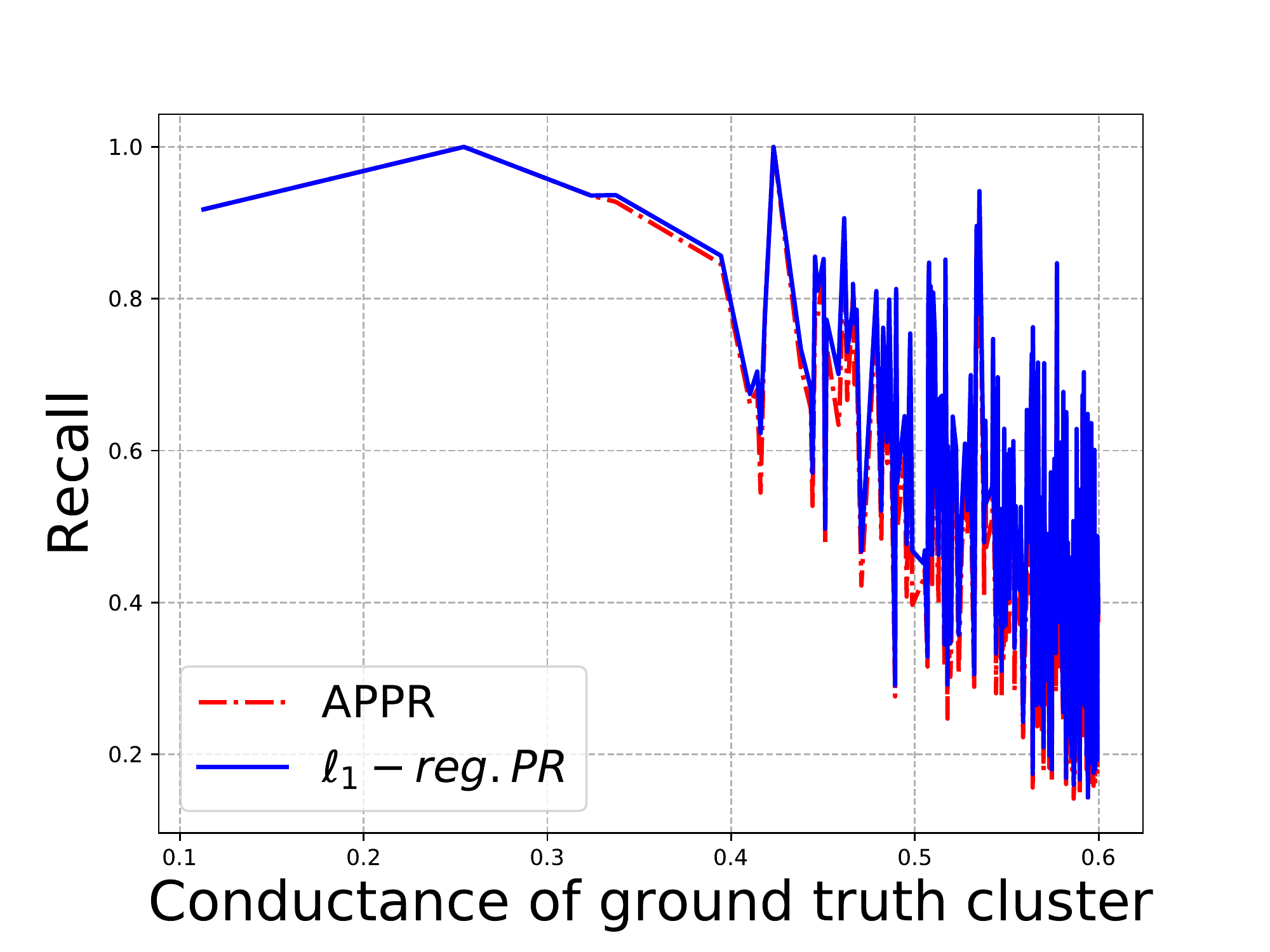}}
\subfigure[Precision vs conductance]{\label{fig:pre_orkut_appr_vs_l1}\includegraphics[scale=0.35]{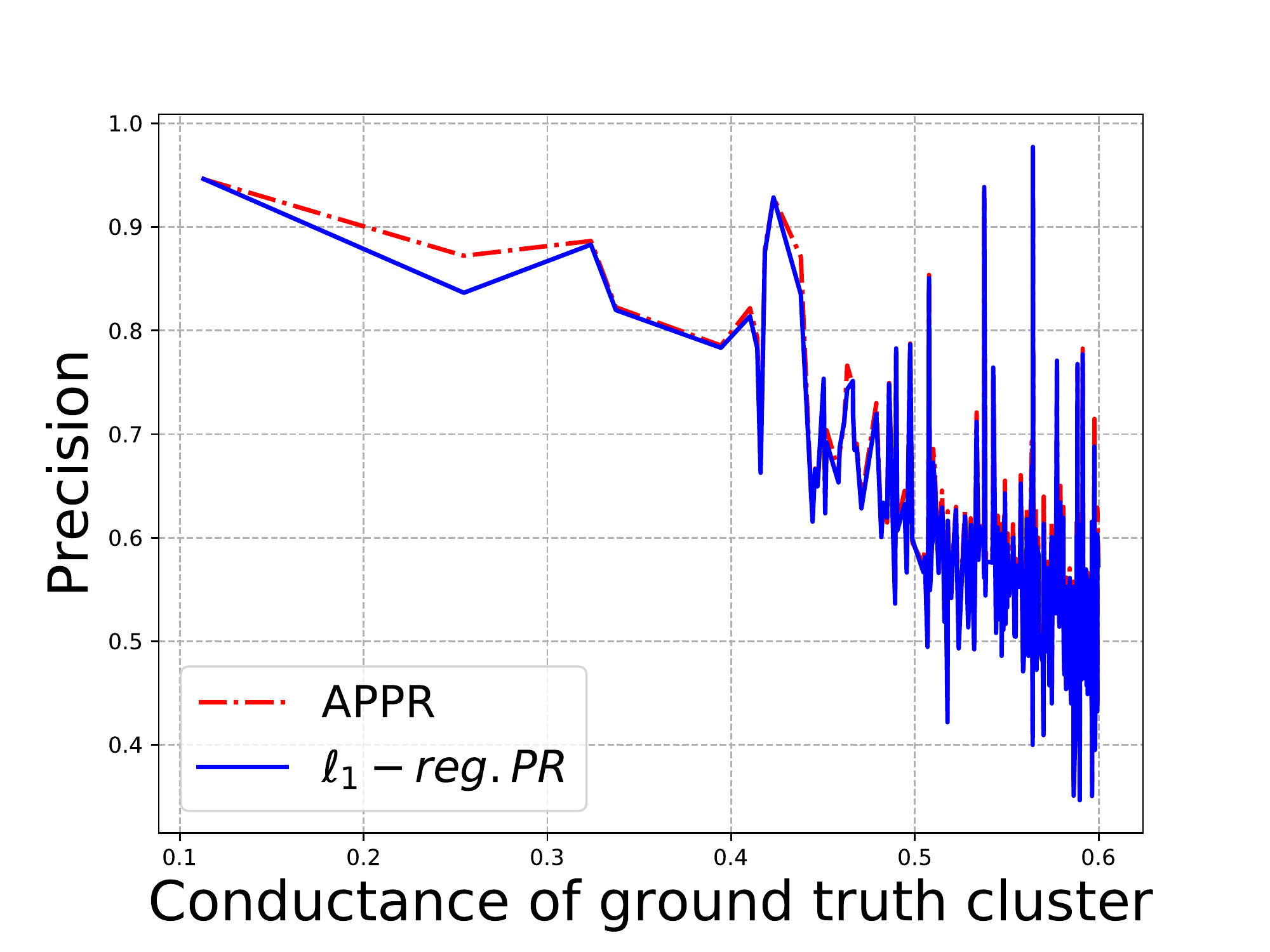}}
\caption{We demonstrate the similarity of APPR and $\ell_1$-reg. PR by illustrating F1score, recall and precision vs conductance of ground truth clusters plots for the Orkut dataset (282 ground truth clusters). This plot demonstrates two properties of the algorithms. First, it demonstrates performance of $\ell_1$-reg. PR and APPR as conductance of the target cluster increases (horizontal-axis). Observe that as conductance becomes larger then overall performances decreases. Second, it demonstrates that the methods $\ell_1$-reg. PR and APPR get nearly identical results, which is also justified by theoretical arguments. See the main text for details.}
\label{fig:orkut_appr_vs_l1}
\end{figure}

\textbf{APPR and $\ell_1$-reg. PR are similar.} We start our empirical observations by comparing APPR and $\ell_1$-reg. PR. We already proved in Theorem \ref{thm:appr_vs_l1} that these methods are similar. However, we would like to point out some additional details, which will help us justify simplification of the experiments for the remainder of this section. 
In Figure \ref{fig:orkut_appr_vs_l1}, we make a much more extended empirical comparison between the two local spectral methods, where we compare their precision, recall and F1score. We use the Orkut dataset, and we compare the two methods using all $282$ ground truth clusters of this dataset. In this figure, we present average results over all nodes for each given ground truth cluster. We observe that APPR and $\ell_1$-reg. PR produce output with nearly identical precision, recall and F1score. We attribute any minor differences between the two methods to the minor differences between the optimality conditions of $\ell_1$-reg. PR and APPR. Moreover, APPR and proximal coordinate descent for $\ell_1$-reg. PR have similar running time in practice for producing the results in Figure \ref{fig:orkut_appr_vs_l1}.  This is justified by worst-case analysis in \citet{fountoulakis2019variational} as well as our average-case analysis in Theorem \ref{thm:appr_thm}. In particular, each method required $75$ hours to produce Figure \ref{fig:orkut_appr_vs_l1}, which required $304960$ calls to APPR and $\ell_1$-reg. PR. Due to the similarity of the two approaches, we will only use $\ell_1$-reg. PR in the subsequent experiments. Note that in this experiment we do not show performance of flow-based methods because based on our empirical observations on small datasets like FB-Johns55 and Colgate88 it would have taken close to a month to obtain similar plots like in Figure \ref{fig:orkut_appr_vs_l1}.  (Improving this situation for local flow improve methods is an important direction for future work.)  For example, see the running times of flow-based methods for FB-Johns55 and Colgate88 in Table \ref{tab:times_clusters_facebook}.

\noindent \textbf{APPR and $\ell_1$-reg. PR work well for low conductance target clusters.} Here, we comment on the performance of APPR and $\ell_1$-reg. PR in Figure \ref{fig:orkut_appr_vs_l1} on the Orkut dataset. We observe that performance of both methods decreases as the conductance of the target cluster increases, i.e., as the cluster quality gets worse. This is an expected outcome that is predicted by the average-case analysis of this paper. However, it is important to mention that, for target clusters with conductance close to $0.4$, the F1score of $\ell_1$-reg. PR is around $0.8$. This is surprisingly good performance since a cluster with conductance $0.4$ means that roughly half of the edges of the target cluster cross the cluster boundary (which implies that such target clusters are actually quite noisy and not of particularly high quality).

\noindent \textbf{Results for biology datasets and Sfld and PPI-mips.}\ The results for the biology datasets are shown in Table \ref{tab:results_clusters_biology}.
We present average results over all nodes for each given ground truth cluster. 
We denote with bold numbers the performance number of a method when it has the \textit{largest F1score} among all~methods. 

We note the consistent state-of-the-art performance of $\ell_1$-reg. PR for all clusters in Table \ref{tab:results_clusters_biology}. For some ground truth clusters, $\ell_1$-regularized PageRank perfectly recovers the target clusters, which is mainly attributed to the fact that the ground truth clusters have strong separability property (see also Corollary \ref{cor:exact_recovery} of the main text). In most experiments, SimpleLocal did not improve the performance of the input of $\ell_1$-reg. PR, but also it did not make it worse. In some cases, like the Spindle ground truth cluster in the PPI-mips dataset, SimpleLocal decreased the performance of $\ell_1$-reg. PR in terms of the F1score. This is because SimpleLocal found clusters that have smaller conductance, but that do not correspond to clusters with the highest F1score. This is a known issue that has been mentioned in \citet{fountoulakis2017optimization}. BFS-SL has the worst performance among all methods in most experiments. In fact, BFS-SL performs well only for the usrease ground truth cluster in the Sfld dataset. The performance of BFS-SL is especially poor for all ground truth clusters in the PPI-mips dataset. It is important to mention that we did experiment with different parameter tuning for both BFS-type Algorithm \ref{algo:bfs} and SimpleLocal for the BFS-SL method, but the performance was poor for all settings of parameters that we tried. We attribute the poor performance of BFS-SL in the BFS-type algorithm, which provides the input to SimpleLocal. In particular, the BFS-type Algorithm \ref{algo:bfs} is not related to clustering in a general sense, and this translates to poor quality input to SimpleLocal. As is mentioned in the theoretical analysis in~\citet{orecchia2014flow,veldt2016simple}, SimpleLocal requires as input the output of a local spectral method such as $\ell_1$-reg. PR in order to perform well, which is also verified by the results in Table \ref{tab:results_clusters_biology}.

The corresponding running times for each method are shown in Table \ref{tab:times_clusters_biology}. Each numeric value in this table demonstrates the total running time to run a method for a given ground truth cluster. The running time is the sum of the running times for each node in the ground truth cluster. Note that $\ell_1$-reg. PR is the fastest method. 

\begin{table}[h] 
    \caption{Results for biology datasets Sfld and PPI-mips. In this table, we present average results of F1score over all nodes for each given ground truth cluster. We denote with bold numbers the performance number of a method when it has the largest score among all methods.}
\vspace{2mm}
  \centering
    \begin{adjustbox}{max width=\textwidth}
    \begin{tabular}{clccc}
    dataset & \multicolumn{1}{c}{feature} & $\ell_1$-reg. PR & BFS-SL & $\ell_1$-reg. PR-SL\\
    \midrule
     \multirow{2}{*}{\rotatebox[origin=c]{90}{Sfld}}          & urease       & $\bf 0.75$   & $0.42$   & $0.38$ \\
      								                  & AMP           & $\bf 0.86$   & $\bf 0.86$ & $\bf 0.86$\\\hdashline
     \multirow{19}{*}{\rotatebox[origin=c]{90}{PPI-mips}} & Actin         & $\bf 0.98$   & $0.04$      & $\bf 0.98$\\
                 								 & Anaphase   & $\bf 1.00$   & $0.09$      & $\bf 1.00$  \\
                 								 &Cdc28p       & $\bf 1.00$    & $0.10$      & $\bf 1.00$   \\
                 								 &Coat            & $\bf 0.85$    & $0.04$      & $\bf 0.85$  \\
                 								 & ct-large      & $\bf 1.00$     & $0.02$      & $\bf 1.00$ \\
                 								 & ct-small      & $\bf 1.00$    & $0.05$       & $\bf 1.00$  \\
                 								 & F0-F1-ATP & $\bf 1.00$    & $0.06$      &  $\bf 1.00$   \\
                 								 & H+              & $\bf 1.00$    & $0.06$     & $\bf 1.00$  \\
                 								 & mc-large     & $\bf 1.00$    & $0.03$     & $\bf 1.00$  \\
                 								 & mc-small     & $\bf 1.00$    & $0.07$    & $\bf 1.00$  \\
                 								 &mc-complex & $0.78$         & $0.07$    & $\bf 0.79$  \\
                 								 &mRNA          & $\bf 0.93$    & $0.03$    & $\bf 0.93$   \\
                 								 & Nuclear       & $\bf 0.85$    & $0.05$    & $\bf 0.85$   \\
                 								 &RNA             & $\bf 0.87$    & $0.03$    & $0.80$  \\
                 								 &Spindle        & $\bf 0.85$    & $0.03$    &  $0.82$   \\
                 								 & TRAPP       &  $\bf 1.00$   & $0.10$    & $\bf 1.00$  \\
                 								 &tRNA           &  $\bf 1.00$   & $0.09$     & $\bf 1.00$  \\
                 								 &19-22S        &  $\bf 1.00$   & $0.05$     & $\bf 1.00$  \\
                 								 & 20S            & $\bf 1.00$    & $0.06$     & $\bf 1.00$   \\
    \midrule
    \end{tabular}%
    \end{adjustbox}
  \label{tab:results_clusters_biology}%
\end{table}%

\begin{table}[h] 
    \caption{Running times for biology datasets Sfld and PPI-mips. The numeric entries of the table show the total time in seconds over all nodes of a given ground truth cluster. The running times of $\ell_1$-reg. PR where less than $0.1$ for most experiments, and they have been rounded up to $0.1$.}
\vspace{2mm}
  \centering
  \begin{adjustbox}{max width=\textwidth}
    \begin{tabular}{clccc}
    dataset & \multicolumn{1}{c}{feature} & $\ell_1$-reg. PR & BFS-SL & $\ell_1$-reg. PR-SL\\
    \midrule
     \multirow{2}{*}{\rotatebox[origin=c]{90}{Sfld}}          & urease       & $17.2$  & $29.8$   & $151.2$ \\
      								                  & AMP           & $0.2$     & $2.5$ & $1.6$    \\\hdashline
     \multirow{19}{*}{\rotatebox[origin=c]{90}{PPI-mips}} & Actin         & $0.1$  & $1.8$   & $0.6$ \\
                 								 & Anaphase   & $0.1$  & $0.8$   & $0.1$ \\
                 								 &Cdc28p        & $0.1$  & $0.7$   & $0.1$ \\
                 								 &Coat             & $0.1$  & $1.5$   & $0.3$ \\
                 								 & ct-large       & $3.6$  & $8.5$   & $16.5$ \\
                 								 & ct-small       & $1.3$  & $10.0$   & $6.6$ \\
                 								 & F0-F1-ATP  & $0.1$  & $1.0$   & $0.1$ \\
                 								 & H+               & $0.1$  & $1.1$   & $0.1$ \\
                 								 & mc-large      & $0.3$  & $3.3$   & $2.0$ \\
                 								 & mc-small     & $0.1$  & $0.9$   & $0.1$ \\
                 								 &mc-complex & $0.1$  & $0.8$   & $0.2$ \\
                 								 &mRNA          & $0.3$  & $3.5$   & $2.4$ \\
                 								 & Nuclear       & $0.1$  & $1.3$   & $0.3$ \\
                 								 &RNA             & $0.2$  & $1.5$   & $2.7$ \\
                 								 &Spindle        & $0.1$  & $2.3$   & $0.7$ \\
                 								 & TRAPP       & $0.1$  & $0.7$   & $0.1$ \\
                 								 &tRNA           & $0.1$  & $0.7$   & $0.1$ \\
                 								 &19-22S        & $0.1$  & $1.3$   & $0.2$ \\
                 								 & 20S            & $0.1$  & $1.0$   & $0.1$ \\
    \midrule
    \end{tabular}%
    \end{adjustbox}
  \label{tab:times_clusters_biology}%
\end{table}%

\noindent \textbf{Results social networks FB-Johns55 and Colgate88.}\ The results for the social network graphs are shown in Table \ref{tab:results_clusters_facebook}. There are a lot of interesting observation for this set of experiments.
First, $\ell_1$-reg. PR outperforms BFS-LS, with the exception of the ground truth clusters of major index 217 in  FB-Johns55, where BFS-LS has a $0.03$ larger F1score, and the ground truth cluster of year 2009 in Colgate88, where BFS-SL has the same F1score as $\ell_1$-reg. PR.
We observed two reasons that BFS-SL has worse performance in most experiments. The first reason is that BFS-SL outputs a cluster that has smaller conductance than the output cluster of $\ell_1$-reg. PR, but better conductance is often not related to the ground truth cluster, especially in cases where the ground truth cluster itself has large conductance. We attribute this behavior to SL because as an algorithm it attempts to find a cluster with small conductance. The second reason is that the input to SL from BFS-type Algorithm \ref{algo:bfs} is not a good approximation to the ground truth cluster, which is a required property of SL such that it performs well.

The second set of observations is about $\ell_1$-reg. PR-SL. For most ground truth clusters, we observe that $\ell_1$-reg. PR-SL performs worse than or on par with $\ell_1$-reg. PR, with the exception of ground truth clusters year 2009 and major index 217 in FB-Johns55 and ground truth clusters of years 2008 and 2009 in Colgate88. When $\ell_1$-reg. PR-SL makes the input of $\ell_1$-reg. PR worse, it is clearly because the former finds a cluster with better conductance value which does not relate to the ground truth cluster. We observe this behavior very often when the target cluster does not have small conductance value, and this is also confirmed through our simulation study (see Figure \ref{fig:FR_TP_d} in Section \ref{sec:simulated_data}). When $\ell_1$-reg. PR-SL performs better it is because the ground truth cluster has small conductance but not small enough such that $\ell_1$-reg. PR performs well by itself. In particular, $\ell_1$-reg. PR leaks more mass outside of the target cluster than it should, and this results in small precision. This is a well-known problem that has been also observed in \citet{fountoulakis2017optimization}, which can be fixed by SL.

The corresponding running times for each method are shown in Table \ref{tab:times_clusters_facebook}. The running time is the sum of the running times for each node in the ground truth cluster. Note that $\ell_1$-reg. PR is the fastest method. 

\begin{table}[h] 
    \caption{Results for Facebook datasets FB-Johns55 and Colgate88. In this table we present average results of F1score over all nodes for each given ground truth cluster. We denote with bold numbers the performance number of a method when it has the largest score among all~methods.}
\vspace{2mm}
  \centering
  \begin{adjustbox}{max width=\textwidth}
    \begin{tabular}{clccc}
    dataset & \multicolumn{1}{c}{feature} & $\ell_1$-reg. PR & BFS-SL & $\ell_1$-reg. PR-SL\\
    \midrule
     \multirow{9}{*}{\rotatebox[origin=c]{90}{FB-Johns55}} 
     										 & year $2006$           & $\bf 0.32$ & $0.13$ & $0.23$ \\
      								                  & year $2007$           & $\bf 0.43$ & $0.17$ & $0.31$ \\
     										 & year $2008$           & $\bf 0.50$ & $0.34$ & $0.36$  \\
      								                  & year $2009$           & $0.84$      & $0.78$ & $\bf 0.89$  \\
     										 & major index $217$  & $0.85$      & $0.88$ & $\bf 0.88$   \\
      								                  & second major $0$   & $\bf 0.41$ & $0.20$ & $0.20$  \\
     										 & dorm $0$                 & $\bf 0.46$ & $0.13$ & $0.08$   \\
      								                  & gender $1$              & $\bf 0.42$ & $0.21$ & $0.21$ \\
      								                  & gender $2$              & $\bf 0.46$ & $0.19$ & $0.18$  \\\hdashline
     \multirow{10}{*}{\rotatebox[origin=c]{90}{Colgate88}} 
     										 & year $2004$           & $0.42$       & $0.29$ & $\bf 0.44$   \\
      								                  & year $2005$           & $\bf 0.44$  & $0.15$ & $0.43$ \\
     										 & year $2006$           & $\bf 0.46$  & $0.27$ & $0.39$  \\
      								                  & year $2007$           & $\bf 0.54$  & $0.25$ & $0.46$  \\
     										 & year $2008$           & $0.75$       & $0.56$ & $\bf 0.88$  \\
      								                  & year $2009$           & $0.96$       & $0.96$ & $\bf 0.98$  \\
      								                  & second major $0$   & $\bf 0.49$  & $0.24$ & $0.25$ \\
     										 & dorm $0$                & $\bf 0.46$   & $0.04$ & $0.26$  \\
      								                  & gender $1$             & $\bf 0.45$   & $0.21$ & $0.25$ \\
      								                  & gender $2$             & $\bf 0.34$   & $0.20$ & $0.26$  \\
    \midrule
    \end{tabular}%
    \end{adjustbox}
  \label{tab:results_clusters_facebook}%
\end{table}%

\begin{table}[h] 
  \caption{Running times for Facebook datasets FB-Johns55 and Colgate88. The numeric entries of the table show the total time in seconds over all nodes of a given ground truth cluster.}
\vspace{2mm}
  \centering
  \begin{adjustbox}{max width=\textwidth}
    \begin{tabular}{clcccc}
    dataset & \multicolumn{1}{c}{feature} & $\ell_1$-reg. PR & BFS-SL & $\ell_1$-reg. PR-SL\\
    \midrule
     \multirow{9}{*}{\rotatebox[origin=c]{90}{FB-Johns55}} 
     										 & year $2006$           & $307$  & $28126$   & $56022$ \\
      								                  & year $2007$           & $453$  & $28083$    & $61462$ \\
     										 & year $2008$           & $1125$  & $25855$    & $62668$ \\
      								                  & year $2009$           & $852$  & $15736$    & $15090$ \\
     										 & major index $217$  & $80$  & $728$        & $768$ \\
      								                  & second major $0$   & $8866$  & $91500$    & $308041$ \\
     										 & dorm $0$                 & $10865$  & $70084$   & $260485$ \\
      								                  & gender $1$              & $1533$  & $69265$   & $205031$ \\
      								                  & gender $2$              & $13080$  & $83836$   & $301843$ \\\hdashline
     \multirow{10}{*}{\rotatebox[origin=c]{90}{Colgate88}} 
     										 & year $2004$           & $11$  & $1482$         & $659$ \\
      								                  & year $2005$           & $70$  & $5800$       & $7679$ \\
     										 & year $2006$           & $99$  & $5471$       & $11398$ \\
      								                  & year $2007$           & $154$  & $6198$       & $11641$ \\
     										 & year $2008$           & $287$  & $4756$        & $6900$ \\
      								                  & year $2009$           & $530$  & $2882$      & $2507$ \\
      								                  & second major $0$  & $2876$  & $24525$   & $71937$ \\
     										 & dorm $0$                & $449$  & $15184$   & $36992$ \\
      								                  & gender $1$             & $1533$  & $20347$   & $53016$ \\
      								                  & gender $2$             & $430$  & $16854$   & $39537$ \\
    \midrule
    \end{tabular}%
    \end{adjustbox}
  \label{tab:times_clusters_facebook}%
\end{table}%

\noindent \textbf{Memory scalability for small- and large-scale graphs.}\ One of the main motivations of the $\ell_1$-regularized PageRank model is the low memory requirements for each seed node.
In Table~\ref{tab:memory} we illustrate performance in terms of memory requirements for solving the $\ell_1$-regularized PageRank using proximal coordinate descent.
In particular, we illustrate how memory requirement increases as $\rho$ decreases. As is expected, for large values of $\rho$ the output solution is very sparse and the algorithms have very low memory requirements. As $\rho$ gets smaller then 
the solutions become denser and memory requirements increase. We observe similar performance for FB-Johns55 (small dataset) and Orkut (large dataset).


\begin{table}[h!]
 \caption{Memory scalability for proximal coordinate descent for solving the $\ell_1$-regularized PageRank as $\rho$ decreases. We illustrate averaged results over $100$ trials, i.e., seed nodes. The numbers in the first row of each dataset are Megabytes. Note that to compute the memory requirements for storing the adjacency matrix we did not store the edge weights since the graphs are unweighted.}
\vspace{2mm}
  \centering
  \begin{adjustbox}{max width=\textwidth}
 \begin{tabular}{c l c c c c c c c} 
 dataset & \multicolumn{1}{c}{statistics}& $\rho=$1.0e-1 & 1.0e-2 & 1.0e-3 & 1.0e-4 & 1.0e-5 & 1.0e-6 & 1.0e-7 \\ 
 \hline
 \multirow{5}{*}{\rotatebox{90}{Orkut}} & Average memory                                                                          & $0.11$        & $0.20$     & $0.27$      & $1.00$      & $3.51$       & $22.8$      & $100$\\ 
                                                             & $\frac{\mbox{Average memory}}{\mbox{mem. for adjacency}}$  & $1.2$e-4  & $2.1$e-4 & $2.9$e-4 & $1.0$e-3 & $3.6$e-3 &$2.4$e-2 &$1.0$e-1  \\ 
                                                             & Average nodes touched                                                               & $72$         & $72$       & $131$     &$1474$    & $10788$ &$83139$  & $443857$ \\ 
                                                             & Average nodes in solution                                                            & $1$          & $1$          & $4$         &$33$        & $266$    & $2401$   & $22159$\\\hdashline 
 \multirow{5}{*}{\rotatebox{90}{FB-Johns55}} & Average memory                                                                          & $0.09$     & $0.17$     & $0.20$   & $0.75$     & $1.44$    & $2.41$   & $2.65$\\ 
                                                                       & $\frac{\mbox{Average memory}}{\mbox{mem. for adjacency}}$  & $6.3$e-2  & $1.1$e-1 & $1.3$e-1 & $4.9$e-1 & $9.5$e-1 &$1.5$e+0 & $1.7$e+0 \\ 
                                                                       & Average nodes touched                                                               & $66$         & $68$       & $114$     &$959$       & $3113$ &$5079$     & 5157 \\ 
                                                                       & Average nodes in solution                                                            & $1$          & $1$          & $3$         &$36$        & $367$    & $4201$   & 5157\\ 
 \hline
 \end{tabular}
    \end{adjustbox}
  \label{tab:memory}%
\end{table}

\section{Conclusion}
\label{sec:discussion}


In this paper, we have examined the $\ell_1$-regularized PageRank optimization problem for local graph clustering.
In a local graph clustering problem, the objective is to find a single target cluster in a large graph, given a seed node in the cluster, and to do so in a running time that does not depend on the size of the entire graph, but instead that depends on the size of the output cluster.\ 
Algorithmic results (i.e., running-time bounds to achieve  a given cluster quality) for local graph clustering abound, but our results are the first statistical results (i.e., where one is interested in recovering a cluster under an hypothesized model).
Under our local random model, we show that the optimal support of $\ell_1$-regularized PageRank identifies the target cluster with bounded false positives, and in certain settings exact recovery is also  possible.\ 
We further establish a strong connection between $\ell_1$-regularized PageRank and approximate personalized PageRank (APPR), based on which we obtain similar statistical guarantees on APPR under the random model.
Additionally, we have brought the idea of solution path algorithms from the sparse regression literature to the local graph clustering literature, and we showed that the forward stagewise algorithm gives a provable approximation to the entire $\ell_1$ regularization path of this algorithm. 
Finally we demonstrate the state-of-the-art performance of $\ell_1$-regularized PageRank on both simulated and real data graphs.



\acks{{We would like to acknowledge support for this project
from the National Science Foundation (NSF grant IIS-9988642)
and the Multidisciplinary Research Program of the Department
of Defense (MURI N00014-00-1-0637).}
This material is 
also 
based 
in part 
on research sponsored by DARPA and the Air Force Research Laboratory under agreement number FA8750-17-2-0122. The U.S. Government is authorized to reproduce and distribute reprints for Governmental purposes notwithstanding any copyright notation thereon.  The views and conclusions contained herein are those of the authors and should not be interpreted as necessarily representing the official policies or endorsements, either expressed or implied, of DARPA and the Air Force Research Laboratory or the U.S. Government.
We would also like to acknowledge ARO, NSF,  ONR, and Berkeley Institute for Data Science for providing partial support of this work.
}


\clearpage

\newpage

\appendix

\section{Some auxiliary lemmas}\label{sec:collect_lemmas}

To establish the theorems, we need a few concentration inequalities and intermediate results that shall be used in the proof. In Section~\ref{sec:concen_lemma}, we give degree concentration inequalities for random graphs, and in Section~\ref{sec:exp_graph}, we state recovery guarantees of $\ell_1$-regularized PageRank on the population graph. Section~\ref{sec:reduced_l1_pagerank} gives a few important results on the $\ell_1$-regularized PageRank when restricted to the target cluster.

\subsection{Concentration lemmas}\label{sec:concen_lemma}

Here, we present several concentration lemmas for degrees of random graphs.
The first lemma is consequence of Chernoff's inequality for the sum of independent Bernoulli random variables, and applying union bound (c.f. see~\citet[Theorem 2.3.1]{vershynin2018high}).

\begin{lemma}[{Adapted from~\citet[Proposition 2.4.1]{vershynin2018high}.}]
\label{lem:concen}
Let $X_i$ be the sum of independent Bernoulli random variables, with $\EE{X_i}=\mu$ for $i=1,2,\ldots,k$. Then, there exists a universal  constant $c_0>0$ such that if $ \mu \geq c_0^{-1}\delta^{-2}\log k$, then with probability at least $1-2e^{-c_0\delta^2 \mu}$,
\begin{align*}\text{For all $i\in K$: } \;  \Abs{X_i - \mu }\leq \delta\mu. \end{align*}
\end{lemma}

According to our random model (Definition~\ref{def:model}), the degree vector $d_K$ for the target cluster $K$ comprises of random variables which are the sum of independent Bernoulli random variables with common mean $\bar{d}$. Therefore, by applying Lemma~\ref{lem:concen}, it is straightforward to see the following result.

\begin{lemma}[{Adapted from~\citet[Proposition 2.4.1]{vershynin2018high}.}]
\label{lem:concen_degree}
There exists a universal  constant $c_0>0$ such that if $\bar{d}\geq c_0^{-1}\delta^{-2}\log k$, then with probability at least $1-2e^{-c_0\delta^2\bar{d}}$,
\begin{align*}\text{For all $i\in K$: } \;  \Abs{d_i - \bar{d}}\leq \delta \bar{d}, \end{align*}
where $\bar{d}$ is the average degree of the vertices in the cluster $K$.
\end{lemma}

The next lemma bounds the number of edges between node $j\in K^c$ and the target cluster $K$ when $q=\O{\frac{1}{n}}$.
\begin{lemma}
\label{lem:concen_degree_K}
Suppose that $q=c/n$  and that $k\geq 2(c+3)$ for a positive constant $c$. Then for $n$ sufficiently large, with probability at least $1-\O{n^{-1}}$,
\[ \max_{j\in K^c}\norm{A_{j,K}}_1\leq c+2 .  \]
\end{lemma}

\subsection{Exact recovery of target cluster in population graph}
\label{sec:exp_graph}

Next, we define the ``ground truth'' PageRank vector, which we obtain by applying $\ell_1$-regularized PageRank to the population graph $\overline{G}$ of our random model.\ Recall the adjacency matrix $\EE{A}$ defined in~\eqref{eqn:adj_avg}, the associated diagonal degree matrix $\EE{D}$, and the Laplacian matrix $\EE{\lap}$.\ Writing $\EE{Q}=\alpha\EE{D}+\frac{1-\alpha}{2}\EE{\lap}$, we denote the population version of $\ell_1$-regularized PageRank as
\begin{equation}
\label{eqn:l1_pagerank_exp}
\xs = \argmin_{x}\left\{\frac{1}{2}x^\top \EE{Q}x - \alpha x^\top \seed+ \rho\alpha \norm{\EE{D}x}_1\right\}.
\end{equation}
Compared to~\eqref{eqn:pagerank2}, we see that the matrices associated with the graph are now replaced by their expected counterparts.\

The following lemma shows that there exist a range of $\rho$ values for which the solution to the population version of the $\ell_1$-regularized PageRank minimization problem~\eqref{eqn:l1_pagerank_exp} gives an exact recovery for the target cluster.

\begin{lemma}
[\textbf{Exact recovery for population PageRank vector.}]
\label{lem:exact_recovery}
Consider the local random model given in Definition~\ref{def:model} and suppose that Assumption~\ref{assump:rate} holds.\ Then, for $n$ sufficiently large, the ``ground truth'' PageRank vector $\xs$, defined in~\eqref{eqn:l1_pagerank_exp}, identifies $K$ correctly, i.e.,
\[\supp(\xs) =K,\]
as long as $\rho \in [\rho^\sharp, \rho^\natural)$, where
\begin{align*}
\rho^\sharp = \frac{q(1-\alpha)}{2\alpha\bar{d}\cdot \EE{{d}_{\ell_\sharp}} +q(1-\alpha)\left[ k\bar{d} + (n-k) \EE{{d}_{\ell_\sharp}}\right]} ,\;\; \text{ and }\;\; \rho^\natural = \frac{p(1-\alpha)}{\bar{d}\left[(1+\alpha)\bar{d} +(1-\alpha)p \right]},
\end{align*}
where $\EE{{d}_{\ell_\sharp}}=\min_{\ell\in K^c}\EE{{d}_{\ell}}$.\ Furthermore, $\xs$ has a closed form expression, $\xs= u\cdot \ones_S + v\cdot \ones_{K}$, where $u$ and $v$ are given by
\begin{align*}\begin{cases}u=\frac{2\alpha}{(1+\alpha)\bar{d} +(1-\alpha)p }; \\ v=\frac{ \frac{1-\alpha}{2}p \cdot u - \rho\alpha \bar{d}}{\alpha\bar{d} + \frac{1-\alpha}{2}q(n-k)}. \end{cases}
 \end{align*}
\end{lemma}
From the closed form expression, we can see that $\xs$ has high probability mass on the single seed node with a long tail further away from the seed node.\ The mass outside the target cluster is being thresholded exactly to zero via $\ell_1$-norm regularization, thus identifying the target cluster without any false positives.\ Of course, in practice, the population graph is unknown, and we are instead given an instance of the graph from the random model.\ In this case, the ground truth vector $\xs$ allows us to estimate the magnitude of the $\ell_1$-regularized PageRank vector $\xh$ by analyzing the error between $\xs$ and $\xh$.\ For more details, we refer the reader to the proof of Lemma~\ref{lem:l_infty_bound}.\

\subsection{$\ell_1$-regularized PageRank restricted to target cluster}\label{sec:reduced_l1_pagerank}

Consider the $\ell_1$-regularized PageRank problem restricted to the target cluster $K$.
\begin{align}
\label{eqn:reduced_pagerank_pen}
\xhr &= \argmin_{x\in\R^{n}} \left\{ \frac{1}{2}x^\top Q x -\alpha x^\top \seed +\rho \alpha \norm{Dx}_1 : x_{K^c}=0\right\} \nonumber \\
&=\argmin_{y\in\R^{k}} \left\{ \frac{1}{2}y^\top Q_{K,K} y -\alpha y^\top \seed_K +\rho \alpha \norm{D_{K,K}y}_1 \right\}.
\end{align}
Since our aim is to recover the target cluster $K$ based on the {\em local} model (Definition~\ref{def:model}), it is natural to analyze the properties of the solution to this reduced problem. Here, abusing notation, we use $\xhr$ to denote the vector either in the original space $\R^n$ or in the reduced space $\R^k$.

We present several lemmas about $\xhr$ that will be critical for the proof of theorems. 
First, we give a guarantee on the recovery of the target cluster for the optimal solution~\eqref{eqn:reduced_pagerank_pen}.

\begin{lemma}
\label{lem:reduced_full_recovery}
Let $\xhr$ be the solution to the reduced problem~\eqref{eqn:reduced_pagerank_pen}. Suppose that $(1-\delta)p^2 k\geq c_0^{-1}\delta^{-2}\log k$. Then, if the regularization parameter satisfies $\rho\leq \rho(\delta)$, we have
\[\supp(\xhr) = K,\]
with probability at least $1-6e^{-c_0 \delta^2 (1-\delta)p^2 k}$.
Here $c_0$ is the same constant appearing in Lemma~\ref{lem:concen}.
\end{lemma}
For the above result, since by construction $\xhr$ is zero outside $K$, all we need to show is that $\xhr_K>0$ when $\rho \approx \O{\frac{\gamma p}{\bar{d}^2}}$. Next we compare the support set of $\xhr$ to  that of $\xh$.

\begin{lemma}
\label{lem:support_lemma}
For any regularization parameter $\rho\in [0,\infty)$, we have
\[ \supp(\xhr) \subseteq \supp(\xh).\]
\end{lemma}

Finally we have the following estimate on the maximum coordinate of $\xhr$ on $K\backslash S$ (recall $u$ and $v$ are respectively the components of $\xs$ on $S$ and $K\backslash S$; see Lemma~\ref{lem:exact_recovery}).
\begin{lemma}
\label{lem:l_infty_bound}
Let $\delta\leq 0.1$ and suppose that $(1-\delta)p^2 k\geq c_0^{-1}\delta^{-2}\log k$.
Assume that Assumption~\ref{assump:rate} holds and that the size of the target cluster $k\geq 5$.
 If the regularization parameter satisfies $\rho\geq \rho(\delta)$,
then for $n$ sufficiently large,  the following bound holds
\[ \norm{\xhr_{K\backslash S}}_\infty\leq  \underbrace{v}_{=\xs_{K\backslash S}}+\underbrace{  \frac{c_1(1-\alpha)u}{\bar{d}} + 2c_1\delta (v + \rho\alpha )}_{\geq \norm{\xhr_{K\backslash S} - \xs_{K\backslash S}}_\infty},\]
with probability at least $1-6e^{-c_0\delta^2(1-\delta) p^2k}$. Here $c_0$ is the same constant appearing in Lemma~\ref{lem:concen} and $c_1$ is a positive numerical constant.
\end{lemma}

\section{Proofs of theorems}\label{sec:proof_theorems}
In this section we prove all of our theorems. Based on the lemmas presented above we give proofs of our five theorems, respectively, in Section~\ref{sec:proof_appr_vs_l1},~\ref{sec:proof_full_recovery},~\ref{sec:proof_bound_FP},~\ref{sec:proof_no_false_pos}, and~\ref{sec:proof_appr_thm}.

\subsection{Proof of Theorem~\ref{thm:appr_vs_l1}}
\label{sec:proof_appr_vs_l1}
Fix $\rho_0>0$. First, we prove the first part of the theorem, i.e., $\supp(\ph(\rho_0)) \subseteq \supp(x^{\text{APPR}}(\rho_0))$.
When $\rho_0> 1/d_S$, this relation holds trivially because both $\ph(\rho_0)$ and $x^{\text{APPR}}(\rho_0)$ do not contain any nodes in the support set. To prove that the relation holds for  $\rho_0 \leq 1/d_S$, we will proceed by induction.
Specifically, we will prove that 
\begin{equation}\label{eqn:appr_and_l1}
\begin{cases}
\ph_i(\rho_0) < x^{\text{APPR}}_i(\rho_0),   & \text{$i\in \supp(x^{\text{APPR}}_i(\rho_0))$,}   \\ \ph_i(\rho_0) = x^{\text{APPR}}_i(\rho_0)=0, &\text{$i\notin \supp(x^{\text{APPR}}_i(\rho_0))$,} \end{cases} \end{equation}
for all $\rho_0\in (0,1/d_S]$. Assuming that this holds, it is straightforward to follow that 
\[\supp(\ph(\rho_0)) \subseteq \supp(x^{\text{APPR}}(\rho_0)), \]
thus proving the claim.

Now we turn to proving~\eqref{eqn:appr_and_l1}. When $\rho_0=1/d_S$, we can  see from the algorithm~\eqref{eqn:appr} that $x^{\text{APPR}}(\rho_0)$ contains the seed node as the only element in the support set, while by Lemma~\ref{lem:nonneg_grad}, we have $\ph=0$, which proves~\eqref{eqn:appr_and_l1}. Next, assuming that it holds at some $\rho_0\in (0,1/d_S]$, let $\widetilde{\rho_0} >0$ be chosen such that $\widetilde{\rho_0}<\rho_0$ and such that there appears a node $i\in V$ that  violates the condition~\eqref{eqn:appr_and_l1} for the first time. If $i\in \supp(x^{\text{APPR}}(\widetilde{\rho_0}))$, since the path $\ph(\rho)$ is continuous by the proof of Lemma~\ref{lem:monotone}, this means that $i$ is the first node such that $\ph_i(\widetilde{\rho_0}) = x^{\text{APPR}}_i(\widetilde{\rho_0})>0$ (there may exist multiple nodes that simultaneously violate~\eqref{eqn:appr_and_l1} at $\rho=\widetilde{\rho_0}$ but the same statement still applies). In this case, by induction, we know that $\ph_j(\widetilde{\rho_0}) \leq x^{\text{APPR}}_j(\widetilde{\rho_0})$ for $j\neq i$. Then 
\begin{align*}
(D^{-1}\nabla f(\ph(\widetilde{\rho_0})))_i  &=\frac{1+\alpha}{2}\ph_i(\widetilde{\rho_0}) - \frac{1-\alpha}{2} (D^{-1}A\ph(\widetilde{\rho_0}))_i -\alpha \seed_i\\ &\geq \frac{1+\alpha}{2}x^{\text{APPR}}_i(\widetilde{\rho_0}) - \frac{1-\alpha}{2} (D^{-1}Ax^{\text{APPR}}(\widetilde{\rho_0}))_i  -\alpha \seed_i \\ &=  (D^{-1}\nabla f(x^{\text{APPR}}(\widetilde{\rho_0})))_i ,
\end{align*}
where the inequality holds since $(D^{-1}Ax)_i=\sum_{j\sim i} w_{ji}x_j/d_i$ for any $x\in\R^{|V|}$. By the optimality condition of~\eqref{eqn:pagerank2} (Lemma~\ref{lem:nonneg_grad}), the left-hand side must be $(D^{-1}\nabla f(\ph(\widetilde{\rho_0})))_i=-\widetilde{\rho_0}\alpha$ which gives 
\[(D^{-1}\nabla f(x^{\text{APPR}}(\widetilde{\rho_0})))_i  \leq -\widetilde{\rho_0}\alpha. \]
However, this is a contradiction to the termination criterion of the APPR algorithm~\eqref{eqn:appr}, and in particular, we must have $i\notin \supp(x^{\text{APPR}}(\widetilde{\rho_0}))$ and $\ph_i(\widetilde{\rho_0})>0$ by the condition~\eqref{eqn:appr_and_l1}.
Then, by Lemma~\ref{lem:nonneg_grad}, this implies that $(D^{-1}\nabla f(\ph(\widetilde{\rho_0})))_i=-\widetilde{\rho_0}\alpha$, and following the same steps as above we can see that $-\widetilde{\rho_0}\alpha >(D^{-1}\nabla f(x^{\text{APPR}}(\widetilde{\rho_0})))_i$. However, this again contradicts the termination criterion of APPR, thus proving the first part of the theorem.

Next, we prove the second part of the theorem, i.e., $\supp(x^{\text{APPR}}(\rho_0)) \subseteq \supp(\ph(\rho_1))$ for $\rho_1=(1-\alpha)\cdot \rho_0/2$. 
We first claim that for every node $i\in \supp(x^{\text{APPR}}(\rho_0))$, we must have $-\rho\alpha d_i < \nabla_i f(x^{\text{APPR}}(\rho_0))  \leq-\frac{1-\alpha}{2}\rho\alpha d_i$.
The first inequality is obvious from the termination criterion of APPR.
To see why the second inequality holds, if a node $i\in V$ is selected at iteration $k$ of the APPR algorithm, then by the update step~\eqref{eqn:appr}, we have $x^{(k+1)}_i=x_i^{(k)} -d_i^{-1}\nabla_i f(x^{(k)})$. Then,
the gradient of $f(x^{(k)})$ on node $i$ is updated as
\begin{align}\label{eqn:grad_update}
\nabla_i f(x^{(k+1)})&=\left[\left(\frac{1+\alpha}{2}D - \frac{1-\alpha}{2}A\right) x^{(k+1)}\right]_i -\alpha\seed_i \\
&= \frac{1+\alpha}{2}d_ix_i^{(k)} -\frac{1+\alpha}{2}\nabla_i f(x^{(k)}) - \frac{1-\alpha}{2}\sum_{j\sim i}w_{ji}x^{(k)}_j - \alpha\seed_i \nonumber
\\&= \nabla_i f(x^{(k)})  -\frac{1+\alpha}{2}\nabla_i f(x^{(k)})  = \frac{1-\alpha}{2}\nabla_i f(x^{(k)}) . \nonumber
\end{align}
By the termination criterion of APPR, we know that $\nabla_i f(x^{(k)})\leq -\rho_0\alpha d_i$ for all $i\in V$. This shows that the right-hand side of the above equation is $\leq -\rho_0\alpha d_i$, and in particular, taking $k\to\infty$, we obtain  $\nabla_i f(x^{\text{APPR}}(\rho_0))  \leq-\frac{1-\alpha}{2}\rho_0\alpha d_i$, which proves the claim.

Now consider the following strategy of continuously adding small mass to the vector $x^{\text{APPR}}(\rho_0)$: 
\begin{enumerate}
\item Start with $x^{\text{APPR}}(\rho_0)$. Find the node $i$ whose gradient is smallest, i.e., 
$$j_1\in \argmin_{i\in V} \nabla_i f(x^{\text{APPR}}(\rho_0)).$$
\item Add mass to the node $j_1$ until some other node $j_2\in V$ has  the gradient as small as  $j_1$. This event always happens because by the work above~\eqref{eqn:grad_update} we can see that the  gradient on node $j_1$ is increasing while the gradients on the other nodes are either decreasing (if it is a neighboring node of $j_1$) or stay the same.
\item Add mass to the nodes $(j_1,j_2)$ such that gradients on $j_1$ and $j_2$ are increasing at the same rate,\footnote{This is always possible since for any positive $\epsilon\in\R^{|V|}$, we know that $\sum_{i\in V}(\nabla_i f(x+\epsilon) - \nabla_i f(x))>0$. So, in order to increase the gradients at the same rate, we need to find a direction $\epsilon$ such that the summands $\nabla_j f(x+\epsilon) - \nabla_j f(x)$ are  same for nodes $j$ that are selected.} and until some other node $j_3\in V$ has  the gradient as small as  $j_1$ and $j_2$.
\item Continue this step until the gradient on the node $j_1$ becomes $-\frac{1-\alpha}{2}\rho_0\alpha d_i$. 
\end{enumerate}
Writing $\widetilde{x}^{\text{APPR}}$ to denote the output vector of the above procedure, it is easy to see that $\widetilde{x}^{\text{APPR}}$ indeed satisfies the optimality condition of the $\ell_1$-regularized PageRank minimization problem, given in Lemma~\ref{lem:nonneg_grad}, with $\rho=\rho_1$. This in turn shows that $\supp(x^{\text{APPR}})\subseteq \supp(\ph(\rho_1))$, since by the uniqueness of the optimal solution we have $\widetilde{x}^{\text{APPR}}=\ph(\rho_1)$. This completes the proof of Theorem~\ref{thm:appr_vs_l1}.

\subsection{Proof of Theorem~\ref{thm:full_recovery}}
\label{sec:proof_full_recovery}

The proof of Theorem~\ref{thm:full_recovery} is  a straightforward combination of Lemma~\ref{lem:reduced_full_recovery} and~\ref{lem:support_lemma}. 
Specifically, by Lemma~\ref{lem:reduced_full_recovery}, we know that for $\rho \leq \rho(\delta)$, we have that $\supp(\xhr)=K$. 
Then applying Lemma~\ref{lem:support_lemma} we have
\[\supp(\xhr) = K\subseteq \supp(\xh), \]
thus proving the result.

\subsection{Proof of Theorem~\ref{thm:bound_FP}}
\label{sec:proof_bound_FP}

To prove Theorem~\ref{thm:bound_FP}, we first need the following lemma for bounding the volume of $\supp(\xh)$. 
This lemma is a stronger version of~\citet[Theorem 2]{fountoulakis2015performance}.
\begin{lemma}
\label{lem:xh_vol_bound}
For any regularization parameter $\rho>0$, it holds that
\[\vol(\supp(\xh(\rho)))  \leq \frac{1-d^\top \xh(\rho)}{\rho} , \]
where $\vol(\supp(\xh))=\sum_{i\in \supp(\xh)} d_i$.
\end{lemma}
Now by our choice of $\rho\geq\rho(\delta)$~\eqref{eqn:rho} we have
\begin{align}
\label{eqn:vol_xhat}
\vol(\supp(\xh)) \leq \frac{1}{\rho}\leq \frac{1}{\rho(\delta)} &=  \left(\frac{1+\alpha}{1-\alpha}\right)^2  \left(\frac{1+\delta}{1-\delta} \right)^2 \frac{ (1+\delta) \bar{d}^2 }{\gamma p }\nonumber \\
&=  \left(\frac{1+\alpha}{1-\alpha}\right)^2  \left(\frac{1+\delta}{1-\delta} \right)^2 \frac{ (1+\delta)k \bar{d} }{\gamma^2 } , \end{align}
where the second step follows from~\eqref{eqn:gamma}. Furthermore, by Theorem~\ref{thm:full_recovery}, $\supp(\xh)$ contains the target cluster $K$, so the errors in $\supp(\xh)$ are solely attributed to the presence of false positives. Denoting by $\text{FP}$ the set of false positives in $\supp(\xh)$ we can write
\begin{equation}
\label{eqn:vol_xhat2} 
\vol(\supp(\xh)) = \vol(K) + \vol(\text{FP}).
\end{equation}
Since  $\vol(K)=\sum_{i\in K}d_i \geq (1-\delta) k\bar{d}$ by Lemma~\ref{lem:concen_degree}, it follows that
\begin{align*}
\vol(\text{FP}) &=  \vol(\supp(\xh)) -   \vol(K) \;\; \text{ by~\eqref{eqn:vol_xhat2}}  \\
&\leq \left(\frac{1+\alpha}{1-\alpha}\right)^2  \left(\frac{1+\delta}{1-\delta} \right)^2 \frac{(1+\delta)k\bar{d}}{\gamma^2 } -\vol(K)\;\; \text{ by~\eqref{eqn:vol_xhat}} \\
&\leq  \vol(K) \left[\left(\frac{1+\alpha}{1-\alpha}\right)^2  \left(\frac{1+\delta}{1-\delta} \right)^3 \frac{1}{\gamma^2 } - 1\right],
\end{align*}
with probability at least $1-2\exp(-c_0\delta^2 \bar{d})$. This proves the theorem.


\subsection{Proof of Theorem~\ref{thm:no_false_pos}}
\label{sec:proof_no_false_pos}

Under Assumption~\ref{assump:rate}, with nonvanishing probability we can find a good seed node in the target cluster that is connected solely  to $K$.
\begin{lemma}
\label{lem:seed_node}
Suppose that Assumption~\ref{assump:rate} holds, i.e., $q= \frac{c}{n}$ for a fixed constant $c>0$.\ Under the local random model given in Definition~\ref{def:model}, if $n$ is sufficiently large, then 
\[ \PP{\text{There exists a node $i\in K$ such that $i$ is not connected to $K^c$}}  \geq 1- (1-\exp(-1.5c))^k.\]
\end{lemma}
\noindent We select this node as a seed node in the $\ell_1$-regularized PageRank problem.\

Next we show that under the conditions of Theorem~\ref{thm:no_false_pos}, the optimal solution to the reduced problem, $\xhr=\xhr(\rho(\delta))$ in~\eqref{eqn:reduced_pagerank_pen},  obeys
\begin{equation}
\label{eqn:opt_cond_Kc} 
|(Q\xhr - \alpha \seed)_j|  = |- \frac{1-\alpha}{2} \cdot A_{j,K} \xhr | < \rho(\delta)\alpha {d}_j \text{ for $j\in K^c$}.  
\end{equation}
Note that the above inequality is simply the optimality condition for  the full-dimensional problem on $K^c$ (see Lemma~\ref{lem:nonneg_grad}). Since, by the optimality conditions for the reduced problem, $\xhr$ also satisfies the full-dimensional optimality condition on $K$. By uniqueness of the solution this means that $\xhr=\xh$. Also, since $\supp(\xhr)\subseteq K$ by construction, this proves that there is no false positive in $\xh$.

Now write $A_{j,K} \xhr$ into the sum of two terms,
\[A_{j,K} \xhr = A_{j,S} \xhr_{S} + A_{j,K\backslash S}\xhr_{K\backslash S}.  \]
The first term can be ignored, since by Lemma~\ref{lem:seed_node} we choose the seed node to be the one that is solely connected to $K$. For the second term we use H\"older's inequality to bound
\[A_{j,K\backslash S}\xhr_{K\backslash S} \leq \norm{A_{j,K\backslash S}}_1\norm{\xhr_{K\backslash S}}_\infty. \]
Applying Lemma~\ref{lem:concen_degree_K},~\ref{lem:l_infty_bound}, and using the fact that $u\leq \frac{2\alpha}{(1+\alpha)\bar{d}}, v\leq \frac{(1-\alpha)p}{(1+\alpha)\bar{d}^2}$ (which can be deduced from Lemma~\ref{lem:exact_recovery}), and that $\delta\leq 0.1$ while $\alpha \in [0.1,0.9]$  (by assumptions of  the theorem), then
\begin{align*}
\Abs{(Q\xhr- \alpha \seed)_j}  \leq \frac{1-\alpha}{2}\norm{A_{j,K\backslash S}}_1\norm{\xhr_{K\backslash S}}_\infty  &\leq (0.5 c+1)\left[ \frac{c_1(1-\alpha)u}{\bar{d}}+(1+2c_1\delta)v+2c_1\delta \cdot \rho(\delta)\alpha \right] \\
& \leq (0.5 c+1)C_1\left(\frac{2}{\bar{d}^2}+\rho(\delta)\alpha \right),
\end{align*}
where $C_1>0$ is a positive constant. Then, to prove~\eqref{eqn:opt_cond_Kc}, it suffices to show 
\[ (0.5c + 1)C_1\left[\frac{2}{\rho(\delta)\alpha \bar{d}^2}+1 \right] < d_j.\]
Plugging in the definition of $\rho(\delta)$~\eqref{eqn:rho} to the above, and using $\alpha\in [0.1,0.9]$ and $\delta\leq 0.1$, we obtain
\begin{align*}
 (0.5c + 1)C_1\left[\frac{2}{\rho(\delta)\alpha \bar{d}^2}+1 \right] &= (0.5c + 1)C_1\left[ \left(\frac{1+\alpha}{1-\alpha}\right)^2\left(\frac{1+\delta}{1-\delta}\right)^2 \frac{1+\delta}{\alpha\gamma p} +1 \right] \\&\leq \frac{ (0.5c + 1)C_2}{\gamma p} ,
\end{align*}
for some constant $C_2>0$. By the assumption~\eqref{eqn:degree_cond} this means that $\rho(\delta)^{-1}\alpha^{-1}|(Q\xhr - \alpha \seed)_j|  < d_j$, and thus we have proved the  claim~\eqref{eqn:opt_cond_Kc}.

\subsection{Proof of Theorem~\ref{thm:appr_thm}}
\label{sec:proof_appr_thm}
First, by Theorem~\ref{thm:appr_vs_l1}, we know that 
\[\supp(\ph(\rho(\delta))) \subseteq \supp(x^{\text{APPR}}(\rho(\delta))) \subseteq \supp(\ph((1-\alpha)\cdot \rho(\delta)/2)) .\]
Since $\xh(\rho(\delta))$ contains the target cluster by Theorem~\ref{thm:full_recovery}, therefore together with the relation above we obtain $K\subseteq \supp(x^{\text{APPR}})$. To see that the false positives in $x^{\text{APPR}}$ are also bounded, we can follow the same step as the proof of Theorem~\ref{thm:bound_FP} (see Section~\ref{sec:proof_bound_FP}) to show that 
\[\vol(\text{FP}(\ph((1-\alpha)\cdot \rho(\delta)/2))) \leq \vol(K) \left[ {\left(\frac{1+\alpha}{1-\alpha}\right)^2  \left(\frac{1+\delta}{1-\delta} \right)^3 \frac{2}{(1-\alpha)\gamma^2 }  -1} \right].\]
From the relation between $\supp(x^{\text{APPR}}(\rho(\delta)))$ and $\supp(\ph((1-\alpha)\cdot \rho(\delta)/2))$ it directly follows that 
\[\vol(\text{FP}(x^{\text{APPR}}))  \leq \vol(K) \left[ {\left(\frac{1+\alpha}{1-\alpha}\right)^2  \left(\frac{1+\delta}{1-\delta} \right)^3 \frac{2}{(1-\alpha)\gamma^2 }  -1} \right].\]
Finally to prove the exact recovery case, it suffices to show that $\supp(\ph((1-\alpha)\cdot \rho(\delta)/2))$ contains no false positives. In order for this all we need to do is to replace $\rho(\delta)$ by $\rho((1-\alpha)/2 \cdot \delta)$ in the proof of Theorem~\ref{thm:no_false_pos}. 
Then the same proof steps go through and we finally get the condition
\[\frac{2C(0.5c+1)}{(1-\alpha)\gamma p}=\O{\frac{1}{\gamma p}} < d_j, \]
in place of~\eqref{eqn:degree_cond}. This completes the proof of the theorem.

\section{Proofs of lemmas}\label{sec:additional_proof}
We prove all of our lemmas in this section.

\subsection{Proof of Lemma~\ref{lem:nonneg}}
\label{sec:proof_nonneg}

	By the KKT condition $\xh$ must satisfy
	\[\nabla_j \fp(\xh) =-\rho\alpha \partial  \norm{D\xh}_1 \text{ for all $j\in V$} ,\]
	or equivalently,
	\begin{equation}\label{eqn:kkt_pre} (Q \xh)_j -\alpha \seed{}_j = \begin{cases} -\rho\alpha d_j, & x_j >0 , \\
	\rho\alpha d_j, & x_j <0, \\ [ -\rho\alpha d_j,  \rho\alpha d_j], & x_j = 0,\end{cases}  \end{equation}
	where $d=\diag(D)$ is the degree vector of the graph. Simple calculation shows that
	\begin{align}\label{eqn:grad_of_g} (Q \xh)_j &= \frac{(1+\alpha)}{2}\left(d_j \xh_j - \frac{1-\alpha}{1+\alpha}\cdot \sum_{i:(i,j)\in E} \xh_i \right) \nonumber\\&= \frac{(1+\alpha)}{2}\left(\sum_{i:(i,j)\in E} \xh_j - \frac{1-\alpha}{1+\alpha}\cdot \sum_{i:(i,j)\in E} \xh_i \right).  \end{align}

	Now assume  $j^\star\in V$ is a node such that $\xh_{j^\star} < 0$ and that $j^\star=\argmin_{j\in V} \xh_j$.\ Then from the condition~\eqref{eqn:kkt_pre}, it must be that $(Q\xh)_{j^\star} > 0 $. However by~\eqref{eqn:grad_of_g} this is only possible if there is at least one node $i\in V$ with $i\sim j^\star$ such that
	\[\xh_i < \frac{1+\alpha}{1-\alpha}\cdot \xh_{j^\star} . \]
	However this means that $\xh_i$ is smaller than $\xh_{j^\star}$, contradicting to the fact that $j^\star\in V$ is the smallest  node in the graph.\ This proves the desired result.

\subsection{Proof of Lemma~\ref{lem:nonneg_grad}}
This lemma follows from Lemma~\ref{lem:nonneg} and the definition of $Q$.

\subsection{Proof of Lemma~\ref{lem:monotone}}
\label{sec:proof_monotone}

First, by~\citet[Proposition 1]{rosset2007piecewise}, we know that the optimal solution path $\xh(\rho)$ for the $\ell_1$-regularized minimization~\eqref{eqn:pagerank2} is piecewise linear as a function of $\rho>0$. In particular, this implies that the path is  continuous.

Next, we prove that if the support set of $\xh(\rho)$ remains constant in the interval $[\rho_1,\rho_0]$ ($\rho_0 >\rho_1 $), then $\xh(\rho)$ is strictly decreasing on  $[\rho_1,\rho_0]$, i.e., $\xh(\rho_1) > \xh(\rho_0) $, where the inequality applies component-wise. This, together with the non-negativity of the solution (Lemma \ref{lem:nonneg}) and the  continuity of the path $\xh(\rho)$, then establishes the lemma.

Write $\Acal(\rho)=\{j: \xh_j(\rho) > 0\} $ for $\rho>0$ and choose $\rho_0 ,\rho_1>0$ with $\rho_0>\rho_1$ such that the support set $\Acal(\rho)=\Acal$ does not change for $\rho \in [\rho_1, \rho_0]$.  For $j\in\Acal$, by Lemma~\ref{lem:nonneg_grad}, we have that
	\[ (Q \cdot \xh(\rho))_j - \alpha \seed{}_j= -\rho \alpha d_j, \]
    and so
	\[ Q_{\Acal,\Acal} (\xh_\Acal(\rho_1) - \xh_\Acal(\rho_0)) = (\rho_0 - \rho_1) \alpha d_\Acal. \]
	Multiplying  $Q_{\Acal,\Acal}^{-1}$ on both sides (note that $Q_{\Acal,\Acal}$ is positive definite),
	\begin{equation}
        \label{eqn:monotone} \xh_\Acal(\rho_1) - \xh_\Acal(\rho_0) = \alpha (\rho_0 - \rho_1) Q_{\Acal,\Acal}^{-1}  d_\Acal  .
        \end{equation}
	Now write
	\begin{align*}Q_{\Acal,\Acal} = \frac{1+\alpha}{2}\cdot\left( D - \frac{1-\alpha}{1+\alpha} \cdot A \right)_{\Acal,\Acal}
	= \frac{1+\alpha}{2}\cdot D^{1/2}_{\Acal,\Acal} \left(\ident - \frac{1-\alpha}{1+\alpha} (D^{-1/2}AD^{-1/2})_{\Acal,\Acal} \right) D^{1/2}_{\Acal,\Acal}.
	\end{align*}
	For all $z\in\R^{\Abs{\Acal}}$, we have that
	\begin{align*}
	z^\top z  \pm z^\top  (D^{-1/2}AD^{-1/2})_{\Acal,\Acal} z  &= \sum_{i\in \Acal}z_i^2 \pm \sum_{(i,j)\in E, i,j\in\Acal} \frac{2w_{ij}z_i z_j}{\sqrt{d_i d_j}} \\
	&>  \sum_{(i,j)\in E, i,j\in\Acal}w_{ij}  \left(\frac{z_i}{\sqrt{d_i}} {\pm} \frac{z_j}{\sqrt{d_j}} \right)^2\geq 0,
	\end{align*}
	where the second step holds since $d_i > \sum_{j:(i,j)\in E, j\in\Acal} w_{ij}$. This shows that all the eigenvalues of $(D^{-1/2}AD^{-1/2})_{\Acal,\Acal} $ have absolute value less than $1$. Using the previous fact,  $Q_{\Acal,\Acal}^{-1}$ can be written~as
	\begin{align}
        \label{eqn:Q_inv}
	Q_{\Acal,\Acal}^{-1} &= 2(1+\alpha)^{-1}\cdot  D^{-1/2}_{\Acal,\Acal} \left[I - \frac{1-\alpha}{1+\alpha}\cdot (D^{-1/2}AD^{-1/2})_{\Acal,\Acal} \right]^{-1 }D^{-1/2}_{\Acal,\Acal} \nonumber \\\nonumber
	&=  2(1+\alpha)^{-1}\cdot  D^{-1/2}_{\Acal,\Acal} \left[I +  \frac{1-\alpha}{1+\alpha} \cdot (D^{-1/2}AD^{-1/2})_{\Acal,\Acal} +   \left(\frac{1-\alpha}{1+\alpha}\right)^2\cdot (D^{-1/2}AD^{-1/2})_{\Acal,\Acal}^2 \right. \\
	&\qquad\qquad\qquad\qquad\qquad\left. + \cdots \vphantom{I +  \frac{1-\alpha}{1+\alpha} \cdot (D^{-1/2}AD^{-1/2})_{\Acal,\Acal} +  \left(\frac{1-\alpha}{1+\alpha}\right)^2\cdot (D^{-1/2}AD^{-1/2})_{\Acal,\Acal}^2} \right] D^{-1/2}_{\Acal,\Acal}.
	\end{align}
	Now it is easy to see that the right hand side in~\eqref{eqn:monotone} is positive, i.e., $\alpha (\rho_0 - \rho_1) Q_{\Acal,\Acal}^{-1}  d_\Acal>0$. This completes the proof of Lemma~\ref{lem:monotone}.

\subsection{Proof of Lemma~\ref{lem:concen}}

This lemma is proved in~\citet[Proposition 2.4.1]{vershynin2018high} which we reproduce here for completeness.
By Chernoff's inequality (\citep[Theorem 2.3.1]{vershynin2018high}), for some constant $c_0>0$,
\[\PP{\Abs{X_i - \mu } \geq \delta \mu} \leq 2e^{-2c_0\delta^2 \mu}. \]
Using union bound,
\[\PP{\max_{i\in K}\Abs{X_i - \mu } \geq \delta\bar{d} } \leq 2e^{-2c_0\delta^2\mu+ \log k}. \]
Since $\log k\leq c_0\delta^2\mu$ by assumption, the result  follows.

\subsection{Proof of Lemma~\ref{lem:concen_degree}}

This lemma follows directly from Lemma~\ref{lem:concen}.

\subsection{Proof of Lemma~\ref{lem:concen_degree_K}}

We have $\norm{A_{j,K}}_1\sim \text{Binom}(k, q)$ for $j\in K^c$, so using tail bound for Binomial distribution (\citep{arratia1989tutorial}), for any $t > c$,
\begin{align*} \PP{\norm{A_{j,K}}_1 \geq t  } &\leq \exp\left[-t\log \left(\frac{ t}{k \cdot q } \right) - k\left(1-\frac{ t}{k} \right)\log \left( \frac{1-t/k}{1-q}\right)  \right].\end{align*}
Set $t=c+3$. Then, since $\log(1-x) \geq -x$ for all $0<x<\frac{1}{2}$, then
\[- k\left(1-\frac{t}{k} \right)\log \left( \frac{1-t/k}{1-q}\right)  \leq  - k\log(1-t/k)-t\log (1-q)\leq t + \log2\cdot t \leq 2t=2(c+3), \]
where the second inequality follows, since $k\geq 2(c+3)$ by assumption, and  that $q\leq 0.5$ for $n$ sufficiently large. Then using union bound,
\begin{align*} 
\PP{\max_{j\in K_1^c}\norm{A_{j,K_1}}_1  \geq c+3 } & \leq \exp\left[-(c+3)\log \left(\frac{(c+3)n}{c\cdot k} \right)   + 2(c+3)+ \log(n-k)\right] \\ 
                                                                                    &   \leq \O{n^{-1}},
\end{align*}
as long as $n\gg k$.


\subsection{Proof of Lemma~\ref{lem:exact_recovery}}
\label{sec:proof_exact_recovery}

First the results in Lemma~\ref{lem:nonneg} and Lemma~\ref{lem:nonneg_grad} hold for any graph, seed vector $\seed$, and $\rho>0$, so the result can also be applied to~\eqref{eqn:l1_pagerank_exp}. In particular, the solution $\xs$ is non-negative, and thus the optimality condition  can be expressed as
\begin{equation}\label{eqn:opt_cond_expected} (\EE{Q} \xs)_j -   \alpha \ones_{j\in S} = \begin{cases} -\rho \alpha \EE{d_j} & \xs_j >0 , \\ [-\rho \alpha \EE{d_j}, 0] & \xs_j = 0.\end{cases}  \end{equation}
(Note $\seed$ is $1$ on the seed node and $0$ on the rest.)
For $\rho> \frac{1}{\bar{d}}$ the optimal solution \eqref{eqn:l1_pagerank_exp} is the zero vector. For $\rho \le \frac{1}{\bar{d}}$ the first nonzero nodes appear.
From~\eqref{eqn:opt_cond_expected}, as $\rho$ decreases less than or equal to $\frac{1}{\bar{d}}$, we can see that the  seed node becomes active at first.

Now let $\rho^\natural>0$ be the regularization parameter for which the next set of nodes enters the active set. Then, for $ \rho \in (\rho^\natural ,\frac{1}{\bar{d}}]$,  we know that $\xs_S >0$ and $\xs_{S^c}=0$. Writing  $\xs_S =u \ones_S$ for some $u>0$,  we can see that for the seed node $j\in S$,
\begin{align*}(\EE{Q} \xs)_j &= \frac{1+\alpha}{2}\cdot \left[\sum_{i:(i,j)\in E} \EE{A_{ij}} \xs_j - \frac{1-\alpha}{1+\alpha}\sum_{i:(i,j)\in E} \EE{A_{ij}} \xs_i \right] = \frac{1+\alpha}{2}\cdot u \cdot \bar{d}.\end{align*}
Substituting in the optimality condition~\eqref{eqn:opt_cond_expected}, and solving with respect to $u$, we get
\begin{equation*}
\label{eqn:seed_nodes}
 \frac{1+\alpha}{2}\cdot u \cdot \bar{d} -\alpha = -\rho \alpha\bar{d}, \; \text{ and so }\; u = \frac{2(\alpha - \rho \alpha \bar{d})}{(1+\alpha)\bar{d}}.  \end{equation*}
It remains to be shown that $\xs_S =u \ones_S$, for the above $u$, satisfies the optimality conditions for $j\in S^c$.  To verify this, we use the definition of $u$ and the optimality conditions for $j\in S^c$. We need
\begin{equation}
\label{eq:optexp_u}
(\EE{Q} \xs)_j \in [-\rho  \alpha \EE{d_j}, 0] \ \text{ for all } j\in S^c.
\end{equation}
We have that
$$
(\EE{Q} \xs)_j =-\frac{1-\alpha}{2}p\cdot  u = -\left(\alpha - \rho \alpha \bar{d}\right) \frac{p(1-\alpha)}{\bar{d}(1+\alpha)} \text{ for all } j \in K\backslash S,
$$
and
$$
(\EE{Q} \xs)_j =-\frac{1-\alpha}{2}q\cdot  u = -\left(\alpha - \rho \alpha \bar{d}\right) \frac{q(1-\alpha)}{\bar{d}(1+\alpha)} \text{ for all } j \in K^c.
$$
Using the above equalities, we get that the optimality conditions are satisfied if and only if
$$
\rho > \rho^\natural:=\frac{(1-\alpha)p}{\bar{d} \left[ (1+\alpha)\bar{d} +(1-\alpha)p \right]}.
$$

Next, we prove that, when $\rho$ reaches $\rho^\natural$, the nodes in $K \backslash S$ appear in the active set. To see this, from~\eqref{eqn:opt_cond_expected}, we can check that the first $\rho$ value allowing nonzero nodes for $K\backslash S$  is given by
\[ -\frac{ 1-\alpha }{2} p \cdot u= -\rho \alpha \bar{d} \;\text{, and so } \; \rho=\rho^\natural = \frac{(1-\alpha)p}{\bar{d} \left[ (1+\alpha)\bar{d} +(1-\alpha)p \right]}. \]
Now assuming that $\xs$ takes the form of $u \ones_S + v \ones_{K}$, and substituting in the optimality condition~\eqref{eqn:opt_cond_expected}, we obtain the following equations:
\begin{align*}
&j\in S:\;\;  \underbrace{\frac{1+\alpha}{2}\left[ (u+v)\cdot \bar{d}- \frac{1-\alpha}{1+\alpha}\cdot p\cdot v(k-1) \right] }_{=(\EE{Q}\xs)_j }-  \alpha = - \rho \alpha \bar{d}, \\
&j\in K\backslash S: \;\;   \underbrace{\frac{1+\alpha}{2}\left[ v\cdot \bar{d} - \frac{1-\alpha}{1+\alpha}\cdot p\cdot (u+v(k-1)) \right] }_{= (\EE{Q}\xs)_j}= - \rho\alpha \bar{d}.
  \end{align*}
 So, solving with respect to $u$ and $v$, we obtain
 \begin{align}\label{eqn:u_and_v}
 \begin{cases}u=\frac{2\alpha}{\left[(1+\alpha)\bar{d} +(1-\alpha)p \right]}, \\ v=\frac{ \frac{1-\alpha}{2}p   u -  \rho\alpha \bar{d}}{\alpha\bar{d} + \frac{1-\alpha}{2}q(n-k)}. \end{cases}
 \end{align}
Also, $\xs = u \ones_S + v \ones_{K}$, with $u$ and $v$ given in~\eqref{eqn:u_and_v}, satisfies the optimality conditions for $K^c$ if and only if
 \[ \rho > \rho^\sharp := \frac{q(1-\alpha)}{2\alpha\bar{d}\cdot \EE{{d}_{\ell_\sharp}} +q(1-\alpha)\left[ k\bar{d} + (n-k) \EE{{d}_{\ell_\sharp}}\right]},\]
 where $\EE{{d}_{\ell_\sharp}}=\min_{\ell\in K^c}\EE{d_\ell}$. We claim that $\rho^\natural > \rho^\sharp$, in which case we have proved that $\xs=u \ones_S + v \ones_{K}$ satisfies the optimality condition~\eqref{eqn:opt_cond_expected} for $\rho \in [\rho^\sharp,\rho^\natural)$, which proves the theorem.

It now remains to check that $\rho^\natural > \rho^\sharp$. With some algebra, we have that
\begin{align*}
\rho^\natural > \rho^\sharp \;\;  &\Longleftrightarrow \;\; 2p\alpha\cdot \bar{d}\cdot \EE{{d}_{\ell_\sharp}} + pq(1-\alpha)\cdot k\bar{d} + pq(1-\alpha)(n-k)\EE{{d}_{\ell_\sharp}} \\ 
                                                 & > q(1+\alpha)\bar{d}^2 + pq(1-\alpha) \bar{d}\\
&\Longleftrightarrow \;\; 2p\alpha\cdot \bar{d}\cdot \EE{{d}_{\ell_\sharp}} + pq(1-\alpha)\cdot k\bar{d} + pq(1-\alpha)(n-k)\EE{{d}_{\ell_\sharp}}  \\ 
& > q(1-\alpha+2\alpha)\bar{d}^2 + pq(1-\alpha) \bar{d}\\
&\Longleftrightarrow \;\; 2\alpha \bar{d}(p \EE{{d}_{\ell_\sharp}} - q\bar{d}) + pq(1-\alpha)(k-1)\bar{d} + pq(1-\alpha)(n-k)\EE{{d}_{\ell_\sharp}}  \\ 
& > q(1-\alpha)\bar{d}^2.
\end{align*}
Also, since $\bar{d}=p(k-1)+q(n-k)$, we have
\[q(1-\alpha)\bar{d}^2 = q(1-\alpha)\bar{d} \left[p(k-1)+q(n-k) \right] ,\]
and so
\begin{equation}\label{eqn:rho_cond}\rho^\natural > \rho^\sharp \;\; \Longleftrightarrow  \;\; 2\alpha \bar{d}(p \EE{{d}_{\ell_\sharp}} - q\bar{d})  + (1-\alpha)q(n-k)(p\EE{{d}_{\ell_\sharp}} - q\bar{d}) >0. \end{equation}
Under Assumption~\ref{assump:rate}, we know that $p\EE{d_{\ell_\sharp}} =\O{1}$ while $q\bar{d} = \O{\frac{\bar{d}}{n}}$, so $p \EE{{d}_{\ell_\sharp}} - q\bar{d}>0$ for a sufficiently large $n$. This verifies~\eqref{eqn:rho_cond}.

\subsection{Proof of Lemma~\ref{lem:reduced_full_recovery}}
Define
\[\xcr(\rho) = \alpha Q_{K,K}^{-1}( \seed_K -\rho d_K).\]
We will prove that $\xcr=\xcr(\rho) >0$ for any $\rho\leq \rho(\delta)$ where $\rho(\delta)$ is defined in~\eqref{eqn:rho}. In particular, this means that $\xcr$ satisfies the optimality condition for the reduced problem~\eqref{eqn:reduced_pagerank_pen}, implying that $\xhr_K=\xcr>0$ and thus proving the result.

First, using~\eqref{eqn:Q_inv}, we can write
\[\xcr=  \frac{2\alpha}{1+\alpha}D_{K,K}^{-1}\left[\ident + \frac{1-\alpha}{1+\alpha}A_{K,K}D_{K,K}^{-1} + \left(\frac{1-\alpha}{1+\alpha}\right)^2 (A_{K,K}D_{K,K}^{-1})^2 + \cdots  \right] (\seed_K - \rho d_K). \]
Let $w=D_{K,K}^{-1}\seed_K$. Then
\begin{align}\label{eqn:x_check}\xcr &=  \frac{2\alpha}{1+\alpha} \Big[\sum_{j=0}^{\infty} \left( \frac{1-\alpha}{1+\alpha}\right)^j (D_{K,K}^{-1} A_{K,K})^j w - \rho \sum_{j=0}^{\infty}  \left( \frac{1-\alpha}{1+\alpha}\right)^{j}(D_{K,K}^{-1} A_{K,K})^j \ones \Big] \nonumber \\
&=\frac{2\alpha}{1+\alpha} \Big[w +\frac{1-\alpha}{1+\alpha} D_{K,K}^{-1}A_{K,K}w+  \sum_{j=0}^{\infty} \left( \frac{1-\alpha}{1+\alpha}\right)^{j+2} (D_{K,K}^{-1} A_{K,K})^{j+2} w \nonumber \\ 
& - \rho \sum_{j=0}^{\infty}  \left( \frac{1-\alpha}{1+\alpha}\right)^{j}(D_{K,K}^{-1} A_{K,K})^j \ones \Big].
\end{align}
Note that $\seed_K$ has mass $1$ on the seed node, so we have $w=\seed_K / d_S$, where $d_S$ is the degree of the seed node. Denoting by $\text{FON}$ the first-order neighbors of the seed node in $K$, i.e., {$\text{FON}=\{i\in K: \text{$i$ is a neighbor of the seed node} \}$}, we then have
\[D_{K,K}^{-1}A_{K,K}w = D_{K,K}^{-1}\ones_{\text{FON}}/d_S ,\]
where $\ones_{\text{FON}}$ is the indicator vector for the set $\text{FON}$. Applying the degree concentration Lemma~\ref{lem:concen_degree},  then with probability at least $1-2e^{-c_0 \delta^2 \bar{d}}$,
\begin{equation}\label{eqn:full_recovery_ineq1}D_{K,K}^{-1}A_{K,K}w\geq \frac{1}{(1+\delta)^2\bar{d}^2}\ones_{\text{FON}}. \end{equation}

Next, using Lemma~\ref{lem:concen}, we have that $|\text{FON}|\geq (1-\delta) pk$ with probability at least $1-2e^{-c_0 \delta^2 p k}$. Furthermore, for each non-seed node  $i\notin \text{FON}$ in $K$, it is connected to  nodes in $\text{FON}$ with probability $p$, independently of all other pairs. Then $A_{i,K}^\top \ones_{\text{FON}}  |\; |\text{FON}| \iidsim \text{Binom}(|\text{FON}|,p) $, and thus applying Lemma~\ref{lem:concen}, and using the assumption $(1-\delta)p^2k \geq c_0^{-1}\delta^{-2}\log k$, we get
\begin{align*}\PPst{A_{i,K}^\top \ones_{\text{FON}} \geq (1-\delta)|\text{FON}| p \text{ for all $i\notin \text{FON}$} }{|\text{FON}|} &\geq  1-2e^{-c_0 \delta^2 |\text{FON}|p} \\
&\geq  1-2e^{-c_0 \delta^2 (1-\delta)p^2k} .\end{align*}
Hence, we can conclude that with probability at least $1-4e^{-c_0\delta^2(1-\delta)p^2 k }$,
\begin{equation}\label{eqn:full_recovery_ineq2} A_{i,K}^\top \ones_{\text{FON}}\geq (1-\delta)^2 p^2 k \text{ for all $i\notin \text{FON}$}.\end{equation}
Then,
\begin{align*}
\left(\frac{1-\alpha}{1+\alpha}\right)^2 (D_{K,K}^{-1}A_{K,K})^2 w &\geq \left(\frac{1-\alpha}{1+\alpha}\right)^2 (D_{K,K}^{-1}A_{K,K})\cdot \frac{1}{(1+\delta)^2\bar{d}^2}\ones_{\text{FON}}\\
&\geq \left(\frac{1-\alpha}{1+\alpha}\right)^2 \left(\frac{1-\delta}{1+\delta}\right)^2 \frac{p^2 k}{\bar{d}^2} D_{K,K}^{-1} \ones\\
&\geq \left(\frac{1-\alpha}{1+\alpha}\right)^2  \left(\frac{1-\delta}{1+\delta} \right)^2 \frac{\gamma p}{(1+\delta) \bar{d}^2}\ones,
\end{align*}
where the first step applies~\eqref{eqn:full_recovery_ineq1}, the second step uses~\eqref{eqn:full_recovery_ineq2}, and the third applies Lemma~\ref{lem:concen_degree} and the fact that $p\cdot  k = \gamma\bar{d}$. Returning to~\eqref{eqn:x_check}, we can now see that $\xcr>0$ as long as $\rho$ is less than $\rho(\delta)=\left(\frac{1-\alpha}{1+\alpha}\right)^2  \left(\frac{1-\delta}{1+\delta} \right)^2 \frac{\gamma p}{(1+\delta) \bar{d}^2}$. This completes the proof of Lemma~\ref{lem:reduced_full_recovery}.


\subsection{Proof of Lemma~\ref{lem:support_lemma}}

Recall that $\xh$ is the $\ell_1$-regularized PageRank on the original graph~\eqref{eqn:pagerank2}. If $\rho>1/d_S$, then we can easily check that $\xh=\xhr=0$, while for $\rho= 1/d_S$, we have $\supp(\xhr) = \supp(\xh)=S$. To prove that $\supp(\xhr) \subseteq \supp(\xh)$ holds for $\rho<1/d_S$, we proceed by induction.

Writing $\Acal_1=\supp(\xhr)$ and $\Acal_2 = \supp(\xh)$, by the optimality condition, we have
\[ \xhr_{\Acal_1} = \alpha Q_{\Acal_1,\Acal_1}^{-1}(\seed_{\Acal_1} - \rho d_{\Acal_1}) \text{ and } \xh_{\Acal_2} = \alpha Q_{\Acal_2,\Acal_2}^{-1}(\seed_{\Acal_2} - \rho d_{\Acal_2}). \]
By inductive hypothesis, assume $\Acal_1\subseteq \Acal_2$. According to the expression~\eqref{eqn:Q_inv}, then we can see that $\xh_{\Acal_1} \geq \xhr_{\Acal_1}$, where the inequality applies component-wise. Now let $i\in K$ be a node such that $\xhr_i=\xh_i=0$.  Using the optimality condition, we know that $i$ becomes active for the full-dimensional problem~\eqref{eqn:pagerank2} whenever the following condition is satisfied:
\begin{equation}\label{eqn:support1}-\frac{1-\alpha}{2}\sum_{j\sim i, j\in \Acal_2} w_{ij} \xh_j = -\rho\alpha d_i. \end{equation}
Analogously, the node $i$ becomes active for the reduced problem~\eqref{eqn:reduced_pagerank_pen} when the following condition is satisfied:
\begin{equation}\label{eqn:support2}-\frac{1-\alpha}{2}\sum_{j\sim i, j\in \Acal_1} w_{ij} \xhr_j = -\rho\alpha d_i. \end{equation}
Comparing the left-hand sides in~\eqref{eqn:support1} and~\eqref{eqn:support2}, it is obvious that under the induction assumption, the left-hand side in~\eqref{eqn:support1} is larger than in~\eqref{eqn:support2}. This implies that $\xh_i$ becomes active earlier than $\xhr_i$, so the induction assumption continues to hold. 
This completes the proof of the lemma.

\subsection{Proof of Lemma~\ref{lem:l_infty_bound}}
\label{sec:proof_l_infty_bound}


Since $\xhr=(\xhr_S,\xhr_{K\backslash S})$ is the solution to the reduced problem~\eqref{eqn:reduced_pagerank_pen}, fixing $\xhr_S$, then $\xhr_{K\backslash S}$ is the minimizer of the following optimization problem:
\begin{equation}\label{eqn:reduced_pagerank_pen2} \xhr_{K\backslash S} = \argmin_{y\in\R^{k-1}} \left\{\frac{1}{2} (\xhr_S , y)^\top Q_{K,K} (\xhr_S,y)+\rho\alpha \norm{D\cdot (\xhr_S,y)}_1 \right\}. \end{equation}
Due to Lemma~\ref{lem:reduced_full_recovery}, the subgradient of $\norm{D\cdot (\xhr_S,y)}_1 $ at $y=\xhr_{K\backslash S}$ is given by $d_{K\backslash S}$. 
Using optimality, it follows that
\[ 0=\underbrace{Q_{K\backslash S,K}\xhr }_{\text{gradient at $y=\xhr_{K\backslash S}$}}  + \rho\alpha d_{K\backslash S}.\]

Next, we can prove that our choice of $\rho=\rho(\delta)$ lies in the interval $ [\rho^\sharp, \rho^\natural)$.
\begin{lemma}
\label{lem:rho}
Under the conditions of Lemma~\ref{lem:l_infty_bound}, we have $\rho=\rho(\delta) \in (\rho^\sharp, \rho^\natural)$.
\end{lemma}

Hence, by Lemma~\ref{lem:exact_recovery},  the population version of $\ell_1$-regularized PageRank~\eqref{eqn:l1_pagerank_exp} has support $K$,  so it follows that $\xs$ is also the solution to the reduced problem
\begin{equation}
\label{eqn:l1_pagerank_exp2}
\xs=\min_{x}\left\{\frac{1}{2}x^\top \EE{Q_{K,K}}x - \alpha x^\top \seed_K + \rho\alpha \norm{\EE{D_{K,K}}x}_1: x_{K^c}=0\right\} .
\end{equation}
(Abusing notation, we use $\xs$ and $\xs_{K}$ interchangeably throughout the proof.) So, using the optimality condition,  we obtain
\[ 0=\underbrace{\EE{Q_{K\backslash S,K}} \xs}_{\text{gradient at $y=\xs_{K\backslash S}$}}  + \rho\alpha \EE{d_{K\backslash S}},\]
where we have $\partial \norm{\EE{D}\cdot (\xs_S,y)}_1 =\EE{d_{K\backslash S}}$ due to Lemma~\ref{lem:exact_recovery} and Lemma~\ref{lem:rho}. 
Combining the two optimality equations, and adding and subtracting $Q_{K\backslash S,K}\xs$, it follows that
\begin{align*}0 &= \EE{Q_{K\backslash S,K}}\xs  - Q_{K\backslash S,K}\xhr  +    \rho\alpha \EE{d_{K\backslash S}} - \rho\alpha d_{K\backslash S} \\
&=  (\EE{Q_{K\backslash S,K}}- Q_{K\backslash S,K})\xs + Q_{K\backslash S,K}(\xs-\xhr)+    \rho\alpha ( \EE{d_{K\backslash S}} - d_{K\backslash S}).
 \end{align*}
Writing $Q_{K\backslash S,K}(\xs-\xhr) =Q_{K\backslash S,S}(\xs_S-\xhr_S) + Q_{K\backslash S,K\backslash S}(\xs_{K\backslash S}-\xhr_{K\backslash S})$, and rearranging terms, we get
\begin{align*}  
Q_{K\backslash S,K\backslash S}(\xhr_{K\backslash S}-\xs_{K\backslash S}) 
   & = (\EE{Q_{K\backslash S,K}}- Q_{K\backslash S,K})\xs + Q_{K\backslash S,S}(\xs_S-\xhr_S) \\ & \qquad +    \rho\alpha  (\EE{d_{K\backslash S}}  - d_{K\backslash S} ). 
\end{align*}
Taking the $\ell_\infty$ norm on both sides,
\begin{multline}\label{eqn:err_bound2}
\norm{Q_{K\backslash S,K\backslash S}(\xhr_{K\backslash S} - \xs_{K\backslash S})}_\infty\leq  \underbrace{\norm{(\EE{Q_{K\backslash S,K}} - Q_{K\backslash S, K})\xs}_\infty}_{\text{term 1}} + \underbrace{ \rho\alpha\norm{ \E{d_{K\backslash S}} -  d_{K\backslash S} }_\infty }_{\text{term 2}}\\ + \underbrace{\norm{Q_{K\backslash S,S}(\xhr_{S}- \xs_S)}_\infty}_{\text{term 3}}.
\end{multline}

\noindent
For term 1 we know that $\xs=u\ones_S + v\ones_{K}$ from Lemma~\ref{lem:exact_recovery}, so plugging in to term 1,
\begin{multline*}
(\EE{Q_{K\backslash S,K}} - Q_{K\backslash S, K})\xs
= - \frac{1-\alpha}{2}(A_{K\backslash S,S} - \E{A_{K\backslash S,S}})\cdot  u  \\+ \frac{1+\alpha}{2}(d_{K\backslash S} - \EE{d_{K\backslash S}}) \cdot v  - \frac{1-\alpha}{2} (  A_{K\backslash S ,K} -\EE{A_{K\backslash S, K} } )\ones_K\cdot  v.
\end{multline*}
%
%
Thus,
\begin{align*}
&\text{term 1} = \norm{(\EE{Q_{K\backslash S,K}} - Q_{K\backslash S, K})\xs}_\infty \\
&\leq  \frac{1-\alpha}{2}\norm{A_{K\backslash S,S} - \E{A_{K\backslash S,S}} }_\infty\cdot  u\\ 
& + \left(\frac{1+\alpha}{2}\norm{d_{K\backslash S} -\EE{d_{K\backslash S}}}_\infty +\frac{1-\alpha}{2}\norm{(  A_{K\backslash S ,K} -\EE{A_{K\backslash S, K} } )\ones_{K}}_\infty \right)\cdot   v  \\
 &\leq   \frac{1-\alpha}{2}\norm{A_{K\backslash S,S} - \E{A_{K\backslash S,S}} }_\infty\cdot  u + \delta \bar{d}\cdot  v\leq \frac{1-\alpha}{2} u+\delta\bar{d}\cdot  v   ,
\end{align*}
with probability at least $1-4e^{-c_0\delta^2 pk}$, where the second step uses the triangle inequality, the third step applies Lemma~\ref{lem:concen} and Lemma~\ref{lem:concen_degree}, and the last step holds since $\abs{A_{j,S}-p}\leq 1$ for all $j\in K\backslash S$. Furthermore, by Lemma~\ref{lem:concen_degree}, we can bound term 2 as
\begin{equation*}\text{term 2}\leq  \rho \alpha \cdot \delta \bar{d} ,\end{equation*}
while for term 3,  simple calculation leads to
\[\text{term 3} \leq \frac{1-\alpha}{2}\Abs{\xhr_S - \xs_S} . \]
Putting the results together,  we have
\begin{equation}\label{eqn:err_bound3}\norm{Q_{K\backslash S,K\backslash S}(\xhr_{K\backslash S} - \xs_{K\backslash S})}_\infty \leq \frac{1-\alpha}{2} u+\delta\bar{d}\cdot  v +  \rho \alpha \cdot \delta\bar{d}  + \frac{1-\alpha}{2}\Abs{\xhr_S - \xs_S}.  \end{equation}

Now it remains to upper bound the term $\Abs{\xhr_S - \xs_S}$. Following the same steps as before, indeed we can check that the following  holds:
\begin{multline}\label{eqn:err_bound4}
\norm{Q_{S,S}(\xhr_{ S} - \xs_{ S})}_\infty\leq  \underbrace{\norm{(\EE{Q_{ S,K}} - Q_{S, K})\xs}_\infty}_{\text{term 4}} + \underbrace{\rho\alpha  \norm{\E{d_{ S}} -   d_{S} }_\infty }_{\text{term 5}}\\ + \underbrace{\norm{Q_{ S,K\backslash S}(\xhr_{K\backslash S}- \xs_{K\backslash S})}_\infty}_{\text{term 6}}.
\end{multline}
Also, we can see that using $\xs=u\ones_S + v\ones_K$,
\[\text{term 4} =\norm{\frac{1+\alpha}{2}(u+v)(d_S - \bar{d}) - \frac{1-\alpha}{2}v((A_{S,K}-\EE{A_{S,K}})\ones_K) }_\infty \leq (u+v)\delta\bar{d},\]
where the inequality applies Lemma~\ref{lem:concen} and Lemma~\ref{lem:concen_degree}. It is also easy to see that $\text{term 5}\leq \rho\alpha\cdot \delta\bar{d}$, while
\begin{multline*} \text{term 6} = \frac{1-\alpha}{2}\norm{A_{S,K\backslash S} (\xhr_{K\backslash S} - \xs_{K\backslash S})}_\infty\leq\frac{1-\alpha}{2} \norm{A_{S,K\backslash S} }_1 \norm{\xhr_{K\backslash S} - \xs_{K\backslash S}}_\infty  \\ \leq \frac{1-\alpha}{2}(1+\delta)p(k-1) \norm{\xhr_{K\backslash S} - \xs_{K\backslash S}}_\infty ,\end{multline*}
where the second step uses H\"older's inequality and the next step applies Lemma~\ref{lem:concen}. Furthermore, $Q_{S,S}=\frac{1+\alpha}{2}d_S$ by definition, which is bounded below by $\frac{1+\alpha}{2}(1-\delta)\bar{d}$. Thus, from~\eqref{eqn:err_bound4}, we get
\begin{align*}
\norm{\xhr_S - \xs_S}_\infty &\leq  \frac{\delta\bar{d}(u+v) + \rho\alpha\cdot \delta\bar{d} + \frac{1-\alpha}{2}(1+\delta)p(k-1)\norm{\xhr_{K\backslash S} - \xs_{K\backslash S}}_\infty }{\frac{1+\alpha}{2}(1-\delta)\bar{d}}.
\end{align*}

Finally, return to~\eqref{eqn:err_bound3} and substitute the above bound in place of $\norm{\xhr_S - \xs_S}_\infty $, then
\begin{multline}\label{eqn:err_bound5}
\norm{Q_{K\backslash S, K\backslash S} (\xhr_{K\backslash S} - \xs_{K\backslash S})}_\infty \leq \frac{1-\alpha}{2}u +\delta\bar{d}\cdot v+\rho\alpha\cdot \delta\bar{d} \\+\left(\frac{1-\alpha}{1+\alpha}\right)\frac{\delta\bar{d}(u+v) + \rho\alpha\cdot\delta\bar{d} + \frac{1-\alpha}{2}(1+\delta)p(k-1)\norm{\xhr_{K\backslash S} - \xs_{K\backslash S}}_\infty  }{(1-\delta)\bar{d}}.
\end{multline}
The following lemma concerns the local strong convexity of the matrix $Q_{K\backslash S, K\backslash S}$ in $\ell_\infty$ norm.
\begin{lemma}
\label{lem:strong_convexity}
Under the conditions of Lemma~\ref{lem:l_infty_bound}, with probability at least $1-2e^{-c_0\delta^2 \bar{d}}$,
\begin{align*}
\norm{Q_{K\backslash S, K\backslash S} y}_\infty \geq \alpha(1-\delta)\bar{d}\norm{y}_\infty \text{  for all $y\in\R^{k-1}$}.
\end{align*}
\end{lemma}

\noindent
Applying Lemma~\ref{lem:strong_convexity} to the left-hand side of~\eqref{eqn:err_bound5}, and rearranging terms and simplifying, we have~that
\begin{multline*}
\left[\alpha(1-\delta)\bar{d} - \frac{\gamma}{2}\left(\frac{1-\alpha}{1+\alpha}\right)\left(\frac{1+\delta}{1-\delta}\right) \right] \norm{\xhr_{K\backslash S} - \xs_{K\backslash S}}_\infty   \\\leq \left(\frac{1-\alpha}{2}+ \frac{1-\alpha}{1+\alpha}\frac{\delta}{1-\delta} \right)u  +\left( \delta \bar{d} +\frac{1-\alpha}{1+\alpha}\frac{\delta}{1-\delta} \right) v + \rho\alpha\left(\delta\bar{d} + \frac{1-\alpha}{1+\alpha}\frac{\delta}{1-\delta} \right),
\end{multline*}
where we use $\alpha <1$ and $p(k-1)=\gamma\bar{d}$. By assumption of the lemma, we have that $\delta\leq 0.1$ and that $\alpha\in [0.1,0.9]$  while $\gamma\in(0,1)$ by definition~\eqref{eqn:gamma}, so
\[ \left[ \alpha (1-\delta)\bar{d} - \frac{\gamma}{2}\left(\frac{1-\alpha}{1+\alpha}\right)\left(\frac{1+\delta}{1-\delta}\right) \right] \geq c_1\bar{d} \;\;  \text{ and } \;\; \frac{1-\alpha}{1+\alpha}\frac{\delta}{1-\delta} \leq \delta\bar{d}, \]
for some numerical constant $c_1>0$. As a result,
\[ c_1\bar{d}\norm{\xhr_{K\backslash S} - \xs_{K\backslash S}}_\infty  \leq (1-\alpha)u + 2\delta\bar{d} (v + \rho\alpha). \]
Dividing both sides by $c_1\bar{d}$,
\[ \norm{\xhr_{K\backslash S} - \xs_{K\backslash S}}_\infty \leq  \frac{c_1^{-1}(1-\alpha)u}{\bar{d}} + 2c_1^{-1}\delta( v + \rho\alpha) . \]
Using the triangle inequality, Lemma~\ref{lem:l_infty_bound} now follows.



\subsection{Proof of Lemma~\ref{lem:xh_vol_bound}}

By Lemma~\ref{lem:nonneg_grad} we know that 
\begin{equation*}
(Q\xh  - \alpha \seed)_i \begin{cases} =-\rho\alpha d_i, & \xh_i >0, \\ \leq 0, & \xh_i = 0. \end{cases}
\end{equation*}
Summing over $i$'s, we have 
\begin{align*}
\sum_{i} (Q\xh  - \alpha \seed)_i  &= \alpha \sum_{i} d_i \xh_i- \alpha \\ 
&\leq -\rho\alpha \sum_{j:\xh_j >0} d_j = -\rho\alpha\cdot \vol(\supp(\xh)),
\end{align*}
where the first equality holds since $\ones^\top Q \xh = \ones^\top (\alpha D + (1-\alpha)/2\cdot \lap) \xh = \alpha\sum_i d_i\xh_i $.
Dividing both sides by $-\rho\alpha$, we get
\[\vol(\supp(\xh))  \leq \frac{1-d^\top \xh}{\rho} . \]


\subsection{Proof of Lemma~\ref{lem:seed_node}}

This result follows from simple probability calculation. 
We have
\begin{align*}
\PP{\cup_{i\in K} (\text{$i$ is not connected to $K^c$})} &=1 - \PP{\cap_{i\in K} (\text{$i$ is connected to $K^c$})} \\
&= 1 - (\PP{ \text{$i$ is connected to $K^c$}})^{k}\\
&= 1 - (1-\PP{ \text{$i$ is not connected to $K^c$}})^{k}\\
&= 1- (1-(1-q)^{n-k})^{k}\\
&\geq 1-(1-(1-q)^{n})^{k}.
\end{align*}
Here, the second and fourth steps hold since each pair of nodes has an edge independently of all other pairs. Now suppose that $n$ is sufficiently large so that $q\leq 0.5$. Then, we have $(1-q)^n \geq \exp(-1.5 c)$, which leads to
\[\PP{\cup_{i\in K} (\text{$i$ is not connected to $K^c$})}  \geq 1- (1-\exp(-1.5c))^k. \]

\subsection{Proof of Lemma~\ref{lem:rho}}

Letting $\bar{c}=\left(\frac{1-\alpha}{1+\alpha}\right) \left( \frac{1-\delta}{1+\delta}\right)^2 \frac{\gamma}{1+\delta}$, we can write $\rho=\rho(\delta)=\frac{\bar{c}(1-\alpha)p}{(1+\alpha)\bar{d}^2}$. Following the same steps as~\eqref{eqn:rho_cond}  (proof of Lemma~\ref{lem:exact_recovery}), we can see
\begin{align}\label{eqn:rho_ineq}
\rho > \rho^\sharp \;\;  &\Longleftrightarrow \;\; 2\bar{c}p\alpha\cdot \bar{d}\cdot \EE{{d}_{\ell_\sharp}} +\bar{c} pq(1-\alpha)\cdot k\bar{d} + \bar{c}pq(1-\alpha)(n-k)\EE{{d}_{\ell_\sharp}} > q(1+\alpha)\bar{d}^2  \nonumber\\
&\Longleftrightarrow \;\; 2\alpha \bar{d}(\bar{c}p  \EE{{d}_{\ell_\sharp}} - q\bar{d})  + (1-\alpha)q(n-k)(\bar{c}p\EE{{d}_{\ell_\sharp}} - q\bar{d}) \\\nonumber 
& \hspace{2.9in}- pq(1-\alpha)(k-1)\bar{d}(1-\bar{c})>0.
\end{align}
Since $\delta\leq 0.1$ and $\alpha\in [0.1,0.9]$, we know that $\bar{c}\geq c' \gamma$ for some constant $c'>0$. Also, since $q=\frac{c}{n}$ while $p=k=\O{1}$ by Assumption~\ref{assump:rate}, we have that $\gamma = \O{1}$, and thus
\[ \begin{cases} 2\alpha \bar{d}(\bar{c}p  \EE{{d}_{\ell_\sharp}} - q\bar{d})  + (1-\alpha)q(n-k)(\bar{c}p\EE{{d}_{\ell_\sharp}} - q\bar{d}) =\O{\bar{d}} ;\\  pq(1-\alpha)(k-1)\bar{d}(1-\bar{c}) = \O{\frac{k\bar{d}}{n}}. \end{cases} \]
Therefore, if $n$ is sufficiently large, the first two terms on the right-hand side of~\eqref{eqn:rho_ineq} are positive and much larger than the last term. 
This proves that $\rho>\rho^\sharp$.

To see $\rho<\rho^\natural$, it suffices to check
\[ \left(\frac{1-\alpha}{1+\alpha}\right)^2\frac{p}{\bar{d}^2} < \frac{p(1-\alpha)}{\bar{d}\left[(1+\alpha)\bar{d} +(1-\alpha)p\right]} = \rho^\natural.\]
Rearranging the terms,  we equivalently need to show $2\alpha(1+\alpha)\bar{d} > (1-\alpha)^2 p$. Since $\bar{d}\geq p\cdot (k-1)$, we then have $2\alpha(1+\alpha) \cdot (k-1) > (1-\alpha)^2$ which, due to the assumption  $\alpha\in [0.1,0.9]$, holds for any $k\geq 5$.


\subsection{Proof of Lemma~\ref{lem:strong_convexity}}

Denoting by $\ones$ the vector whose entries are all ones, we have
\begin{align*}\label{eqn:q_str}
Q_{K\backslash S, K\backslash S} &= \alpha D_{K\backslash S, K\backslash S} + \frac{1-\alpha}{2}(D_{K\backslash S, K\backslash S}- A_{K\backslash S, K\backslash S}) \nonumber \\
&=  \alpha D_{K\backslash S, K\backslash S}  + \frac{1-\alpha}{2}(\text{diag}\left(A\ones \right)_{K\backslash S, K\backslash S} -  A_{K\backslash S, K\backslash S}) \nonumber \\
&= \alpha D_{K\backslash S, K\backslash S} + \frac{1-\alpha}{2}  \text{diag}\left(A_{K\backslash S, S\cup K^c} \ones_{S\cup K^c} \right) + \frac{1-\alpha}{2} L_{K\backslash S, K\backslash S},
\end{align*}
where $L_{K\backslash S, K\backslash S}$ is the graph Laplacian of the sub-graph induced by $A_{K\backslash S, K\backslash S}$.

Now assume $\norm{y}_\infty=1$. Without loss of generality, we will assume $y_1=1$ (the same argument holds in the case of $y_1=-1$). Then, by definition of graph Laplacian, it is easy to see $(L_{K\backslash S, K\backslash S}\cdot y)_1 \geq 0$, and so
\[ (Q_{K\backslash S, K\backslash S}\cdot y)_1 \geq \alpha d_1 . \]
Hence, applying degree concentration to $d_1$ (Lemma~\ref{lem:concen_degree}),
\[ \norm{Q_{K\backslash S, K\backslash S}\cdot y}_\infty \geq (Q_{K\backslash S, K\backslash S}\cdot y)_1 \geq \alpha d_1   \geq  (1-\delta)\alpha\bar{d} ,\]
with probability at least $1-2e^{-c_0\delta^2 \bar{d}}$, proving the result.

\vskip 0.2in
\bibliography{lsc}

\end{document}